\documentclass{article}
\usepackage{spconf,amsmath,graphicx}

\usepackage[colorlinks=true,allcolors=black]{hyperref}
\usepackage{url}
\usepackage{textcomp} 
\usepackage{gensymb} 
\usepackage{arydshln}
\usepackage{booktabs}
\usepackage{multirow}
\usepackage[utf8]{inputenc}
\usepackage[acronym]{glossaries}
\usepackage[caption=false]{subfig}
\usepackage[table,dvipsnames]{xcolor}
\usepackage{xspace} 
\usepackage[sort&compress,numbers]{natbib}
\usepackage{balance}

\usepackage{amssymb}
\usepackage{latexsym}
\usepackage{url}
\usepackage{enumitem}

\newacronymstyle{long-short-br}
{%
  \GlsUseAcrEntryDispStyle{long-short}%
}%
{%
  \GlsUseAcrStyleDefs{long-short}%
}
\setacronymstyle{long-short-br}

\title{\Large Ocular Recognition Databases and Competitions: A Survey}

\makeatletter
\def\@name{ \emph{Luiz~A.~Zanlorensi\textsuperscript{1}, Rayson Laroca\textsuperscript{1}, Eduardo~Luz\textsuperscript{2},}  \\[0.1ex] \emph{Alceu~S.~Britto~Jr.\textsuperscript{3}, Luiz~S.~Oliveira\textsuperscript{1}, David~Menotti\textsuperscript{1}}\thanks{
This is an author-prepared version. The final publication is available at \textit{Springer}~(DOI: \href{http://doi.org/10.1007/s10462-021-10028-w}{\textcolor{blue}{10.1007/s10462-021-10028-w}}).} \\}

\makeatother

\address{\textsuperscript{1}Department of Informatics, Federal University of Paran\'a, Curitiba, Brazil\\\textsuperscript{2}Computing Department, Federal University of Ouro Preto, Belo Horizonte, Brazil\\\textsuperscript{3}Postgraduate Program in Informatics, Pontifical Catholic University of Paran\'a, Curitiba, Brazil\\[0.5ex]
\normalsize
\textsuperscript{1}\textit{\{lazjunior, rblsantos, lesoliveira, menotti\}@inf.ufpr.br} \qquad \textsuperscript{2}\textit{eduluz@ufop.edu.br}\qquad \textsuperscript{3}\textit{alceu.junior@pucpr.br}}

\begin{document}
\ninept
\sloppy
\maketitle
\newacronym{nir}{NIR}{near-infrared wavelength}
\newacronym{vis}{VIS}{visible wavelength}
\newacronym{rr}{RR}{Recognition Rate}
\newacronym{auc}{AUC}{Area Under the Curve}
\newacronym{eer}{EER}{Equal Error Rate}
\newacronym{fnmr}{FNMR}{False Non-Match Rate}
\newacronym{fmr}{FMR}{False Match Rate}
\newacronym{svm}{SVM}{Support Vector Machines}
\newacronym{pca}{PCA}{Principal Component Analysis}
\newacronym{lda}{LDA}{Linear Discriminant Analysis}
\newacronym{gfar}{GFAR}{Generalized False Accept Rate}
\newacronym{gffr}{GFFR}{Generalized False Reject Rate}
\newacronym{cnn}{CNN}{Convolutional Neural Network}
\newacronym{fcn}{FCN}{Fully Convolutional Network}
\newacronym{fte}{FTE}{Failure-to-enroll}
\newacronym{fta}{FTA}{Failure-to-acquire}
\newacronym{safe}{SAFE}{Symmetry Patterns}
\newacronym{gabor}{GABOR}{Gabor Spectral Decomposition}
\newacronym{sift}{SIFT}{Scale-Invariant Feature Transform}
\newacronym{lbp}{LBP}{Local Binary Patterns}
\newacronym{hog}{HOG}{Histogram of Oriented Gradients}
\newacronym{gan}{GAN}{Generative Adversarial Network}
\newacronym{vlad}{VLAD}{Vector of Locally Aggregated Descriptors}

\newcommand{\casiavOne}{CASIA-IrisV1\xspace}
\newcommand{\casiavTwo}{CASIA-IrisV2\xspace}
\newcommand{\ndIris}{ND-IRIS-0405\xspace}
\newcommand{\iceOne}{ICE 2005\xspace}
\newcommand{\iceTwo}{ICE 2006\xspace}
\newcommand{\wvuSIT}{WVU Synthetic Iris Texture Based\xspace}
\newcommand{\wvuSIM}{WVU Synthetic Iris Model Based\xspace}
\newcommand{\fakeIris}{Fake Iris Database\xspace}
\newcommand{\casiaInterval}{CASIA-IrisV3-Interval\xspace}
\newcommand{\casiaLamp}{CASIA-IrisV3-Lamp\xspace}
\newcommand{\casiaTwins}{CASIA-IrisV3-Twins\xspace}
\newcommand{\casiaThousand}{CASIA-IrisV4-Thousand\xspace}
\newcommand{\casiaSyn}{CASIA-IrisV4-Syn\xspace}
\newcommand{\iitdIris}{IIT Delhi Iris\xspace}
\newcommand{\ndICL}{ND Iris Contact Lenses 2010\xspace}
\newcommand{\ndITA}{ND Iris Template Aging\xspace}
\newcommand{\ndTLI}{ND TimeLapseIris\xspace}
\newcommand{\iiitdIUAI}{IIITD IUAI\xspace}
\newcommand{\iiitdCLI}{IIITD CLI\xspace}
\newcommand{\ndCCL}{ND Cosmetic Contact Lenses\xspace}
\newcommand{\ndcsIris}{ND Cross-Sensor-Iris-2013\xspace}
\newcommand{\irisPrint}{Database of Iris Printouts\xspace}
\newcommand{\iiitdIS}{IIITD Iris Spoofing\xspace}
\newcommand{\ndCLD}{NDCLD15\xspace}
\newcommand{\iiitdCS}{IIITD Combined Spoofing\xspace}
\newcommand{\ndGFI}{ND-GFI\xspace}
\newcommand{\berc}{BERC mobile-iris database\xspace}
\newcommand{\casiaMob}{CASIA-Iris-Mobile-V1.0\xspace}
\newcommand{\openEDS}{OpenEDS\xspace}
\newcommand{\csi}{Cataract Surgery on Iris\xspace}
\newcommand{\muid}{MUID\xspace}
\newcommand{\ornl}{ORNL\xspace}

\newcommand{\upol}{UPOL\xspace}
\newcommand{\ubirisvOne}{UBIRIS.v1\xspace}
\newcommand{\utiris}{UTIRIS\xspace}
\newcommand{\ubirisvTwo}{UBIRIS.v2\xspace}
\newcommand{\ubipr}{UBIPr\xspace}
\newcommand{\bdcp}{BDCP\xspace}
\newcommand{\mobbioFake}{MobBIOfake\xspace}
\newcommand{\iiitdMSP}{IIITD Multi-spectral Periocular\xspace}
\newcommand{\polyu}{PolyU Cross-Spectral\xspace}
\newcommand{\miche}{MICHE-I\xspace}
\newcommand{\vssiris}{VSSIRIS\xspace}
\newcommand{\csip}{CSIP\xspace}
\newcommand{\visob}{VISOB\xspace}
\newcommand{\crossEyed}{CROSS-EYED\xspace}
\newcommand{\qutMP}{QUT Multispectral Periocular\xspace}
\newcommand{\visobtwo}{VISOB 2.0\xspace}
\newcommand{\ufprEyeglass}{UFPR-Eyeglasses\xspace}
\newcommand{\ufprPerioc}{UFPR-Periocular\xspace}
\newcommand{\pmhi}{Post-mortem Human Iris\xspace}
\newcommand{\isocialdb}{I-SOCIAL-DB\xspace}

\newcommand{\biosec}{BioSec\xspace}
\newcommand{\biosecur}{BiosecurID\xspace}
\newcommand{\bmdb}{BMDB\xspace}
\newcommand{\mbgc}{MBGC\xspace}
\newcommand{\qFire}{Q-FIRE\xspace}
\newcommand{\focs}{FOCS\xspace}
\newcommand{\casiaDistance}{CASIA-IrisV4-Distance\xspace}
\newcommand{\sdumla}{SDUMLA-HMT\xspace}
\newcommand{\mobbio}{MobBIO\xspace}
\newcommand{\gbMOD}{gb2s$\mu$ MOD\xspace}

\newcommand{\imgsize}{0.19}

\begin{abstract}
\textit{The use of the iris and periocular region as biometric traits has been extensively investigated, mainly due to the singularity of the iris features and the use of the periocular region when the image resolution is not sufficient to extract iris information.
In addition to providing information about an individual's identity, features extracted from these traits can also be explored to obtain other information such as the individual's gender, the influence of drug use, the use of contact lenses, spoofing, among others.
This work presents a survey of the databases created for ocular recognition, detailing their protocols and how their images were acquired.
We also describe and discuss the most popular ocular recognition competitions (contests), highlighting the submitted algorithms that achieved the best results using only iris trait and also fusing iris and periocular region information.
Finally, we describe some relevant works applying deep learning techniques to ocular recognition and point out new challenges and future directions. 
Considering that there are a large number of ocular databases, and each one is usually designed for a specific problem, we believe this survey can provide a broad overview of the challenges in ocular biometrics.}
\end{abstract}

\section{Introduction}
\label{sec:intro}

\glsresetall

Several corporations and governments fund biometrics research due to various applications such as combating terrorism and the use of social networks, showing that this is a strategically important research area~\cite{Daugman2006, Phillips2009}.
A biometric system exploits pattern recognition techniques to extract distinctive information/signatures of a person.
Such signatures are stored and used to compare and determine the identity of a person sample within a population.
As biometric systems require robustness against acquisition and/or preprocessing fails, as well as high accuracy, the challenges and the methodologies for identifying individuals are constantly~developing.

Methods that identify a person based on their physical or behavioral features are particularly important since such characteristics cannot be lost or forget, as may occur with passwords or identity cards~\cite{Bowyer2008}.
In this context, the use of ocular information as a biometric trait is interesting regarding a noninvasive technology and also because the biomedical literature indicates that irises are one of the most distinct biometric~sources~\cite{Wildes1997}.

The most common task in ocular biometrics is recognition, which can be divided into verification ($1:1$ comparison) and identification ($1:n$ comparison).
Also, recognition can be performed in two distinct protocols called closed-world and open-world.
In the closed-world protocol, samples of an individual are present in the training and test set.
The open-world protocol must have samples from different subjects both in the training and test sets.
The identification process generally is performed on the closed-world protocol (except the open-set scenario, which has imposters that are only in the test set, i.e., individuals who should not match any subject in the gallery set), while verification can be performed in both, being the open-world most common protocol adopted in this setup.
In addition to identification and verification, there are other tasks in ocular biometrics such as gender classification~\cite{Tapia2017}, spoofing~\cite{Menotti2015} and liveness~\cite{He2016} detection, recognition of left and right iris images~\cite{Du2016}, ocular region detection~\cite{severo2018benchmark, lucio2019simultaneous}, iris/sclera segmentation~\cite{lucio2018fully, bezerra2018robust}, and sensor model identification~\cite{Marra2017}.

\newcolumntype{C}[1]{>{\centering\let\newline\\\arraybackslash\hspace{0pt}}m{#1}}
\newcolumntype{L}[1]{>{\let\newline\\\arraybackslash\hspace{0pt}}m{#1}}

\begin{figure}[!ht]
\centering
\includegraphics[width=\imgsize\columnwidth]{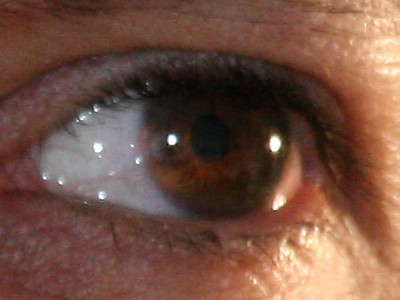}
\includegraphics[width=\imgsize\columnwidth]{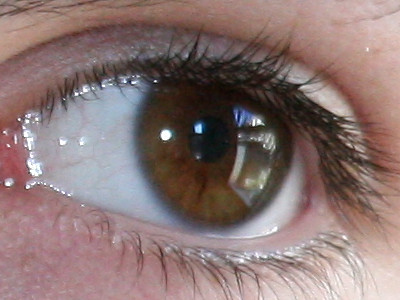}
\includegraphics[width=\imgsize\columnwidth]{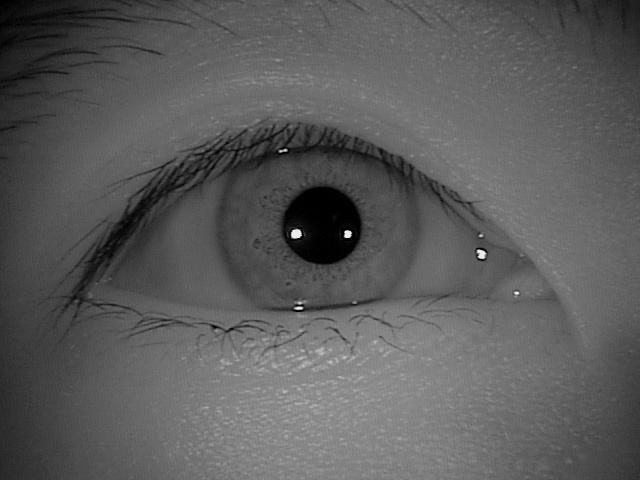} 
\includegraphics[width=\imgsize\columnwidth]{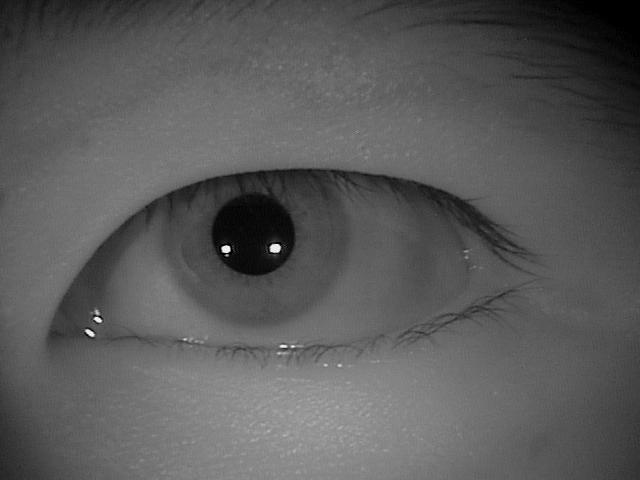} 

\vspace{0.75mm}

\includegraphics[width=\imgsize\columnwidth]{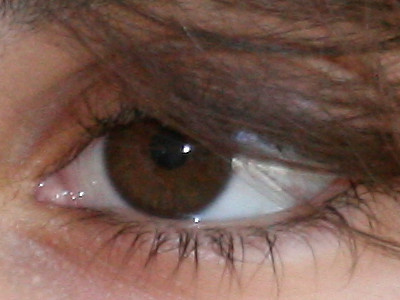}
\includegraphics[width=\imgsize\columnwidth]{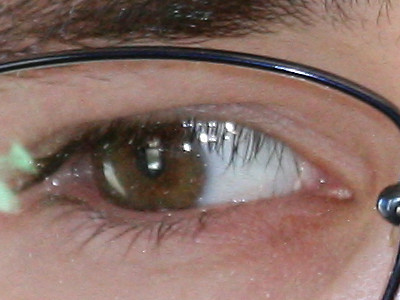}
\includegraphics[width=\imgsize\columnwidth]{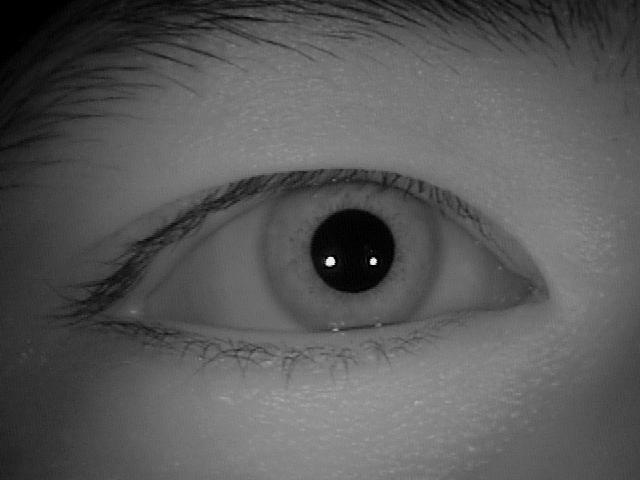}
\includegraphics[width=\imgsize\columnwidth]{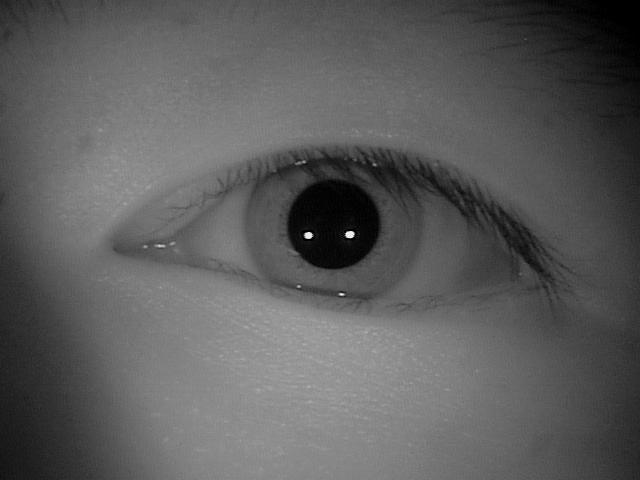} 

\vspace{0.75mm}

\includegraphics[width=\imgsize\columnwidth]{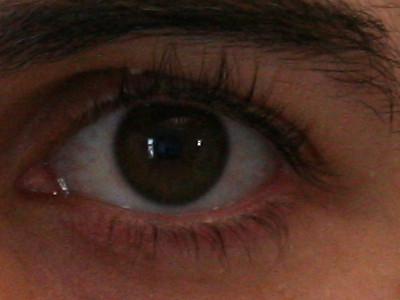} \includegraphics[width=\imgsize\columnwidth]{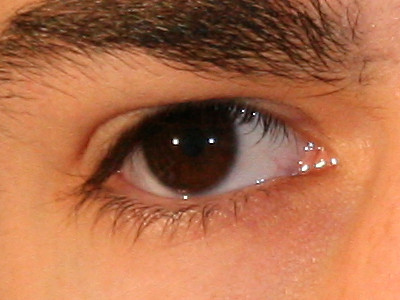} \includegraphics[width=\imgsize\columnwidth]{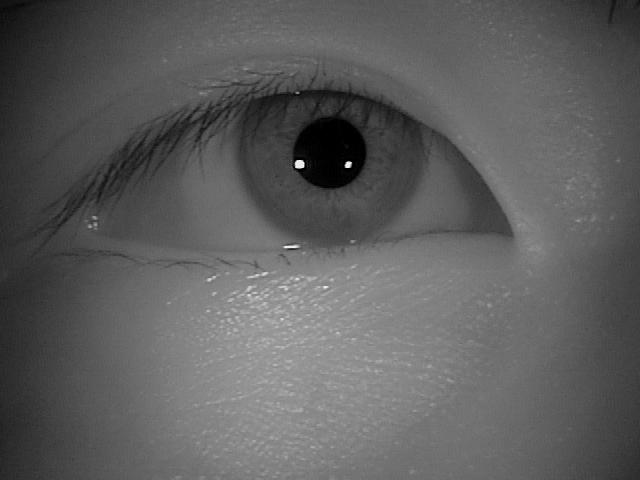} \includegraphics[width=\imgsize\columnwidth]{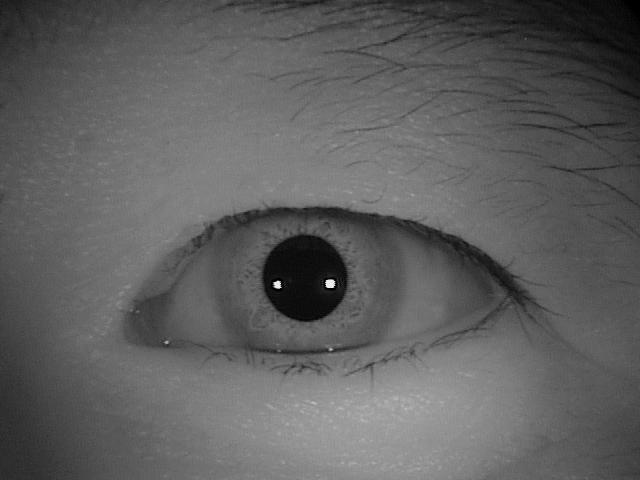} 
\caption{\ubirisvTwo \cite{proenca2010}: uncontrolled environment images at \acrlong*{vis}~(left) and \casiaThousand \cite{CASIA2010}: controlled environment images at \acrlong*{nir}~(right).}
\label{figubiriscasia}
\end{figure}

Iris recognition under controlled environments at \gls*{nir} demonstrates impressive results, and as reported in several works~\cite{Bowyer2008, Phillips2008, Phillips2010, Proenca2017irina, Proenca2019segmentation} can be considered a mature technology.
The use of ocular images captured in uncontrolled environments is currently one of the greatest challenges~\cite{Proenca2012, Rattani2016}.
As shown in Fig.~\ref{figubiriscasia}, such images usually present noise caused by illumination, occlusion, reflection, motion blur, among others.
Therefore, to improve the biometric systems performance in these scenarios, recent approaches have used information extracted only from the periocular region~\cite{Padole2012, Proenca2018, Luz2018} or fusing them with iris features~\cite{Tan2012, Tan2013, Ahmed2016, Ahmed2017}.

\begin{table*}[!ht]
\setlength{\tabcolsep}{8pt}
\centering
\caption{Ocular databases in previous surveys.}
\label{surveys}
\resizebox{\textwidth}{!}{
\begin{tabular}{lcc}
\toprule
Survey                                                                                  & Year    & Databases Described   \\
\midrule
Image understanding for iris biometrics: A survey \cite{Bowyer2008}                       & 2008    &  $10$ iris \\
Iris image classification: A survey~\cite{Hake2015}                                       & 2015    &  $8$ iris \\
Ocular biometrics: A survey of modalities and fusion approaches~\cite{Nigam2015}          & 2015    &  $23$ iris, $5$ periocular, $5$ iris/periocular  \\
Periocular biometrics: databases, algorithms and directions~\cite{Alonso-Fernandez2016a}  & 2016    &  $5$ iris, $4$ periocular \\
A survey on periocular biometrics research~\cite{Alonso-Fernandez2016b}                   & 2016    &  $5$ iris, $4$ periocular \\
Long range iris recognition: A survey~\cite{Nguyen2017}                                   & 2017    &  $3$ iris \\
Ocular biometrics in the visible spectrum: A survey~\cite{Rattani2017}                    & 2017    &  $7$ ocular \\
Overview of the combination of biometric matchers \cite{Lumini2017}                       & 2017    &  $8$ multimodal with iris \\
\midrule
\multirow{2}{*}{\textbf{This paper}} & \multirow{2}{*}{\textbf{2021}}    & $\textbf{40}$ \textbf{iris,} $\textbf{5}$ \textbf{iris/periocular}, \\ &  & $\textbf{7}$ \textbf{periocular}, $\textbf{10}$ \textbf{multimodal} \\

\bottomrule

\end{tabular}
}
\end{table*}

\begin{figure}[!ht]
\begin{center}
 \includegraphics[width=.7\linewidth]{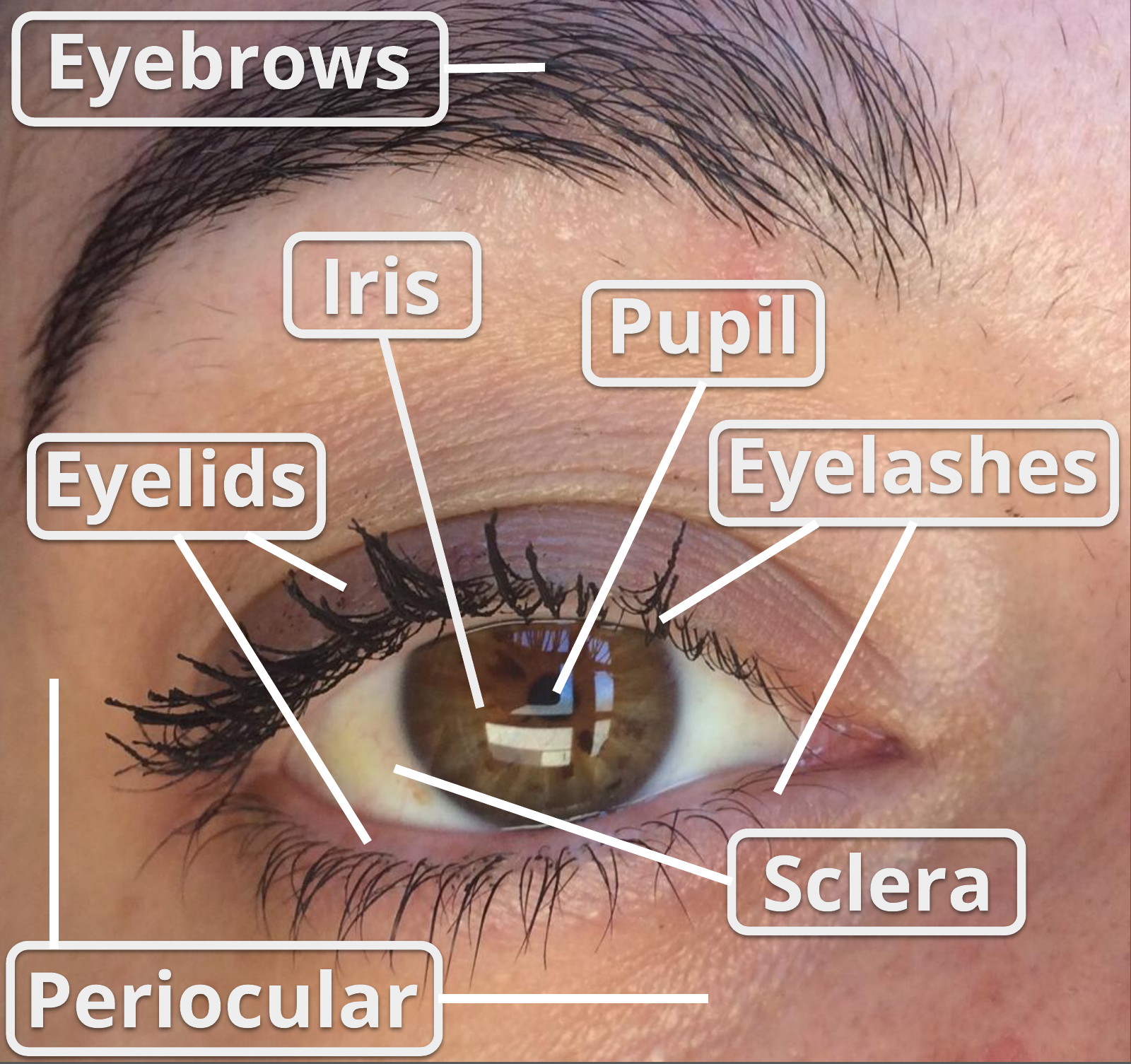}
\end{center}  
\vspace{-3mm}
\caption{Ocular components.}
\label{fig:eye}
\end{figure}

The term periocular is associated with the region around the eye, composed of eyebrows, eyelashes, and eyelids~\cite{Park2009, Park2011, Uzair2015}, as illustrated in Fig.~\ref{fig:eye}.
Usually, the periocular region is used when there is poor quality in the iris region, commonly in~\gls{vis} images or part of the face is occluded (in face images)~\cite{Park2009, Luz2018}.
In the literature, regarding the periocular region, there are works that kept the iris and sclera regions~\cite{Luz2018, Proenca2012, DeMarsico2017} and others that removed them~\cite{Sequeira2016, Sequeira2017, Proenca2018}.

Although there are several surveys in the literature describing ocular recognition methodologies~\cite{Wildes1997, Bowyer2008, Ross2010, Hake2015, Nigam2015, Alonso-Fernandez2016a, Alonso-Fernandez2016b, DeMarsico2016, Nguyen2017, Rattani2017, Lumini2017}, such surveys do not specifically focus on databases and competitions.
Table~\ref{surveys} summarizes the number of ocular databases/competitions described in these surveys.

One of the first surveys on iris recognition was presented by Wildes~\cite{Wildes1997}, who examined iris recognition biometric systems as well as issues in the design and operation of such systems.
Bowyer et al.~\cite{Bowyer2008} described both the historical and the state-of-the-art development in iris biometrics focusing on segmentation and recognition methodologies.
Addressing long-range iris recognition, the literature review described in~\cite{Nguyen2017} presents and describes iris recognition methods at a distance system.
Alonso-Fernandez et al.~\cite{Alonso-Fernandez2016a, Alonso-Fernandez2016b} surveyed methodologies focusing only on periocular biometrics, while Rattani and Derakhshani~\cite{Rattani2017} described state-of-the-art methods applied to periocular region, iris, and conjunctival vasculature recognition using \gls*{vis} images.
Nigam et al.~\cite{Nigam2015} described in detail methodologies for specific topics such as iris acquisition, preprocessing techniques, segmentation approaches, in addition to feature extraction, matching and indexing~methods.
Lastly, Omelina et al.~\cite{omelina2021survey} recently performed an extensive survey regarding iris databases, describing properties of popular databases and recommendations to create a good iris database. 
The authors also made a brief description of some ocular~competitions.

This work describes ocular databases and competitions (or~contests) on biometric recognition using iris and/or periocular traits.
We present the databases according to the images that compose them, i.e.,~\gls{nir}, \gls{vis} and Cross-Spectral, and multimodal databases.
We also detailed information such as image wavelength, capture environment, cross-sensor, database size and  ocular modalities employed, as well as the protocol used for image acquisition and database construction.

The main contributions of this paper are the following: \textbf{(i)}~we survey and describe the types of existing ocular images databases and image acquisition protocols; \textbf{(ii)}~a detailed description of the applications and goals in creating these~databases; \textbf{(iii)}~a discussion and description of the main and most popular ocular recognition competitions in order to illustrate the methodology strategies in each challenge; and \textbf{(iv)}~we drawn new challenging tasks and scenarios in ocular~biometrics.

To the best of our knowledge, this is the first survey specifically focused on ocular databases and competitions.
Thus, we believe that it can provide a general overview of the challenges in ocular recognition over the years, the databases used in the literature, as well as the best performance methodologies in competitions for different scenarios.

The remainder of this work is organized as follows.
In Section~\ref{sec:eyeDataBase}, we detail the ocular databases separating them into three categories: \gls{nir}, \gls{vis} and cross-spectral, and multimodal databases. 
In Section~\ref{sec:eyeCont}, we present a survey and discussion of ocular recognition competitions using iris and periocular region information and describe the top-ranked methodologies.
Section~\ref{sec:deepeye} presents recent works applying deep learning frameworks focusing on encoding and matching to iris/periocular recognition and other tasks regarding ocular biometrics (ocular preprocessing methods are beyond the scope of our review). 
Finally, future challenges and directions are pointed out in~Section~\ref{sec:future} and conclusions are given in~Section~\ref{sec:conclusion}.
\section{Ocular Databases}
\label{sec:eyeDataBase}

Currently, there are various databases of ocular images, constructed in different scenarios and for different purposes.
These databases can be classified by \gls*{vis} and \gls*{nir} images and separated into controlled (cooperatives) and uncontrolled (non-cooperatives) environments, according to the process of image acquisition.
Controlled databases contain images captured in environments with controlled conditions, such as lighting, distance, and focus.
On the other hand, uncontrolled databases are composed of images obtained in uncontrolled environments and usually present problems such as defocus, occlusion, reflection, off-angle, to cite a few.
A database containing images captured at different wavelengths is referred to as cross-spectral, while a database with images acquired by different sensors is referred to as cross-sensor.
The summary of all databases cited in this paper as well as links to find more information about how they are available can be found at [\url{www.inf.ufpr.br/vri/publications/ocularDatabases.html}].

In this Section, the ocular databases are presented and organized into three subsections.
First, we describe databases that contain only \gls{nir} images, as well as synthetic iris databases.
Then, we present databases composed of images captured at both \gls{vis} and cross-spectral scenarios (i.e., \gls{vis} and \gls{nir} images from the same subjects).
Finally, we describe multimodal databases, which contain data from different biometric traits, including iris and/or periocular.

\subsection{Near-Infrared Ocular Images Databases}
\label{subsec:nir-databases}

\begin{figure}[!ht]
\centering
\includegraphics[width=\imgsize\columnwidth]{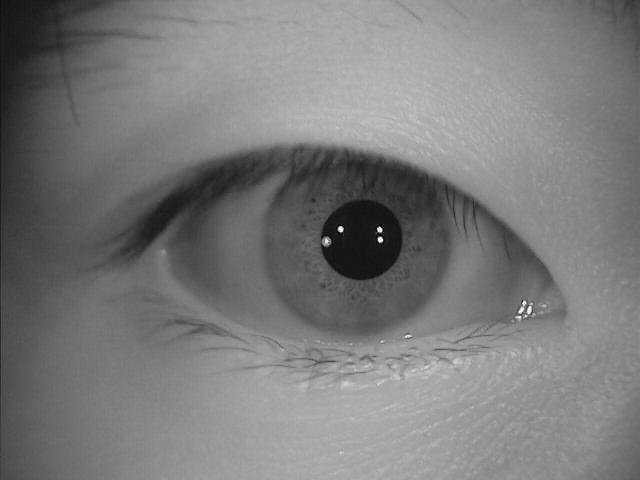}
\includegraphics[width=\imgsize\columnwidth]{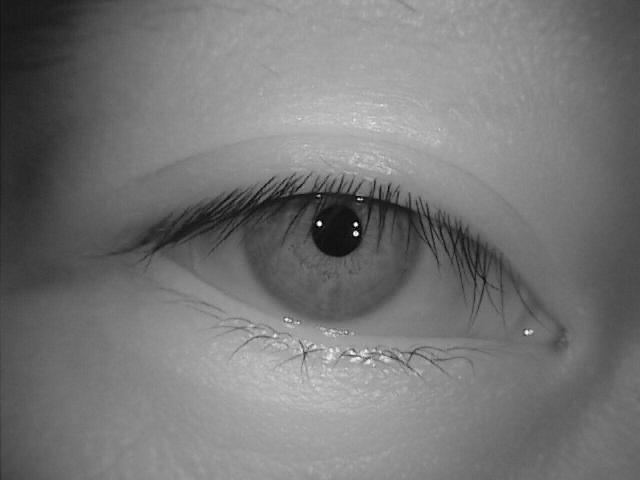}
\includegraphics[width=\imgsize\columnwidth]{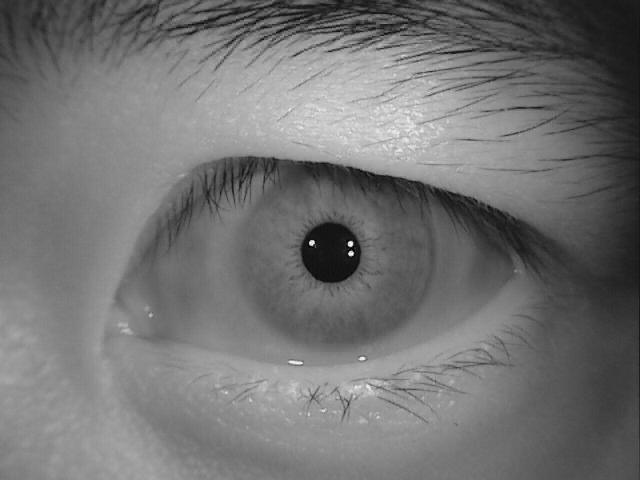}
\includegraphics[width=\imgsize\columnwidth]{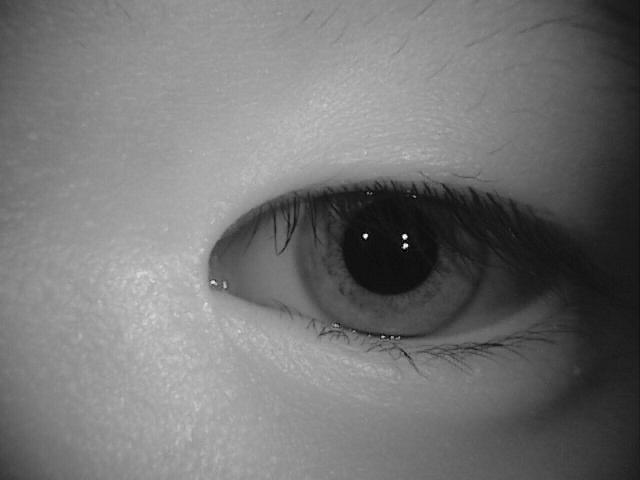}
\includegraphics[width=\imgsize\columnwidth]{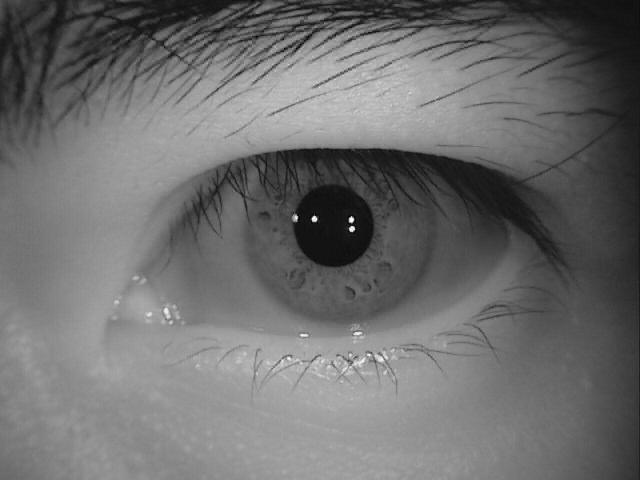}

\vspace{1mm}

\includegraphics[width=\imgsize\columnwidth]{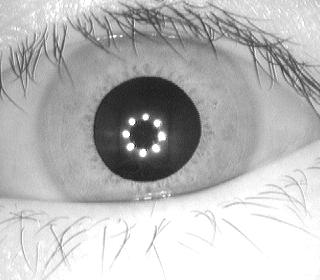}
\includegraphics[width=\imgsize\columnwidth]{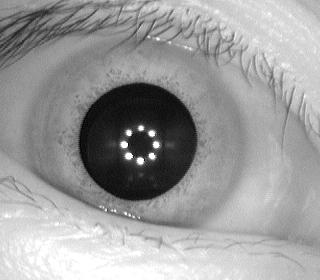}
\includegraphics[width=\imgsize\columnwidth]{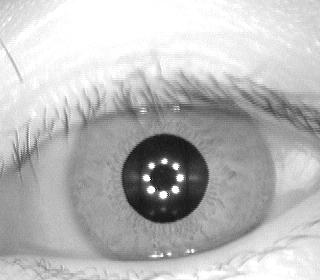}
\includegraphics[width=\imgsize\columnwidth]{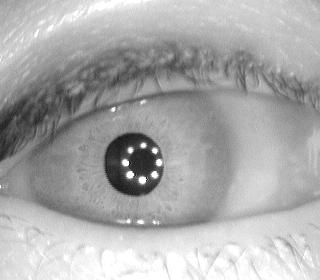}
\includegraphics[width=\imgsize\columnwidth]{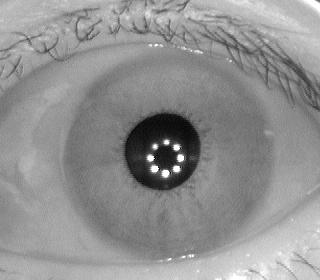}

\vspace{1mm}

\includegraphics[width=\imgsize\columnwidth]{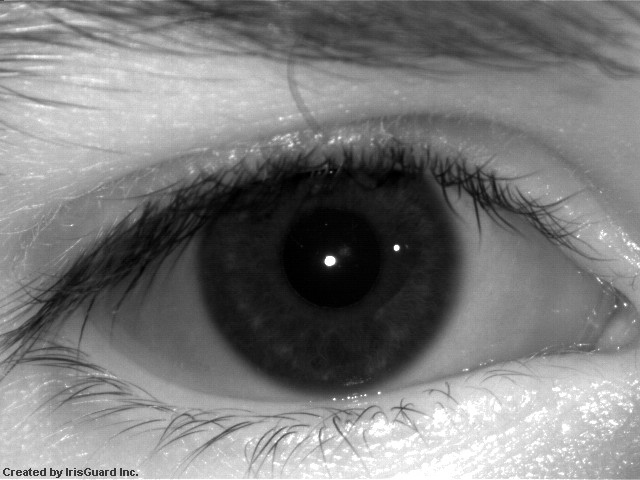}
\includegraphics[width=\imgsize\columnwidth]{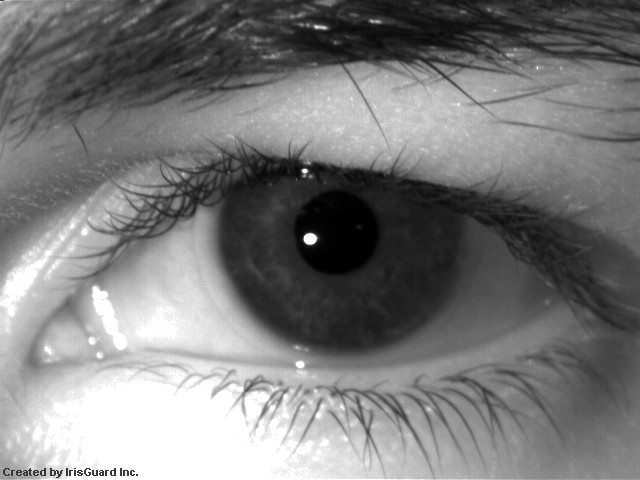}
\includegraphics[width=\imgsize\columnwidth]{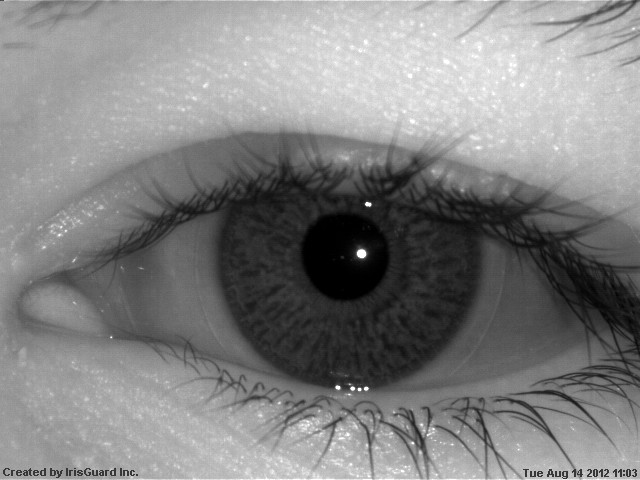}
\includegraphics[width=\imgsize\columnwidth]{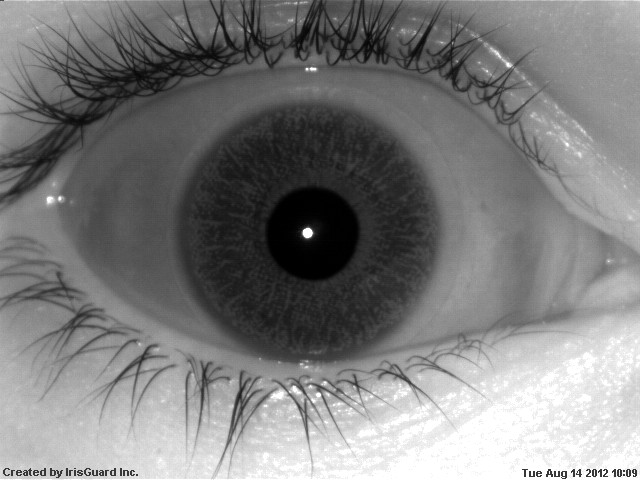}
\includegraphics[width=\imgsize\columnwidth]{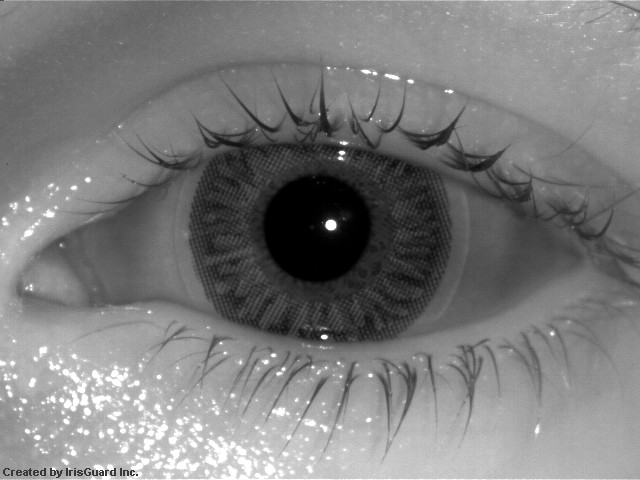}

\vspace{1mm}

\includegraphics[width=\imgsize\columnwidth]{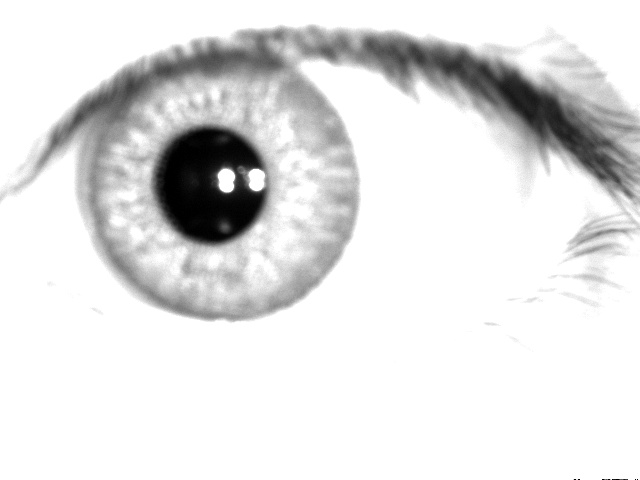}
\includegraphics[width=\imgsize\columnwidth]{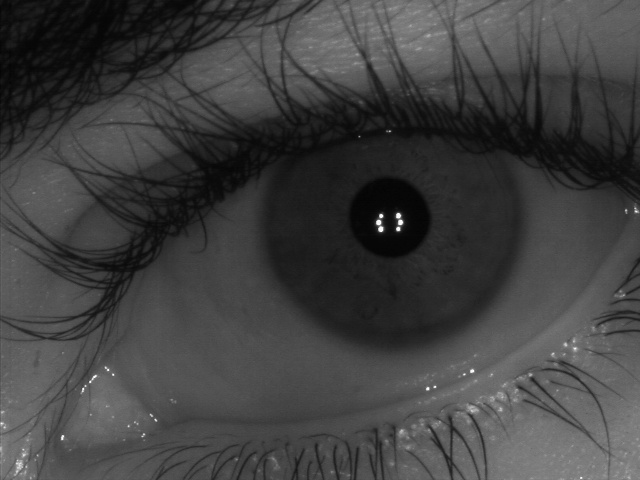}
\includegraphics[width=\imgsize\columnwidth]{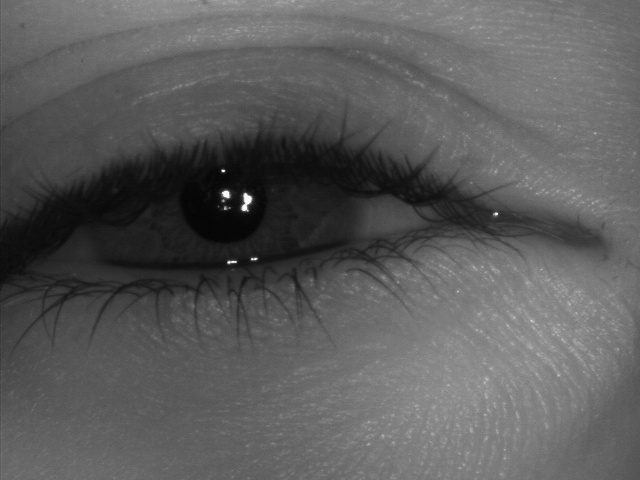}
\includegraphics[width=\imgsize\columnwidth]{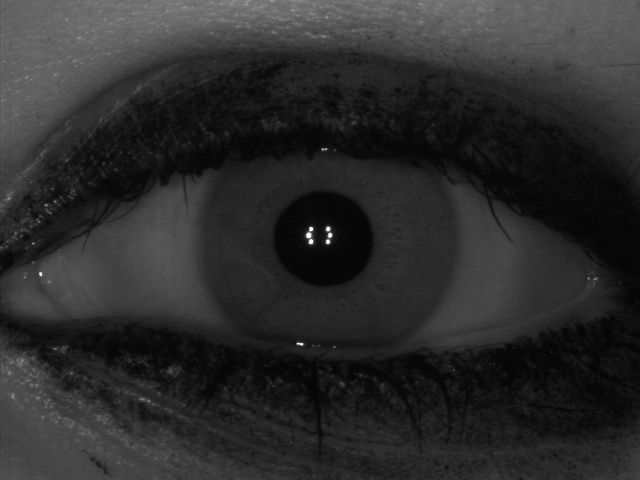}
\includegraphics[width=\imgsize\columnwidth]{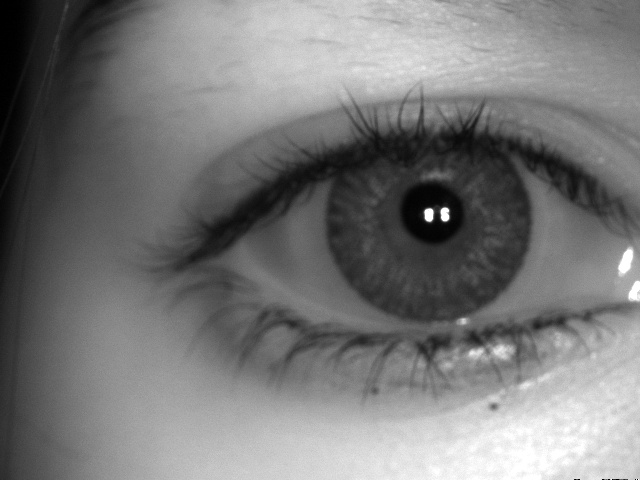}

\vspace{1mm}

\includegraphics[width=\imgsize\columnwidth]{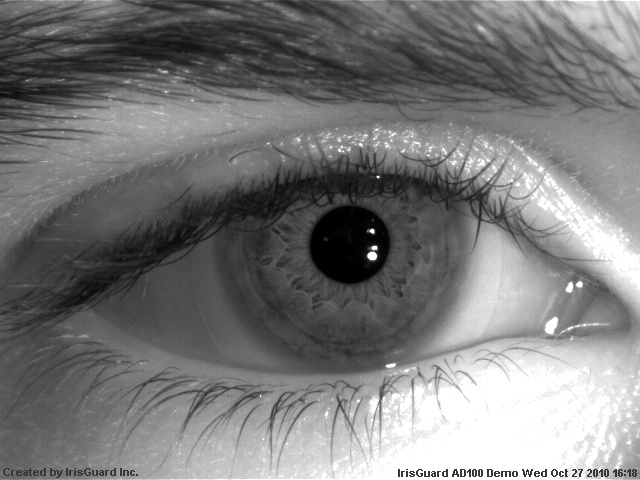}
\includegraphics[width=\imgsize\columnwidth]{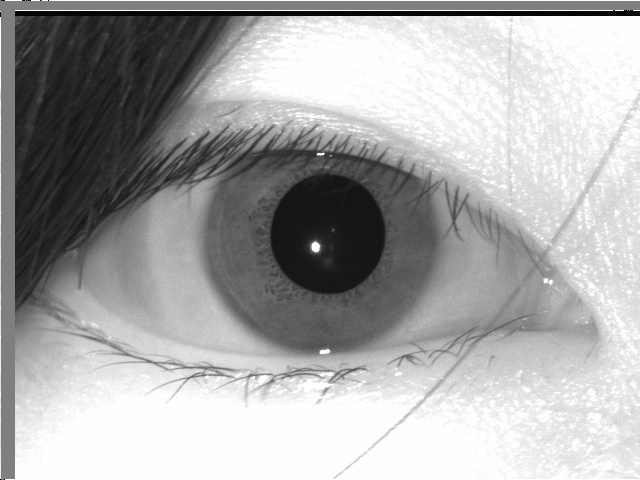}
\includegraphics[width=\imgsize\columnwidth]{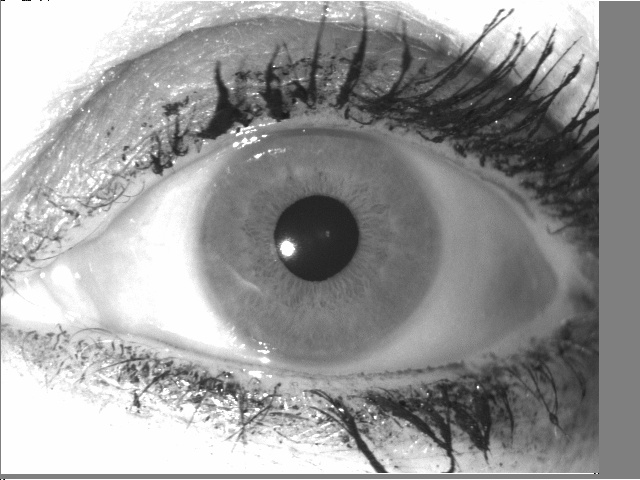}
\includegraphics[width=\imgsize\columnwidth]{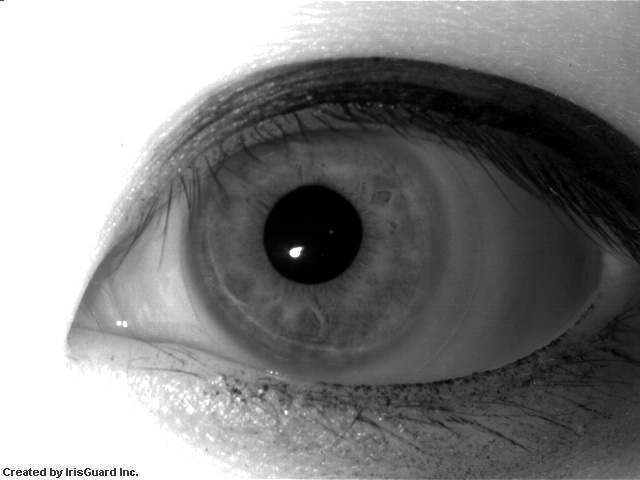}
\includegraphics[width=\imgsize\columnwidth]{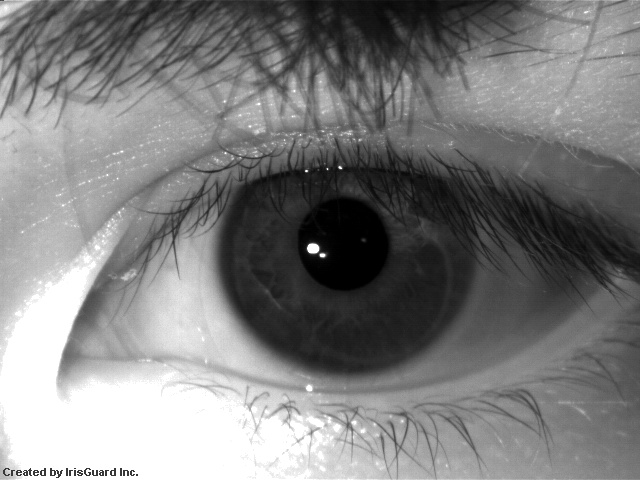}

\caption{From top to bottom:~\gls{nir} ocular image samples from the \casiaLamp~\cite{CASIA2010}, \casiaInterval~\cite{CASIA2010}, \ndCLD~\cite{Doyle2015}, \iiitdCLI~\cite{Kohli2013, Yadav2014} and \ndCCL~\cite{NDLENS2013, Doyle2013} databases.}
\label{fig:nirsamples}
\end{figure}

Ocular images captured at~\gls{nir} wavelength are generally used to study the features present in the iris~\cite{CASIA2010, Phillips2008, Phillips2010}.
As even darker pigmentation irises reveal rich and complex features~\cite{Daugman2004}, most of the visible light is absorbed by the melanin pigment while longer wavelengths of light are reflected~\cite{Bowyer2008}.
Other studies can also be performed with this kind of databases, such as methodologies to create synthetic irises~\cite{Shah2006, Zuo2007}, vulnerabilities in iris recognition and liveness detection~\cite{Ruiz-Albacete2008, Czajka2013, Gupta2014, Kohli2016},
impact of contact lenses in iris recognition~\cite{Baker2010, Kohli2013, Doyle2013, Doyle2015},
template aging~\cite{Fenker2012, Baker2013},
influence of alcohol consumption~\cite{Arora2012} and
study of gender recognition through the iris~\cite{Tapia2016}.
The databases used for these and other studies are described in Table~\ref{tab:niririsdata} and detailed in this session.
Some samples of ocular images from~\gls{nir} databases are shown in Figure~\ref{fig:nirsamples}.

\begin{table*}[!ht]
\setlength{\tabcolsep}{8pt}
\scriptsize
\centering
\caption{NIR ocular databases. Modalities: Iris [IR] and Periocular [PR]. }
\label{tab:niririsdata}
\vspace{0.5mm}
\resizebox{\linewidth}{!}{
\begin{tabular}{lcccccc}
\toprule
Database                            & Year    & \begin{tabular}[c]{@{}c@{}}Controlled\\ Environment\end{tabular}  & Cross-sensor & Subjects & Images & Modality \\
\midrule
\casiavOne~\cite{CASIA2010}         & $2002$  & Yes    & No           & *$108$ eyes        & $756$        & [IR] \\
\casiavTwo \cite{CASIA2010}         & $2004$  & Yes    & Yes          & *$120$ classes     & $2{,}400$    & [IR] \\
\ndIris \cite{Phillips2010}         & $2005$  & Yes    & No           & $356$              & $64{,}980$   & [IR] \\
\iceOne \cite{Phillips2008}         & $2005$  & Yes    & No           & $132$              & $2{,}953$    & [IR] \\
\iceTwo \cite{Phillips2010}         & $2006$  & No     & No           & $240$              & $59{,}558$   & [IR] \\
\wvuSIT \cite{Shah2006}             & $2006$  & N/A    & N/A          & *$1{,}000$ classes & $7{,}000$    & [IR] \\
\wvuSIM \cite{Zuo2007}              & $2007$  & N/A    & N/A          & $5{,}000$          & $160{,}000$  & [IR] \\
\fakeIris \cite{Ruiz-Albacete2008}  & $2008$  & N/A    & No           & $50$               & $800$        & [IR] \\
\casiaInterval \cite{CASIA2010}     & $2010$  & Yes    & No           & $249$              & $2{,}639$    & [IR] \\
\casiaLamp \cite{CASIA2010}         & $2010$  & Yes    & No           & $411$              & $16{,}212$   & [IR] \\
\casiaTwins \cite{CASIA2010}        & $2010$  & Yes    & No           & $200$              & $3{,}183$    & [IR] \\
\casiaThousand \cite{CASIA2010}     & $2010$  & Yes    & No           & $1{,}000$          & $20{,}000$   & [IR] \\
\casiaSyn \cite{CASIA2010}          & $2010$  & N/A    & N/A          & *$1{,}000$ classes & $10{,}000$   & [IR] \\
\iitdIris \cite{Kumar2010}          & $2010$  & Yes    & No           & $224$              & $1{,}120$    & [IR] \\
\ndICL \cite{Baker2010}             & $2010$  & Yes    & No           & $124$              & $21{,}700$   & [IR] \\
\ndITA \cite{Fenker2012}            & $2012$  & Yes    & No           & $322$              & $22{,}156$   & [IR] \\
\ndTLI \cite{Baker2013}             & $2012$  & Yes    & No           & $23$               & $6{,}797$    & [IR] \\
\iiitdIUAI \cite{Arora2012}         & $2012$  & Yes    & No           & $55$               & $440$        & [IR] \\
\iiitdCLI \cite{Kohli2013}          & $2013$  & Yes    & Yes          & $101$              & $6{,}570$    & [IR] \\
\ndCCL \cite{NDLENS2013, Doyle2013} & $2013$  & Yes    & Yes          & N/A                & $5{,}100$    & [IR] \\
\ndcsIris \cite{ndcsiris2013}       & $2013$  & Yes    & Yes          & $676$              & $146{,}550$  & [IR] \\
\irisPrint \cite{Czajka2013}        & $2013$  & Yes    & No           & *$243$ eyes        & $1{,}976$    & [IR] \\
\iiitdIS \cite{Gupta2014}           & $2014$  & Yes    & Yes          & $101$              & $4{,}848$    & [IR] \\
\ndCLD \cite{Doyle2015}             & $2015$  & Yes    & Yes          & N/A                & $7{,}300$    & [IR] \\
\iiitdCS \cite{Kohli2016}           & $2016$  & N/A    & Yes          & $1{,}872$          & $20{,}693$   & [IR] \\
\ndGFI \cite{Tapia2016}             & $2016$  & Yes    & No           & $1{,}500$          & $3{,}000$    & [IR] \\
\berc \cite{Kim2016}                & $2016$  & No     & No           & $100$              & $500$        & [IR] \\
\csi \cite{raghavendra2016cataract} & $2016$  & Yes    & No           & $84$               & $504$        & [IR] \\
\ornl \cite{karakaya2016angle}      & $2016$  & Yes    & No           & $50$               & $1{,}100$    & [IR] \\
\muid \cite{kurtuncu2016offangle}   & $2016$  & Yes    & No           & $111$              & $24{,}360$   & [IR] \\
\casiaMob \cite{Zhang2018fusion}    & $2018$  & Yes    & Yes          & $630$              & $11{,}000$   & [IR]/[PR] \\
\openEDS \cite{Garbin2019}          & $2019$  & Yes    & No           & $152$              & $356{,}649$  & [IR] \\
\bottomrule

\end{tabular}
}
\end{table*}

One of the first iris databases found in the literature was created and made available by CASIA (Chinese Academy of Science). The first version, called~\casiavOne, was made available in 2002.
The~\casiavOne database has $756$ images of $108$ eyes with a size of $320\times280$ pixels.
The \gls{nir} images were captured in two sections with a homemade iris camera~\cite{CASIA2010}.
In a second version (\casiavTwo), made available in $2004$, the authors included two subsets captured by an OKI IRISPASS-h and CASIA-IrisCamV2 sensors.
Each subset has $1{,}200$ images belonging to $60$ classes with a resolution of $640\times480$ pixels~\cite{CASIA2010}.
The third version of the database (CASIA-IrisV3), made available in $2010$, has a total of $22{,}034$ images from more than $700$ individuals, arranged among its three subsets: CASIA-Iris-Interval, CASIA-Iris-Lamp and CASIA-Iris-Twins.
Finally, CASIA-IrisV4, an extension of CASIA-IrisV3 and also made available in $2010$, is composed of six subsets: three from the previous version and three new ones: CASIA-Iris-Distance, CASIA-Iris-Thousand and CASIA-Iris-Syn. 
All six subsets together contain $54{,}601$ ocular images belonging to more than $1{,}800$ real subjects and $1{,}000$ synthetic ones.
Each subset will be detailed below, according to the specifications described in~\cite{CASIA2010}.

The CASIA-Iris-Interval database has images captured under a near-infrared LED illumination.
In this way, these images are used to study the texture information contained in the iris traits.
The database is composed of $2{,}639$ images, obtained in two sections, from $249$ subjects and $395$ classes with a resolution of $320\times280$ pixels.

The images from the CASIA-Iris-Lamp database were acquire by a non-fixed sensor (OKI IRISPASS-h) and thus the individual collected the iris image with the sensor in their own hands.
During the acquisition, a lamp was switched on and off to produce more intra-class variations due to contraction and expansion of the pupil, creating a non-linear deformation.
Therefore, this database can be used to study problems such as iris normalization and robust iris feature representation.
A total of $16{,}212$ images, from $411$ subjects, with a resolution of $640\times480$ pixels were collected in a single~section.

During an annual twin festival in Beijing, iris images from $100$ pairs of twins were collected to form the CASIA-Iris-Twins database, enabling the study of similarity between iris patterns of twins.
This database contains $3{,}183$ images ($400$ classes from $200$ subjects) captured in a single section with the OKI~IRISPASS-h~camera at a resolution of $640\times480$ pixels.

The CASIA-Iris-Thousand database is composed of $20{,}000$ ocular images from $1{,}000$ subjects, with a resolution of $640\times480$ pixels, collected in a single section by an IKEMB-100 IrisKing camera~\cite{IrisKing}.
Due to a large number of subjects, this database can be used to study the uniqueness of iris features.
The main source of intra-class variations that occur in this database is due to specular reflections and~eyeglasses.

The last subset of CASIA-IrisV4, called CASIA-IRIS-Syn, is composed of iris images generated with iris textures automatically synthesized from the ~\casiavOne subset.
The generation process applied the segmentation approach proposed by Tan et al.~\cite{Tan2010}.
Factors such as blurring, deformation, and rotation were introduced to create some intra-class variations.
In total, this database has $10{,}000$ images belonging to $1{,}000$~classes.

The images from the \ndIris~\cite{Phillips2010} database were captured with the LG2200 imaging system using \gls{nir} illumination.
The database contains $64{,}980$ images from $356$ subjects and there are several images with subjects wearing contact lenses.
Even the images being captured under a controlled environment, some conditions such as blur, occlusion of part of the iris region, and problems like off-angle may occur.
The \ndIris is a superset of the databases used in the \iceOne~\cite{Phillips2008} and \iceTwo~\cite{Phillips2010} competitions.

The \iceOne database was created for the Iris Challenge Evaluation 2005 competition~\cite{Phillips2008}.
This database contains a total of $2{,}953$ iris images from $132$ subjects.
The images were captured under \gls{nir} illumination using a complete LG EOU 2200 acquisition system with a resolution of $640\times480$ pixels.
Images that did not pass through the automatic quality control of the acquisition system were also added to the database.
Experiments were performed independently for the left and right eyes.
The results of the competition can be seen in~\cite{Phillips2008}.

The \iceTwo database has images collected using the LG EOU 2200 acquisition system with a resolution of $640\times480$ pixels.
For each subject, two `shots' of $3$ images of each eye were performed per session, totaling $12$ images.
The imaging sessions were held in three academic semesters between 2004 and 2005.
The database has a total of $59{,}558$ iris images from $240$ subjects~\cite{Phillips2010}.

The \wvuSIT database, created at West Virginia University, has $1{,}000$ classes with $7$ grayscale images each.
It consists exclusively of synthetic data, with the irises being generated in two phases.
First, a Markov Random Field model was used to generate the overall iris appearance texture.
Then, a variety of features were generated (e.g., radial and concentric furrows, crypts and collarette) and incorporated into the iris texture.
This database was created to evaluate iris recognition algorithms since, at the time of publication, there were few available iris databases and they had a small number of individuals~\cite{Shah2006}.

The \wvuSIM database also consists of synthetically generated iris images.
This database contains $10{,}000$ classes from $5{,}000$ individuals, with degenerated images by a combination of several effects such as specular reflection, noise, blur, rotation, and low contrast.
The image gallery was created in five steps using a model and anatomy-based approach~\cite{Zuo2007}, which contains $40$ randomized and controlled parameters.
The evaluation of their synthetic iris generation methodology was performed using a traditional Gabor filter-based iris recognition system.
This database provides a large amount of data that can be used to evaluate ocular biometric systems.

The \fakeIris was created using images from $50$ subjects belonging to the \biosec baseline database~\cite{Fierrez2007} and has $800$ fake iris images~\cite{Ruiz-Albacete2008}.
The process for creating new images is divided into three steps.
The original images were first reprocessed to improve quality using techniques such as noise filtering, histogram equalization, opening/closing, and top hat.
Then, the images were printed on paper using two commercial printers: an HP Deskjet 970cxi and an HP LaserJet 4200L, with six distinct types of papers: white paper, recycled paper, photographic paper, high-resolution paper, butter paper, and cardboard~\cite{Ruiz-Albacete2008}.
Finally, the printed images were recaptured by an LG IrisAccess EOU3000~camera.

The \iitdIris database consists of $1{,}120$ images, with a resolution of $320\times240$ pixels, from $224$ subjects captured with the JIRIS JPC1000 digital CMOS camera.
This database was created to provide a large-scale database of iris images of Indian users.
In~\cite{Kumar2010}, Kumar and Passi employed these images to compare the performance of different approaches for iris identification (e.g., Discrete Cosine Transform, Fast Fourier Transform, Haar wavelet, and Log-Gabor filter) and to investigate the impact in recognition performance using a score-level~combination.

The images from the \ndICL database were captured using the LG 2200 iris imaging system.
Visual inspections were performed to reject low-quality images or those with poor results in segmentation and matching.
To compose the database, the authors captured $9{,}697$ images from $124$ subjects that were not wearing contact lenses and $12{,}003$ images from $87$ subjects that were wearing contact lenses.
More specifically, the images were acquired from $92$ subjects not wearing lenses, $52$ subjects wearing the same lens type in all acquisitions, $32$ subjects who wore lenses only in some acquisitions and $3$ subjects that changed the lens type between acquisitions~\cite{Baker2010}.
According to Baker et al.~\cite{Baker2010}, the purpose of this database is to verify the degradation of iris recognition performance due to non-cosmetic prescription contact lenses.

The \ndITA database, described and used by Fenker and Bowyer~\cite{Fenker2012}, was created to analyze the template aging in iris biometrics.
The images were collected from 2008 to 2011 using an LG 4000 sensor, which captures images at \gls{nir}.
This database has $22{,}156$ images, being $2{,}312$ from 2008, $5{,}859$ from 2009, $6{,}215$ from 2010 and $7{,}770$ from 2011, corresponding to $644$ irises from $322$ subjects.
The ND-Iris-Template-Aging-2008-2010 subset belongs to this database.

All images from the \ndTLI database~\cite{Baker2013} were taken with the LG 2200 iris imaging system, without hardware or software modifications throughout $4$ years.
Imaging sessions were held at each academic semester over $4$ years, with $6$ images of each eye being captured per individual in each session.
From 2004 to 2008, a total of $6{,}797$ images were obtained from $23$ subjects who were not wearing eyeglasses, $5$ subjects who were wearing contact lenses, and $18$ subjects who were not wearing eyeglasses or contact lenses in any session.
This database was created to investigate template aging in iris~biometrics.

To investigate the effect of alcohol consumption on iris recognition, Arora et al.~\cite{Arora2012} created the Iris Under Alcohol Influence (\iiitdIUAI) database, which contains $440$ images from $55$ subjects, with $220$ images
being acquired before alcohol consumption and $220$ after it.
The subjects consumed approximately $200$~ml of alcohol in approximately $15$ minutes, and the second half of the images were taken between $15$ and $20$ minutes after consumption.
Due to alcohol consumption, there is a deformation in iris patterns caused by the dilation of the pupil, affecting iris recognition performance~\cite{Arora2012}.
The images were captured using the Vista IRIS scanner at \gls{nir} wavelength.

The IIITD Contact Lens Iris (\iiitdCLI) database is composed of $6{,}570$ iris images belonging to $101$ subjects.
The images were captured by two different sensors: Cogent CIS~202 dual iris sensor and VistaFA2E single iris sensor with each subject (i)~not wearing contact lenses, (ii)~wearing color cosmetic lenses, and (iii)~wearing transparent lenses.
Four lens colors were used: blue, gray, hazel and green.
At least $5$ images of each iris were collected in each lens category for each sensor~\cite{Kohli2013}.

The images from the \ndCCL database~\cite{NDLENS2013} were captured by two iris cameras, an LG4000 and an IrisGuard AD100, in a controlled environment under \gls{nir} illumination with a resolution of  $640\times480$ pixels.
These images are divided into four classes, (i)~no contact lenses, (ii)~soft, (iii)~non-textured and (iv)~textured contact lenses.
Also, this database is organized into two subsets: Subset1 (LG4000) and Subset2 (AD100).
Subset1 has $3{,}000$ images in the training set and $1{,}200$ images in the validation set.
Subset2 contains $600$ and $300$ images for training and validation, respectively~\cite{Doyle2013, Yadav2014, severo2018benchmark}.
Both subsets have $10$ equal folds of training images for testing purposes.

The \ndcsIris database~\cite{ndcsiris2013} is composed of $146{,}550$~\gls{nir} images belonging to $676$ unique subjects, being $29{,}986$ images captured using an LG4000 and $116{,}564$ taken by an LG2200 iris sensor with $640\times480$ pixels of resolution.
The images were captured in $27$ sessions over three years, from 2008 to 2010, and in at least two sessions there are images of the same subject.
The purpose of this database is to investigate the effect of cross-sensor images on iris recognition.
Initially, this database was released for a competition to be held at the BTAS 2013 Conference, but the competition did not have enough submission.

The \irisPrint was created for liveness detection in iris images and contains $729$ printout images of $243$ eyes, and $1{,}274$ images of imitations from genuine eyes.
The database was constructed as follows.
First, the iris images were obtained with an IrisGuard AD100 camera. Then, they were printed using the HP LaserJet 1320 and Lexmark c534dn printers.
To check the print quality, the printed images were captured by the Panasonic ET-100 camera using an iris recognition software, and the images that were successfully recognized were recaptured by an AD100 camera with a resolution of $640\times480$ pixels to create the imitation subset.
Initially, images from $426$ distinct eyes belonging to $237$ subjects were collected.
After the process of recognizing the printed images, $243$ eyes images (which compose the database) were successfully~verified~\cite{Czajka2013}.

The \iiitdIS (IIS) database was created to study spoofing methods.
To this end, printed images from the \iiitdCLI~\cite{Kohli2013} database were used.
Spoofing was simulated in two ways.
In the first, the printed images were captured by a specific iris scanner (Cogent CIS 202 dual eye), while in the second, the printed images were scanned using an HP flatbed optical scanner.
The database contains $4{,}848$ images from $101$ individuals~\cite{Gupta2014}.

The Notre Dame Contact Lenses 2015 (\ndCLD) database contains $7{,}300$ iris images.
The images were obtained under consistent lighting conditions by an LG4000 and an IrisGuard AD100 sensor.
All images have $640\times480$ pixels of resolution and are divided into three classes based on the lens type: no lens, soft, and textured.
This database was created to investigate methods to classify iris images based on types of contact lenses~\cite{Doyle2015}.

The \iiitdCS database was proposed to simulate a real-world scenario of attacks against iris recognition systems.
This database consists of joining the following databases: \iiitdCLI~\cite{Kohli2013}, IIITD IIS~\cite{Gupta2014}, SDB~\cite{Shah2006}, \iitdIris~\cite{Kumar2010} and, to represent genuine classes, iris images from $547$ subjects were collected.
The CSD database has a total of $1{,}872$ subjects, with $9{,}325$ normal image samples and $11{,}368$ samples of impostor images~\cite{Kohli2016}.

The Gender from Iris (\ndGFI) database was created to study the recognition of the subject's gender through the iris, specifically using the binary iris code (which is normally used in iris recognition systems)~\cite{Tapia2016}.
The images were obtained at \gls{nir} wavelength by an LG4000 sensor and labeled by gender.
The \ndGFI database contains a single image of each eye (left and right) from $750$ men and $750$ women, totaling $3{,}000$ images.
About a quarter of the images were captured with the subjects wearing clear contact lenses.
This database has another set of images that can be used for validation, called UND\_V, containing $1{,}944$ images, being $3$ images of each eye from $175$ men and $149$ women.
In this subset, there are also images using clear contact lenses and some cosmetics~\cite{Tapia2016}.

According to~\cite{ISO19794-6}, an iris image has good quality if the iris diameter is larger than $200$ pixels, and if the diameter is between $150$ and $200$ pixels, the image is classified as adequate quality. 
In this context, the images from the \berc have irises with a diameter between $170$ and $200$ pixels, obtained at \gls{nir} wavelength with $1280\times960$ pixels of resolution.
Using a mobile iris recognition system, the images were taken in sequences of $90$ shots~\cite{Kim2016} moving the device at three distances: $15$ to $25$~cm, $25$ to $15$~cm, and $40$ to $15$~cm.
In total, the database has $500$ images from $100$ subjects, which were the best ones selected by the authors of each sequence.

Raghavendra et al.~\cite{raghavendra2016cataract} created the \csi database to analyze the impact of cataract surgery on the verification performance of iris recognition systems.
The database contains $504$ images belonging to $84$ subjects who were affected by cataracts.
The subjects' ages vary from $50$ to $80$ years, being $34$ males and $49$ females.
Three eye samples of each subject were collected before ($24$ hours) and after ($36$~-~$42$ hours) the surgery to remove the cataractous lens.
The images were captured using a commercial dual-iris \gls{nir} device with a resolution of $640\times480$~pixels.

The Oak Ridge National Laboratory (\ornl) Off-angle database was created to study how the gaze angle affects the performance of iris biometrics~\cite{karakaya2013limbus, karakaya2016angle, karakaya2018angle}.
This database encompasses $1{,}100$~\gls{nir} iris images from $50$ subjects varying the angle acquisition from $-50\degree$ to $+50\degree$ with a step-size of $10\degree$.
The gender distribution consists of $56\%$ male and $44\%$ female subjects, and iris color of $64\%$ with dark colors and $36\%$ with light-colors. 
The images were collected by a Toshiba Teli CleverDragon series camera and have a resolution of $4096\times3072$~pixels.

The Meliksah University Iris Database (\muid) was collected to investigate the off-angle iris recognition.
The authors developed an iris image capture system composed of two cameras to simultaneously capture frontal and off-angle samples.
Thus, it is possible to isolate the effect of the gaze angle from pupil dilation and accommodation~\cite{kurtuncu2016offangle}.
In total, the database has $24{,}360$~\gls{nir} images from $111$ subjects, $64$ males and $57$ females, with an average age of $26$ years.
The images were captured by two infrared-sensitive IDS-UI-3240ML-NIR cameras varying from $-50\degree$ to $+50\degree$ angles with a step-size of $10\degree$ and have a resolution of $1280\times1024$ pixels. 
More details about the iris image acquisition platform are described in~\cite{kurtuncu2016offangle}.

The \casiaMob database is composed of $11{,}000$~\gls{nir} images belonging to $630$ subjects, divided into three subsets: CASIA-Iris-M1-S1~\cite{Zhang2015fusionfaceiris}, CASIA-Iris-M1-S2~\cite{Zhang2016exploring} and a new one called CASIA-IRIS-M1-S3.
The images were captured simultaneously from the left and right eyes and stored in $8$ bits gray-level JPG files.
The CASIA-Iris-M1-S1 subset has $1{,}400$ images from $70$ subjects with a resolution of $1920\times1080$ pixels, acquired using a~\gls{nir} imaging module attached to a mobile phone.
The CASIA-Iris-M1-S2 subset has images captured using a similar device.
In total, this subset contains $6{,}000$ images from $200$ subjects with a resolution of $1968\times1024$ pixels, collected at three distances: $20$, $25$ and $30$ cm.
At last, the CASIA-Iris-M1-S3 subset is composed of $3{,}600$ images belonging to $360$ subjects with a resolution of $1920\times1920$ pixels, which were taken with a~\gls{nir} iris-scanning technology equipped on a mobile phone.

The Open Eye Dataset (\openEDS) was created to investigate the semantic segmentation of eyes components, and background~\cite{Garbin2019}.
This database is composed of $356{,}649$ eye images, being $12{,}759$ images with pixel-level annotations, $252{,}690$ unlabeled ones, and $91{,}200$ images from video sequences belonging from $152$ subjects.
The images were captured with a head-mounted display with two synchronized cameras under controlled~\gls{nir} illumination with a resolution of  $640\times400$ pixels.

\subsection{Visible and Cross-Spectral Ocular Images Databases}
\label{subsec:vis-databases}

\begin{figure}[!ht]
\centering
\includegraphics[width=\imgsize\columnwidth]{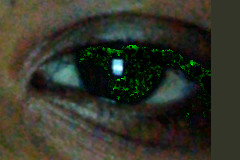}
\includegraphics[width=\imgsize\columnwidth]{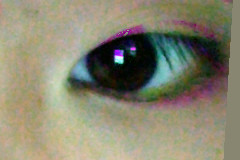}
\includegraphics[width=\imgsize\columnwidth]{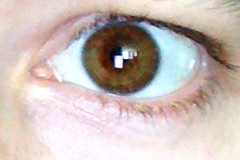}
\includegraphics[width=\imgsize\columnwidth]{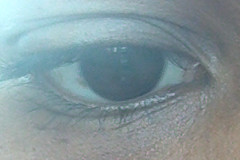}
\includegraphics[width=\imgsize\columnwidth]{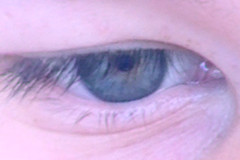}

\vspace{0.6mm}

\includegraphics[width=\imgsize\columnwidth]{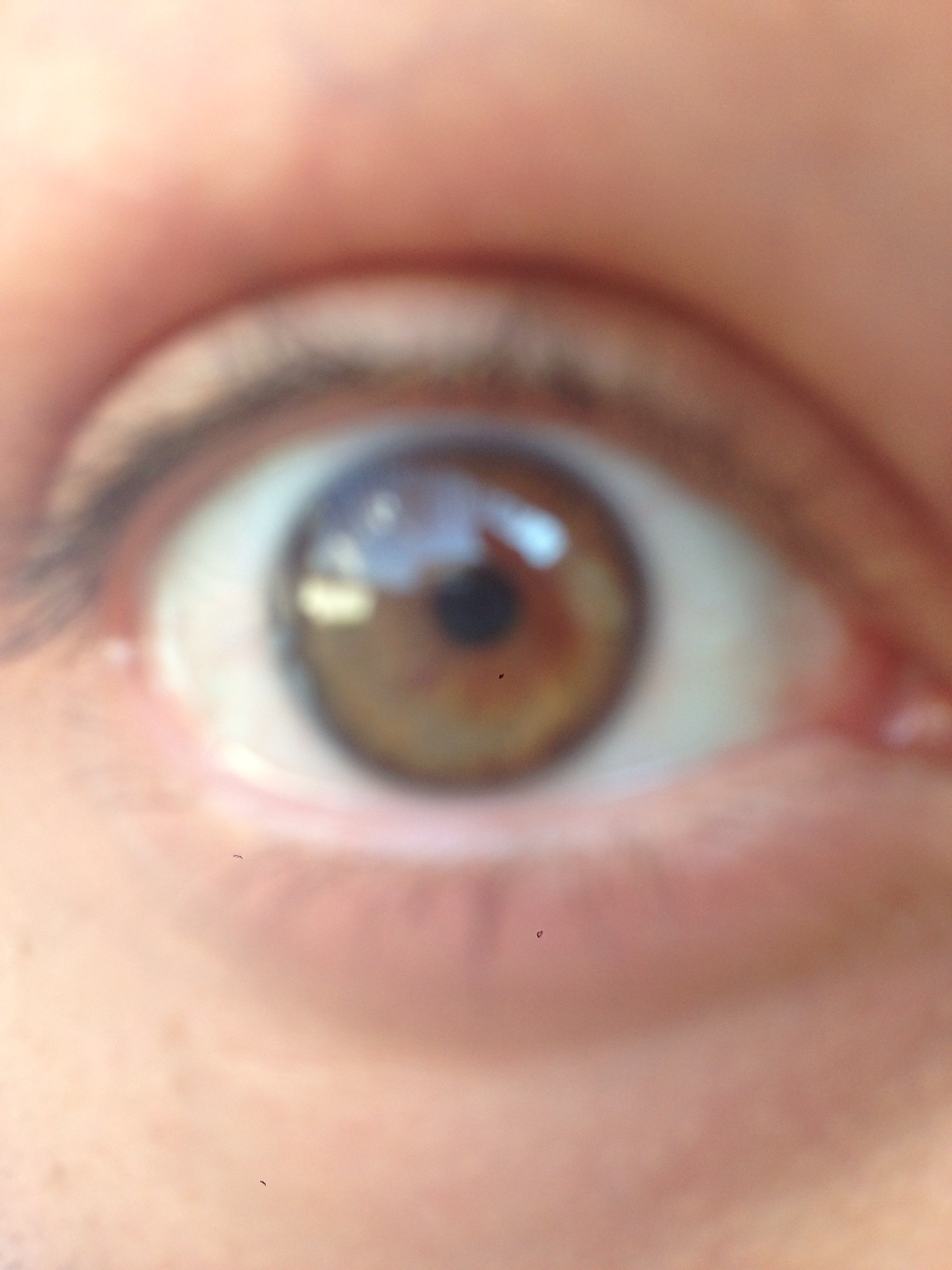}
\includegraphics[width=\imgsize\columnwidth]{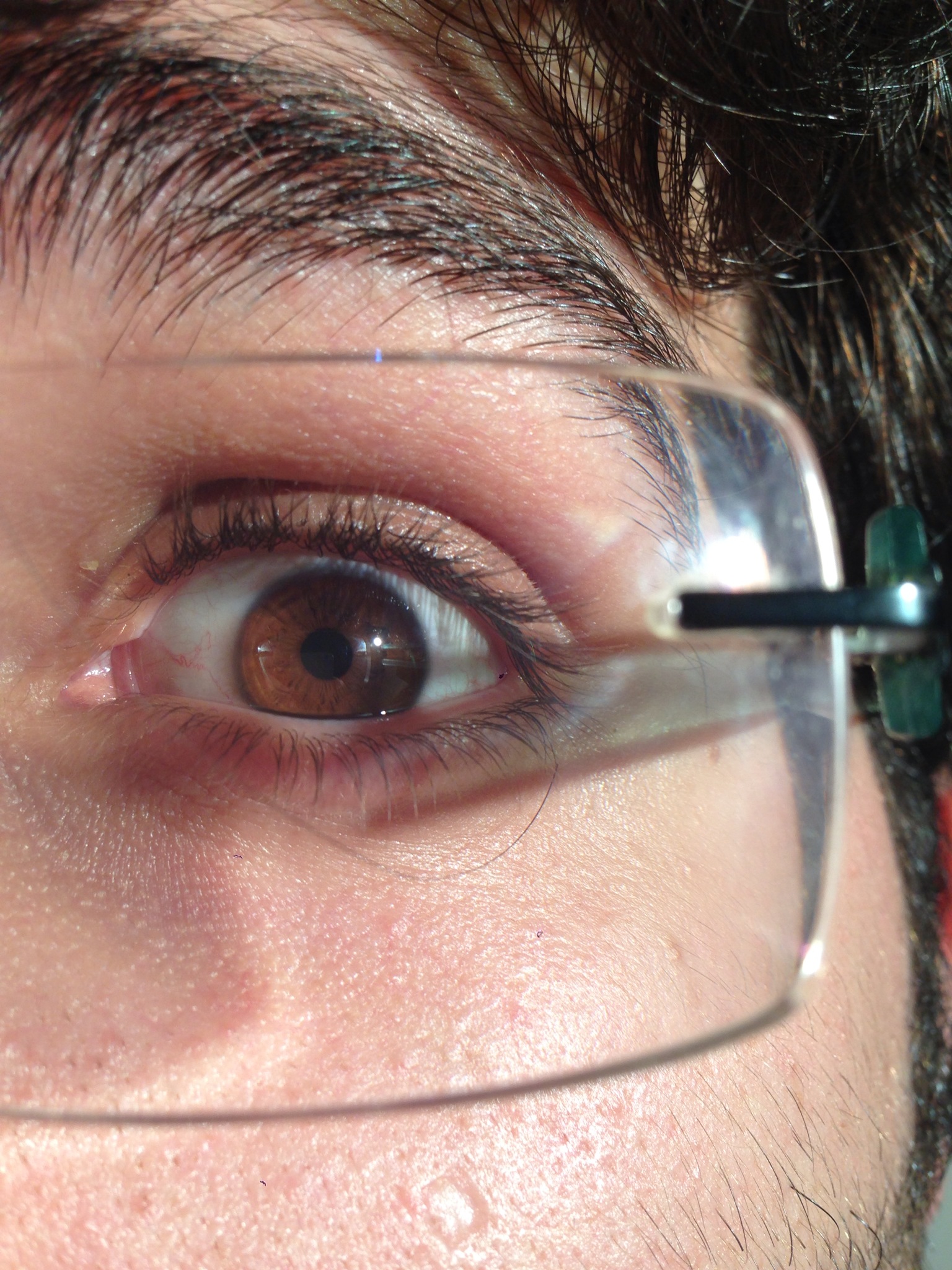}
\includegraphics[width=\imgsize\columnwidth]{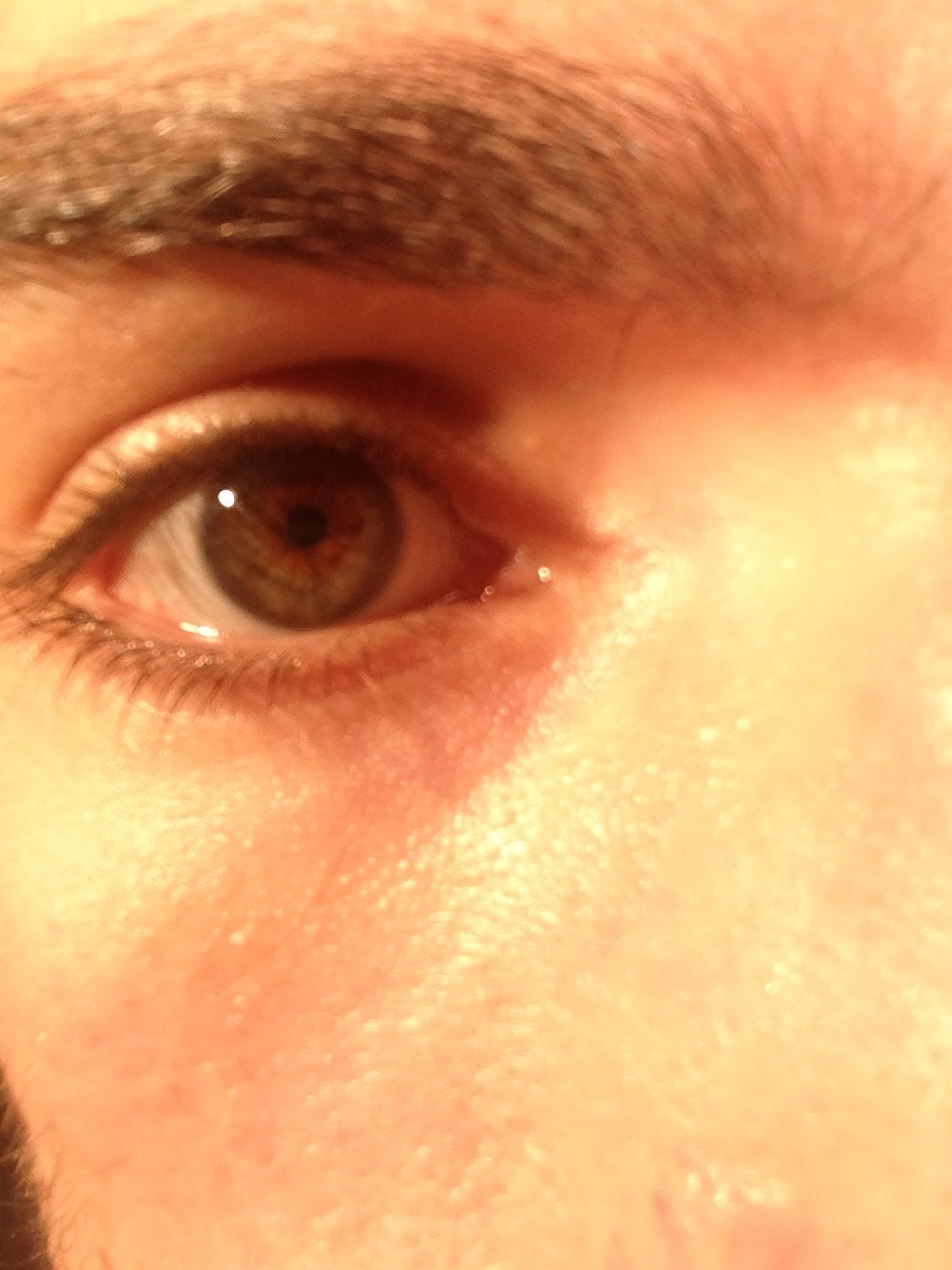}
\includegraphics[width=\imgsize\columnwidth]{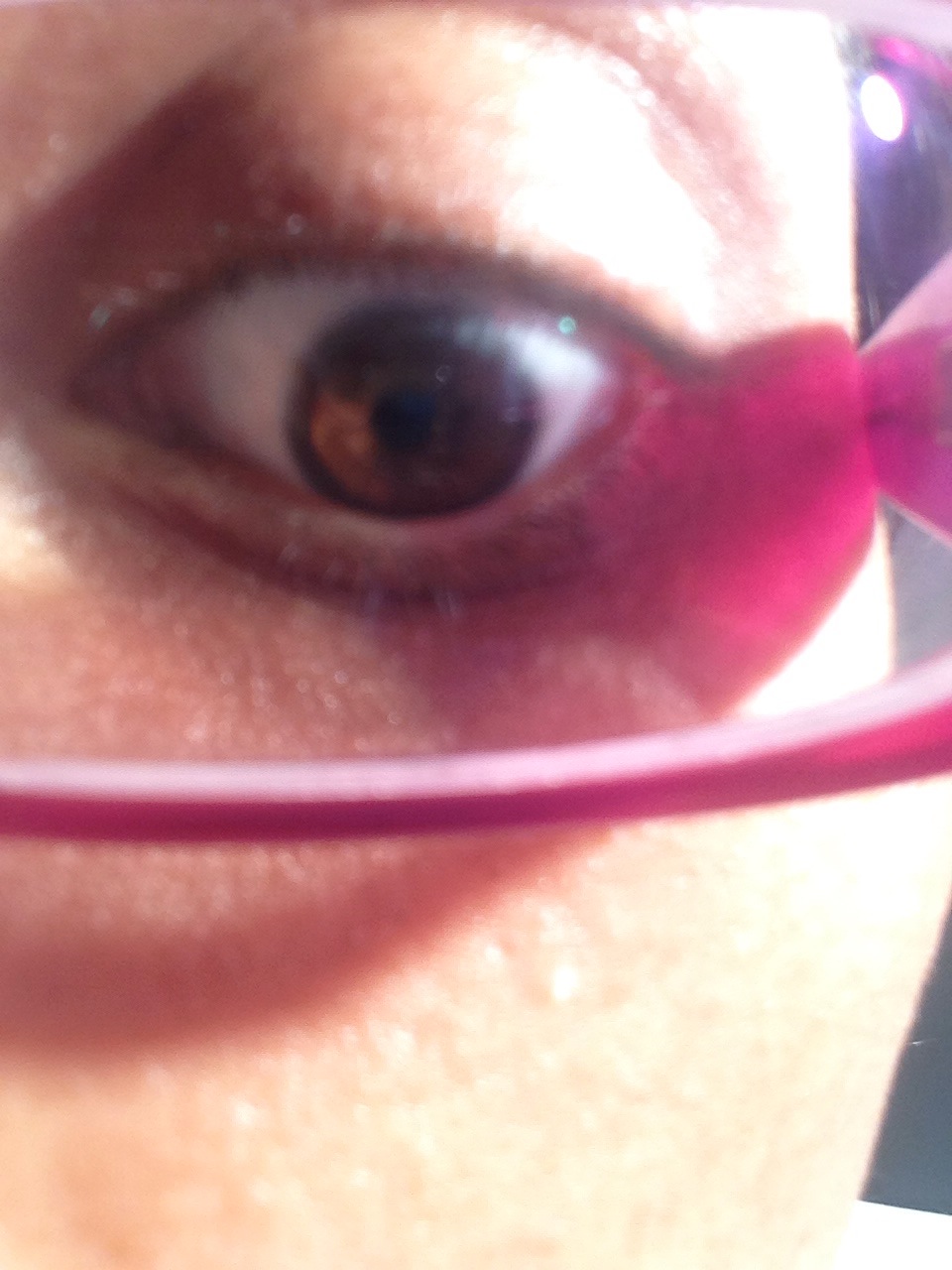}
\includegraphics[width=\imgsize\columnwidth]{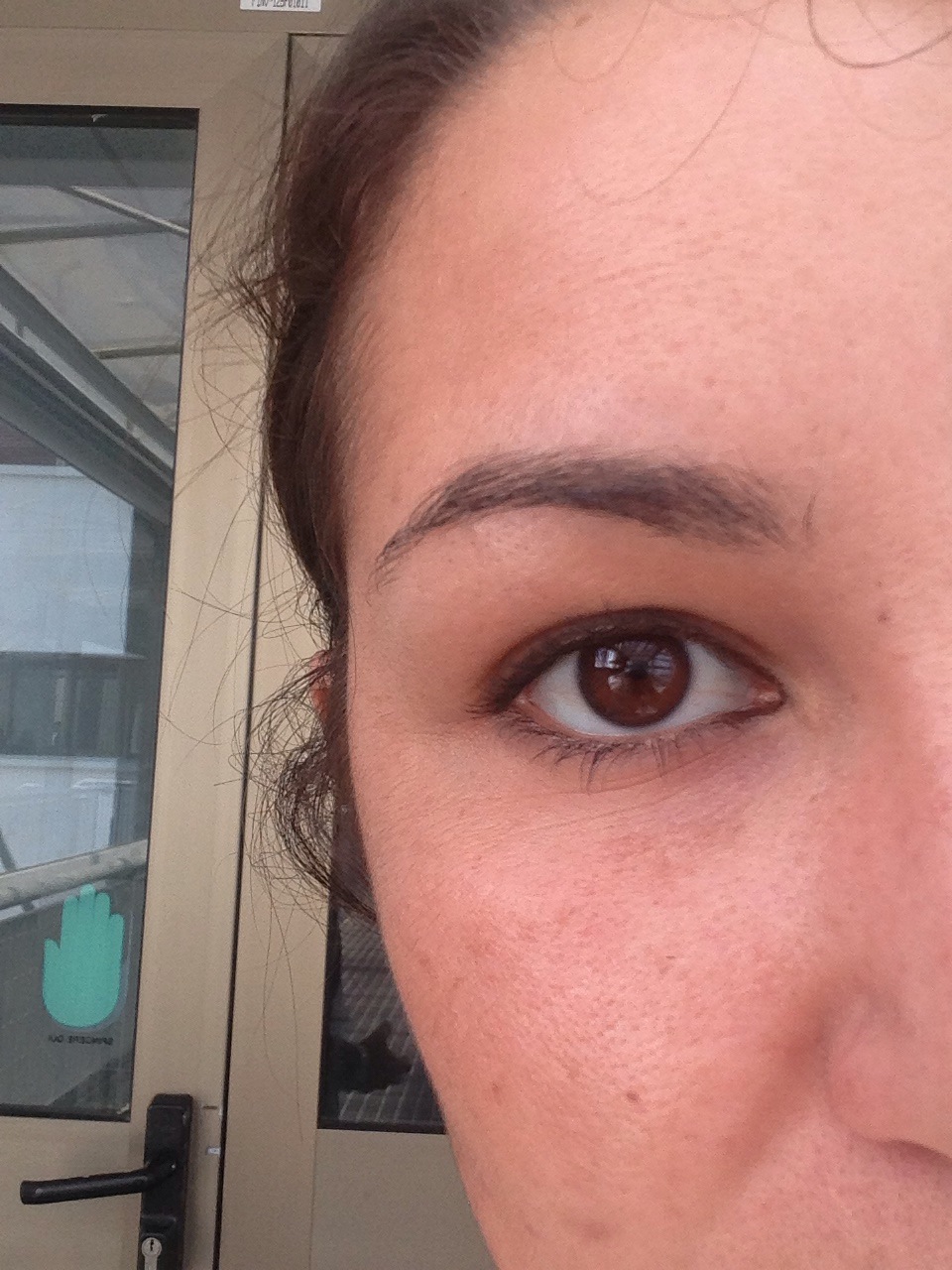}

\vspace{0.6mm}

\includegraphics[width=\imgsize\columnwidth]{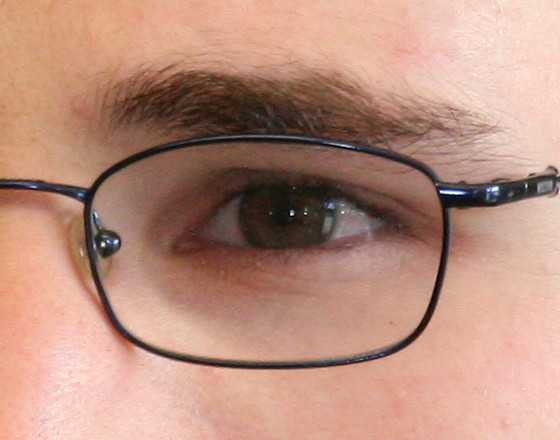}
\includegraphics[width=\imgsize\columnwidth]{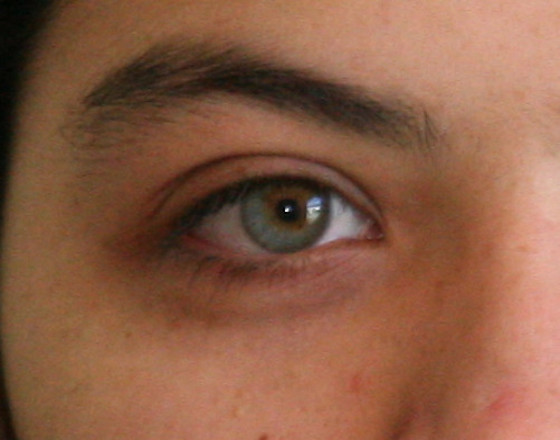}
\includegraphics[width=\imgsize\columnwidth]{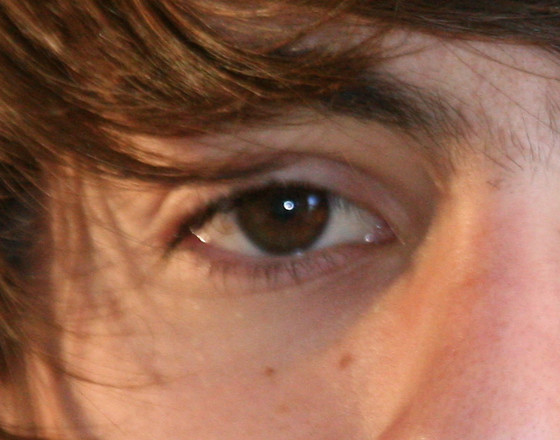}
\includegraphics[width=\imgsize\columnwidth]{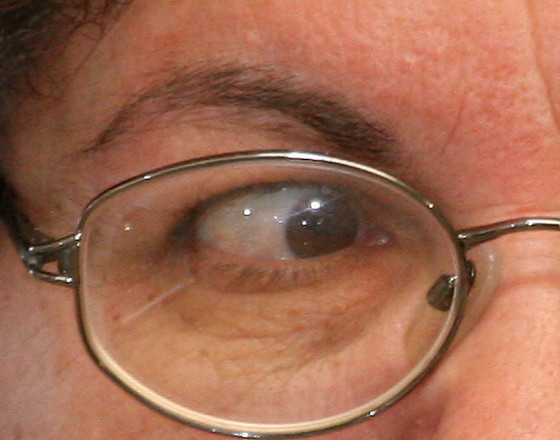}
\includegraphics[width=\imgsize\columnwidth]{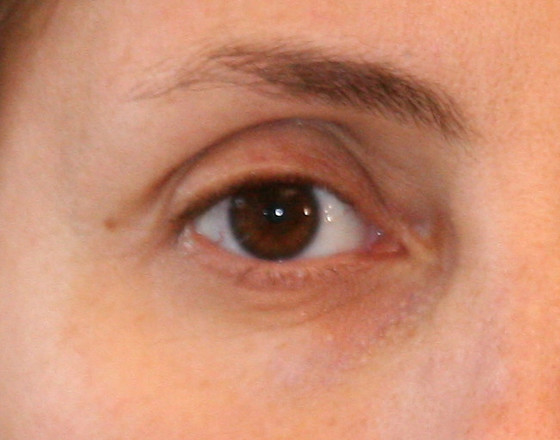}

\vspace{0.6mm}

\includegraphics[width=\imgsize\columnwidth]{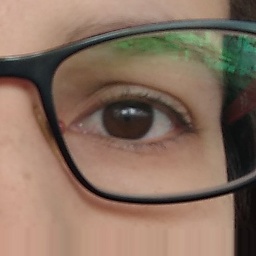}
\includegraphics[width=\imgsize\columnwidth]{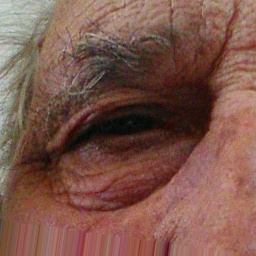}
\includegraphics[width=\imgsize\columnwidth]{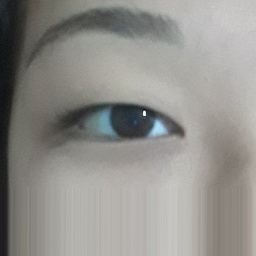}
\includegraphics[width=\imgsize\columnwidth]{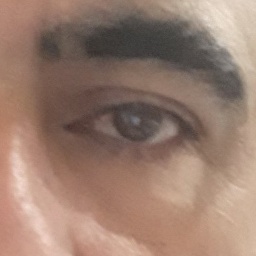}
\includegraphics[width=\imgsize\columnwidth]{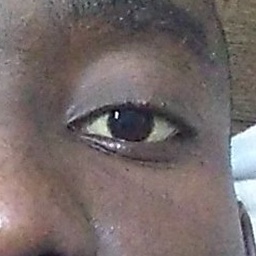}

\vspace{0.6mm}

\includegraphics[width=\imgsize\columnwidth]{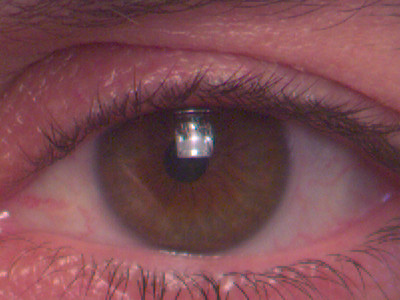}
\includegraphics[width=\imgsize\columnwidth]{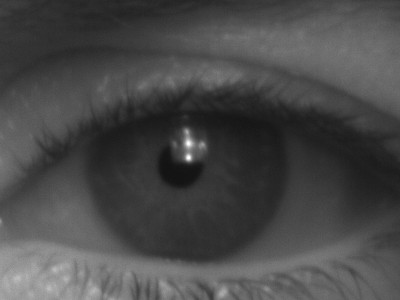}
\includegraphics[width=\imgsize\columnwidth]{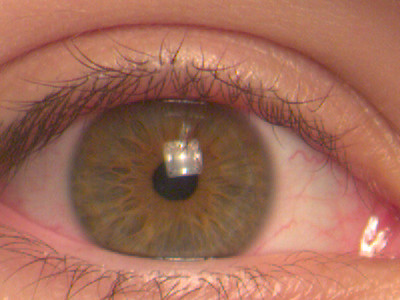}
\includegraphics[width=\imgsize\columnwidth]{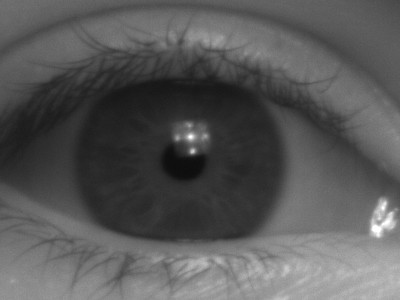}
\includegraphics[width=\imgsize\columnwidth]{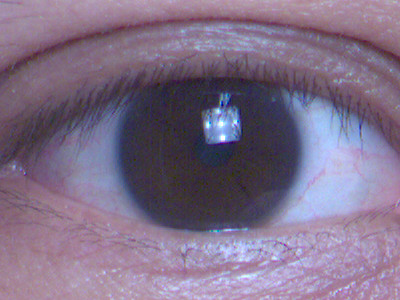}

\vspace{0.6mm}

\includegraphics[width=\imgsize\columnwidth]{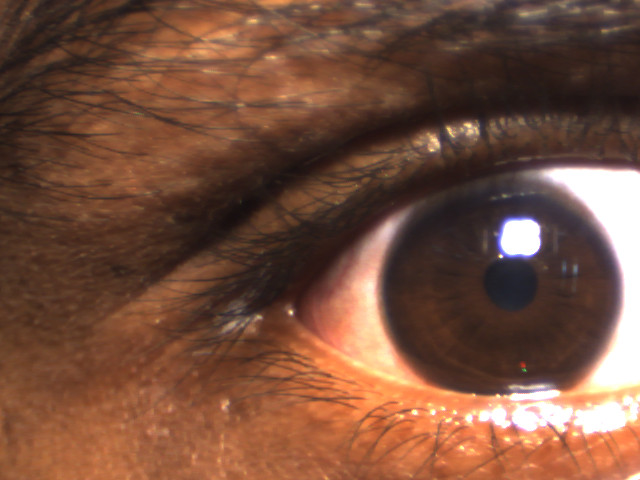}
\includegraphics[width=\imgsize\columnwidth]{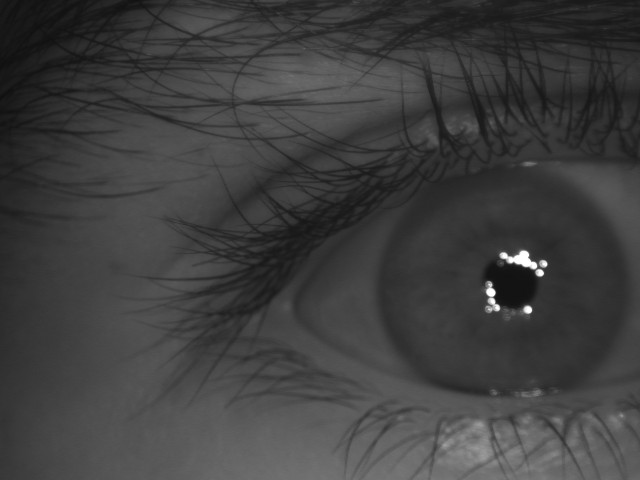}
\includegraphics[width=\imgsize\columnwidth]{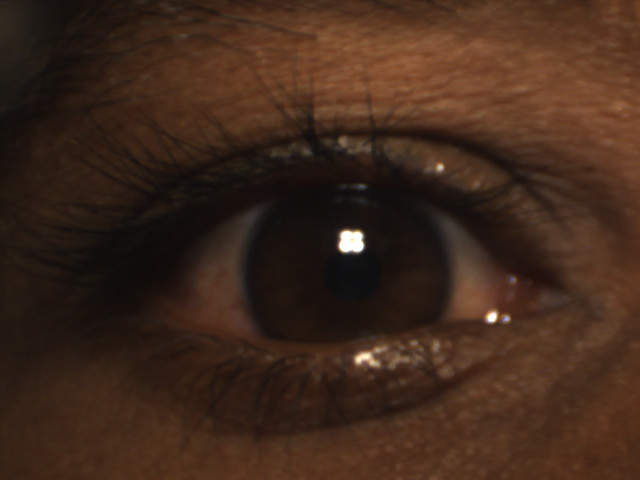}
\includegraphics[width=\imgsize\columnwidth]{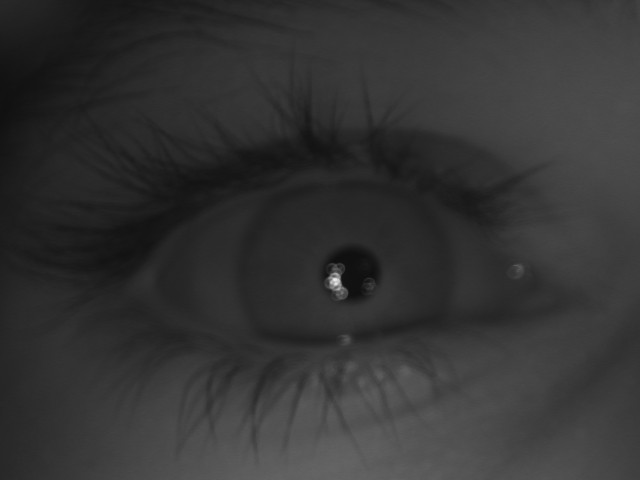}
\includegraphics[width=\imgsize\columnwidth]{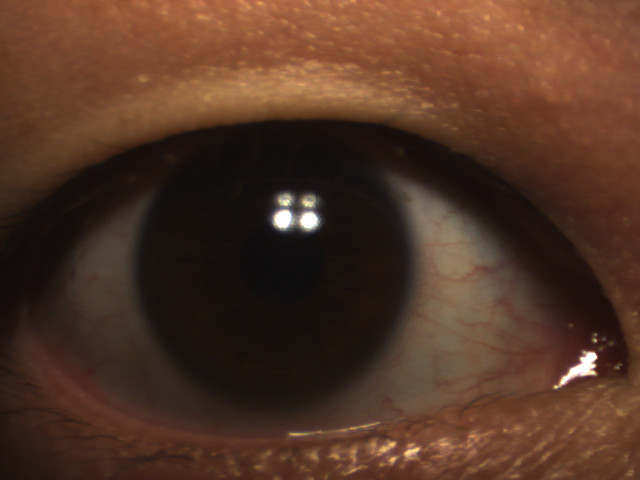}

\caption{From top to bottom:~\gls{vis} and Cross-spectral ocular image samples from the \visob~\cite{Rattani2016}, \miche~\cite{DeMarsico2015}, \ubipr~\cite{Padole2012}, \ufprPerioc~\cite{zanlorensi2020ufprperiocular}, \crossEyed~\cite{Sequeira2016, Sequeira2017}, \polyu~\cite{Nalla2017}~databases.}
\label{fig:vissamples}
\end{figure}

Iris recognition using images taken at controlled \gls{nir} wavelength environments is a mature technology, proving to be effective in different scenarios~\cite{Bowyer2008, Phillips2008, Phillips2010, Proenca2012, Proenca2017irina, Proenca2019segmentation}.
Databases captured under controlled environments have few or no noise factors in the images.
However, these conditions are not easy to achieve and require a high degree of collaboration from subjects.
In a more challenging/realistic scenario, investigations on biometric recognition employing iris images obtained in uncontrolled environments and at \gls{vis} wavelength have begun to be conducted~\cite{proenca2005, proenca2010}.
There is also research on biometric recognition using cross-spectral databases, i.e., databases with ocular images from the same individual obtained at both \gls{nir} and \gls{vis} wavelengths~\cite{Hosseini2010, Sharma2014, Nalla2017, Algashaam2017, Wang2019}.
Currently, many types of research have been performed on biometric recognition using iris and periocular region with images obtained from mobile devices, obtained in an uncontrolled environment and by different types of sensors~\cite{DeMarsico2015, Raja2015, Rattani2016}.
In this subsection, we describe databases with these characteristics. Table~\ref{tab:visirisdata} summarize these databases.
Some samples of ocular images from~\gls{vis} and Cross-spectral databases are shown in Figure~\ref{fig:vissamples}.

\begin{table*}[!ht]
\scriptsize
\setlength{\tabcolsep}{8pt}
\centering
\caption{Visible and Cross-spectral ocular databases. Wavelengths: Near-Infrared (NIR), Visible (VIS) and Night Vision (NV). Modalities: Iris [IR] and Periocular [PR].}
\label{tab:visirisdata}
\resizebox{\linewidth}{!}{
\begin{tabular}{lccccccc}
\toprule
Database                                     & Year    & \begin{tabular}[c]{@{}c@{}}Controlled\\ Environment\end{tabular}   & Wavelength & Cross-sensor & Subjects & Images & Modality \\
\midrule
\upol \cite{Dobes2004}                             & $2004$   & Yes  & VIS             & No           & $64$       & $384$       & [IR] \\
\ubirisvOne \cite{proenca2005}                     & $2005$   & No   & VIS             & No           & $241$      & $1{,}877$   & [IR] \\
\utiris \cite{Hosseini2010}                        & $2007$   & Yes  & VIS / NIR       & Yes          & $79$       & $1{,}540$   & [IR] \\
\ubirisvTwo \cite{proenca2010}                     & $2010$   & No   & VIS             & No           & $261$      & $11{,}102$  & [IR] \\
\ubipr \cite{Padole2012}                           & $2012$   & No   & VIS             & No           & $261$      & $10{,}950$  & [PR] \\
\bdcp \cite{Siena2012}                             & $2012$   & No   & VIS / NIR       & Yes          & $99$       & $4{,}314$   & [IR]/[PR] \\
\mobbioFake \cite{Sequeira2014b}                   & $2013$   & No   & VIS             & No           & N/A        & $1{,}600$   & [IR] \\
\iiitdMSP \cite{Sharma2014}                        & $2014$   & Yes  & VIS / NIR / NV  & Yes          & $62$       & $1{,}240$   & [PR] \\
\polyu \cite{Nalla2017}                            & $2015$   & N/A  & VIS / NIR       & Yes          & $209$      & $12{,}540$  & [IR] \\
\miche \cite{DeMarsico2015}                        & $2015$   & No   & VIS             & Yes (Mobile) & $92$       & $3{,}732$   & [IR] \\
\vssiris \cite{Raja2015}                           & $2015$   & No   & VIS             & Yes (Mobile) & $28$       & $560$       & [IR] \\
\csip \cite{Santos2015}                            & $2015$   & No   & VIS             & Yes (Mobile) & $50$       & $2{,}004$   & [IR]/[PR] \\
\visob \cite{Rattani2016}                          & $2016$   & No   & VIS             & Yes (Mobile) & $550$      & $158{,}136$ & [PR] \\
\crossEyed \cite{Sequeira2016, Sequeira2017}       & $2016$   & No   & VIS / NIR       & Yes          & $120$      & $3{,}840$   & [IR]/[PR] \\
\pmhi \cite{trokielewicz2016postmortem}            & $2016$   & Yes  & VIS / NIR       & Yes          & $6$        & $104$       & [IR] \\
\qutMP~\cite{Algashaam2017}                        & $2017$   & N/A  & VIS / NIR / NV  & Yes          & $53$       & $212$       & [PR] \\
\visobtwo~\cite{nguyen2020visob2}                  & $2020$   & No   & VIS             & Yes          & $150$      & $75{,}428$  & [PR] \\
\isocialdb \cite{labati2020socialdb}               & $2020$   & No   & VIS             & No           & $400$      & $3{,}286$   & [IR]/[PR] \\
\ufprPerioc~\cite{zanlorensi2020ufprperiocular}    & $2020$   & No   & VIS             & No           & $1{,}122$  & $33{,}660$  & [PR] \\
\ufprEyeglass~\cite{zanlorensi2020attnormalization}& $2020$   & No   & VIS             & No           & $83$       & $2{,}270$   & [PR] \\

\bottomrule

\end{tabular}
}
\end{table*}

The \upol (University of Palackeho and Olomouc) database has high-quality iris images obtained at \gls{vis} wavelength using the optometric framework (TOPCON TRC501A) and the Sony DXC-950P 3CCD camera.
In total, $384$ images of the left and right eyes were obtained from $64$ subjects at a distance of approximately $0.15$~cm with a resolution of $768\times576$ pixels, stored in $24$ bits~(RGB)~\cite{Dobes2004}.

The \ubirisvOne database~\cite{proenca2005} was created to provide images with different types of noise, simulating image capture with minimal collaboration from the users.
This database has $1{,}877$ images belonging to $241$ subjects, obtained in two sections by a Nikon E5700 camera.
For the first section (enrollment), some noise factors such as reflection, lighting, and contrast were minimized.
However, in the second section, natural lighting factors were introduced by changing the location to simulate an image capture with minimal or without active collaboration from the subjects.
The database is available in three formats: color with a resolution of $800\times600$ pixels, color with $200\times150$ pixels, and $200\times150$ pixels in grayscale~\cite{proenca2005}.

The \utiris is one of the first databases containing iris images captured at two different wavelengths (cross-spectral)~\cite{Hosseini2010}.
The database is composed of $1{,}540$ images of the left and right eyes from $79$ subjects, resulting in $158$ classes.
The \gls{vis} images were obtained by a Canon EOS 10D camera with $2048\times1360$ pixels of resolution.
To capture the \gls{nir} images, the ISW Lightwise LW camera was used, obtaining iris images with a resolution of $1000\times776$ pixels.
As the melanin pigment provides a rich source of features at the \gls{vis} spectrum, which is not available at \gls{nir}, this database can be used to investigate the impact of the fusion of iris image features extracted at both wavelengths.

The \ubirisvTwo database was built representing the most realistic noise factors.
For this reason, the images that constitute the database were obtained at \gls{vis} without restrictions such as distance, angles, light, and movement.
The main purpose of this database is to provide a tool for the research on the use of \gls{vis} images for iris recognition in an environment with adverse conditions.
This database contains images captured by a Canon EOS 5D camera, with a resolution of $400\times300$ pixels, in RGB from $261$ subjects containing $522$ irises and $11{,}102$ images taken in two sessions~\cite{proenca2010}.

The \ubipr (University of Beira Interior Periocular) database~\cite{Padole2012} was created to investigate periocular recognition using images taken under uncontrolled environments and setups.
The images from this database were captured by a Canon EOS 5D camera with a 400mm focal length.
Five different distances and resolutions were configured: $501\times401$ pixels (8m), $561\times541$ pixels (7m), $651\times501$ pixels (6m), $801\times651$ pixels (5m), and $1001\times801$ pixels (4m).
In total, the database has $10{,}950$ images from $261$ subjects (the images from $104$ subjects were obtained in $2$~sessions).
Several variability factors were introduced in the images, for example, different distances between the subject and the camera, as well as different illumination, poses and occlusions levels.

The \bdcp (Biometrics Development Challenge Problem) database~\cite{Siena2012} contains images from two different sensors: an LG4000 sensor that captures images in gray levels, and a Honeywell Combined Face and Iris Recognition System (CFAIRS) camera~\cite{Siena2012}, which captures \gls{vis} images.
The resolutions of the images are $640\times480$ pixels for the LG4000 sensor and $750\times600$ pixels for the CFAIRS camera.
To compose the database, $2{,}577$ images from $82$ subjects were acquired by the CFAIRS sensor and $1{,}737$ images belonging to $99$ subjects were taken by an LG4000 sensor.
Images of the same subject were obtained for both sensors~\cite{Smereka2015}.
The main objective of this database is the cross-sensor evaluation, matching \gls{nir} against \gls{vis} images~\cite{Rattani2017}.
It should be noted that this database was used only in~\cite{Smereka2015} and no availability information is reported.

Sequeira et al.~\cite{Sequeira2014b} built the \mobbioFake database to investigate iris liveliness detection using images taken from mobile devices under an uncontrolled environment.
It consists of $1{,}600$ fake iris images obtained from a subset of the \mobbio database~\cite{Sequeira2014a}.
The fake images were generated by printing the original images using a professional printer in a high-quality photo paper and recapturing the image with the same device and environmental conditions used in the construction of \mobbio.

The images that compose the \iiitdMSP database were obtained under a controlled environment at \gls{nir}, \gls{vis}, and night-vision spectra.
The \gls{nir} images were captured by a Cogent iris Scanner sensor at a distance of 6 inches from the subject, while the night vision subset was created using the Sony Handycam camera in night vision mode at a distance of 1.3 meters.
The \gls*{vis} images were captured with the Nikon SLR camera, also at a distance of 1.3 meters.
The database contains $1{,}240$ images belonging to $62$ subjects, being $310$ images, $5$ from each subject, at \gls*{vis} and night vision spectra, and $620$ images, $10$ from each subject, at \gls*{nir}~spectrum~\cite{Sharma2014}.

Nalla and Kumar~\cite{Nalla2017} developed the \polyu database to study iris recognition in the cross-spectral scenario.
The images were obtained simultaneously under \gls{vis} and \gls{nir} illumination, totaling $12{,}540$ images from $209$ subjects with $640\times480$ pixels of resolution, being $15$ images from each eye in each~spectrum.

To evaluate the state of the art on iris recognition using images acquired by mobile devices, the Mobile Iris Challenge Evaluation (MICHE) competition (Part I) was created~\cite{DeMarsico2015}.
The \miche (or MICHEDB) database consists of $3{,}732$ \gls{vis} images obtained by mobile devices from $92$ subjects.
To simulate a real application, the iris images were obtained by the users themselves, indoors and outdoors, with and without glasses.
Images of only one eye of each individual were captured.
The mobile devices used and their respective resolutions are iPhone5 ($1536\times2048$), Galaxy S4 ($2322\times4128$) and Galaxy Tablet II ($640\times480$).
Due to the acquisition mode and the purpose of the database, several noises are found in images such as specular reflections, focus, motion blur, lighting variations, occlusion due to eyelids, among others.
The authors also proposed a subset, called MICHE FAKE, containing $80$ printed iris images.
Such images were created as follows.
First, they were captured using the iPhone5 the Samsung Galaxy S4 mobile devices.
Then, using a LaserJet printer, the images were printed and captured again by a Samsung Galaxy S4 smartphone.
There is still another subset, called MICHE Video, containing videos of irises from $10$ subjects obtained indoor and outdoor.
A Samsung Galaxy S4 and a Samsung Galaxy Tab 2 mobile devices were used to capture these videos.
In total, this subset has $120$ videos of approximately $15$ seconds each.

The \vssiris database, proposed by Raja et al.~\cite{Raja2015}, has a total of $560$ images captured in a single session under an uncontrolled environment from $28$ subjects.
The purpose of this database is to investigate the mixed lighting effect (natural daylight and artificial indoor) for iris recognition at the \gls*{vis} spectrum with images obtained by mobile devices~\cite{Raja2015}.
More specifically, the images were acquired by the rear camera of two smartphones: an iPhone 5S, with a resolution of $3264\times2448$ pixels, and a Nokia Lumia~1020, with a resolution of $7712\times5360$ pixels.

Santos et al.~\cite{Santos2015} created the \csip (Cross-Sensor Iris and Periocular) database simulating mobile application scenarios.
This database has images captured by four different device models: Xperia Arc~S~(Sony Ericsson), iPhone~4~(Apple), w200~(THL) and U8510~(Huawei).
The resolutions of the images taken with these devices are as follows: Xperia Arc S (Rear $3264\times2448$), iPhone 4 (Front $640\times480$, Rear $2592\times1936$), W200 (Front $2592\times1936$, Rear $3264\times2448$) and U8510 (Front $640\times480$, Rear $2048\times1536$).
Combining the models with front and rear cameras, as well as flash, $10$ different setups were created with the images obtained.
In order to simulate noise variation, the image capture sessions were carried out in different sites with the following lighting conditions: artificial, natural and mixed.
Several noise factors are presented in these images, such as different scales, off-angle, defocus, gaze, occlusion, reflection, rotation and distortions~\cite{Santos2015}.
The database has $2{,}004$ images from $50$ subjects and the binary iris segmentation masks were obtained using the method described by Tan et al.~\cite{Tan2010}~(winners of the NICE~I~contest).

The \visob database was created for the ICIP 2016 Competition on mobile ocular biometric recognition, whose main objective was to evaluate methods for mobile ocular recognition using images taken at the visible spectrum~\cite{Rattani2016}.
The front cameras of $3$ mobile devices were used to obtain the images: iPhone 5S at 720p resolution, Samsung Note 4 at 1080p resolution and Oppo N1 at 1080p resolution.
The images were captured in $2$ sessions for each one of the 2 visits, which occurred between 2 and 4 weeks, counting in the total $158{,}136$ images from $550$ subjects.
At each visit, it was required that each volunteer (subject) capture their face using each one of the three mobile devices at a distance between $8$ and $12$ inches from the face.
For each image capture session, $3$ light conditions settings were applied: regular office light, dim light, and natural daylight.
The collected images were preprocessed using the Viola-Jones eye detector and the region of the image containing the eyes was cropped to a size of $240\times160$~pixels.

Sequeira et al.~\cite{Sequeira2016, Sequeira2017} created the Cross-Spectral Iris/Periocular~(\crossEyed) database to investigate iris and periocular region recognition in cross-spectral scenarios.
\crossEyed is composed of \gls{vis} and \gls{nir} spectrum images obtained simultaneously with $2$K$\times2$K pixel resolution cameras.
The database is organized into three subsets: ocular, periocular (without iris and sclera regions) and iris.
There are $3{,}840$ images from $120$ subjects ($240$ classes), being $8$ samples from each of the classes for every spectrum.
The periocular/ocular images have dimensions of $900\times800$ pixels, while the iris images have dimensions of $400\times300$ pixels.
All images were obtained at a distance of $1.5$ meters, under uncontrolled indoor environment, with a wide variation of ethnicity and eye colors, and lightning reflexes.

The \pmhi database was collected to investigate the post-mortem human iris recognition.
Due to the difficulty and restriction in collecting such images, this database has only $104$ images from $6$ subjects.
The images were acquired in three sessions with an interval of approximately $11$ hours using the IriShield M2120U~\gls{nir} and Olympus~TG-3~\gls{vis} cameras.

The \qutMP database was developed and used by Algashaam et al.~\cite{Algashaam2017} to study multi-spectral periocular recognition.
In total, $212$ images belonging to $53$ subjects were captured at \gls{vis}, \gls{nir} and night vision spectrum with $800\times600$ pixels of resolution.
The \gls{vis} and \gls{nir} images were taken using a Sony DCR-DVD653E camera, while the night vision images were acquired with an IP2M-842B~camera.

Regarding some ocular biometrics problems caused by substantial degradation due to variations on illumination, distance, noise, and blur when using single-frame mobile captures, Nguyen et al.~\cite{nguyen2020visob2} created the \visobtwo database.
This database comprises multi-frame captures and has stacks of eye images acquired using the burst mode of two mobile devices: Samsung Note 4 and Oppo N1.
It is the second version of the \visob database and was used in the 2020 IEEE WCCI competition~\cite{nguyen2020visob2}.
The images were collected in two visits.
At each visit, the subjects collected their own images under three lighting conditions in two sessions.
The available subset of the \visobtwo database (competition training set) has $75{,}428$ images of left and right eyes belonging to $150$ subjects.
The \visobtwo can also be employed to investigate the probing fairness of ocular biometrics across gender~\cite{krishnan2020probing}.

The Iris Social Database (\isocialdb) has $3{,}286$ \gls{vis} images from $400$ subjects, being $43.75\%$ male and $56.25\%$ female.
It is composed of images of public persons such as artists and athletes.
This database was created by collecting $1{,}643$ high-resolution portrait images using Google Image Search.
Then, the ocular regions were cropped as rectangles of $350\times300$ pixels.
The binary masks for the iris region (created by a human expert) are also available.
This database can be employed to evaluate iris segmentation and recognition under unconstrained~scenarios.

The \ufprPerioc database has \gls{vis} images acquired in unconstrained environments by mobile devices.
These images were captured by the subjects themselves using their own smartphone models through a mobile application~(app) developed by the authors~\cite{zanlorensi2020ufprperiocular}.
In total, this database contains $33{,}660$ samples from $1{,}122$ subjects acquired during $3$ sessions by $196$ different mobile devices.
The image resolutions vary from $360\times160$ to $1862\times1008$ pixels.
The main intra- and inter-class variability are caused by occlusion, blur, motion blur, specular reflection, eyeglasses, off-angle, eye-gaze, makeup, facial expression, and variations in lighting, distance, and angles.
The authors manually annotated the eye corners and used them to normalize the periocular images regarding scale and rotation.
This database can also be employed to investigate gender recognition, age estimation, and the effect of intra-class variability in biometric systems.
The \ufprPerioc database, which includes the manual annotations of the eye corners, as well as information on the subjects' age and gender, is publicly available for the research~community.

Zanlorensi et al.~\cite{zanlorensi2020attnormalization} created the \ufprEyeglass database to investigate intra-class variability and also the effect of the occlusion by eyeglasses in periocular recognition under uncontrolled environments.
This database has $2{,}270$ images captured by mobile devices from $83$ subjects with a resolution of $256\times256$ pixels.
The subjects captured the images using the same mobile app used to collect the \ufprPerioc database.
This database can be considered a subset of the \ufprPerioc database containing some additional images.
The authors manually annotated the iris's bounding box in each image and used it to perform scale and rotation normalization.
The intra-class variations in this database are mainly caused by illumination, occlusions, distances, reflection, eyeglasses, and image quality.
The \ufprEyeglass database, which includes the authors' manual annotations, is publicly available to the research community.

\subsection{Multimodal Databases}

\begin{figure}[!ht]
\centering
\includegraphics[width=\imgsize\columnwidth]{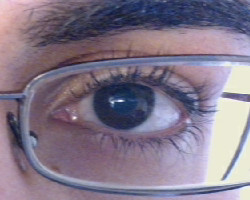}
\includegraphics[width=\imgsize\columnwidth]{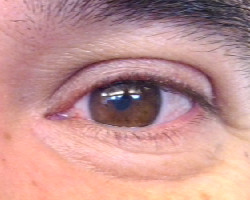}
\includegraphics[width=\imgsize\columnwidth]{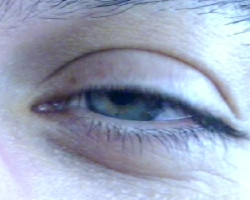}
\includegraphics[width=\imgsize\columnwidth]{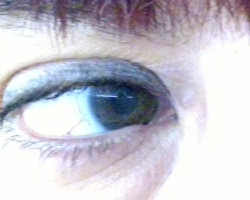}
\includegraphics[width=\imgsize\columnwidth]{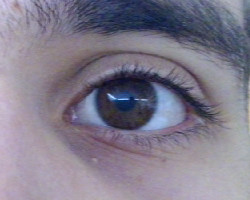}

\vspace{1mm}

\includegraphics[width=\imgsize\columnwidth]{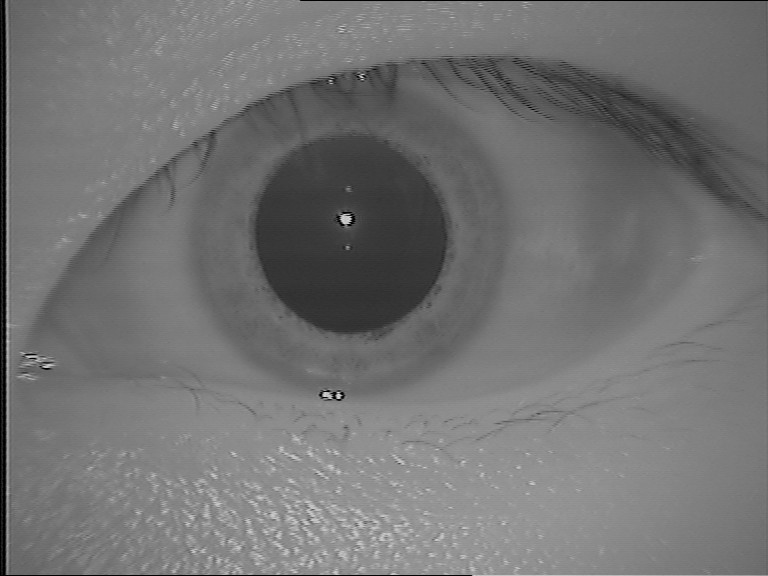}
\includegraphics[width=\imgsize\columnwidth]{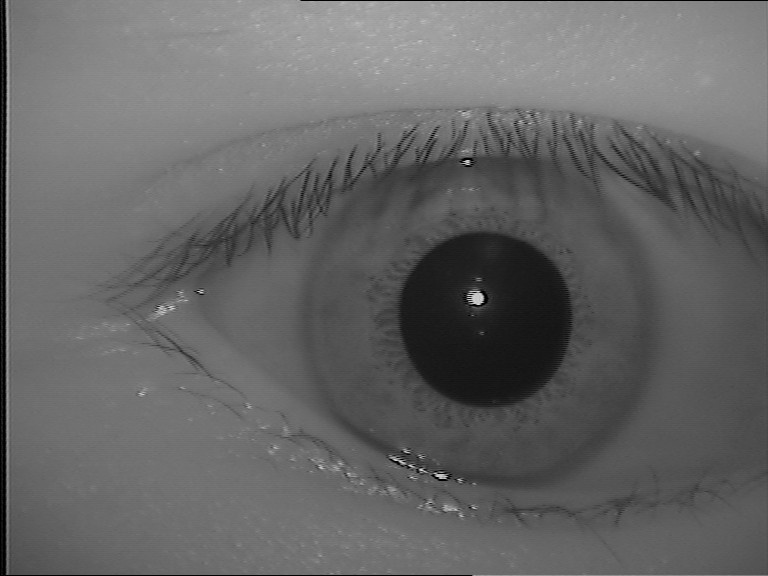}
\includegraphics[width=\imgsize\columnwidth]{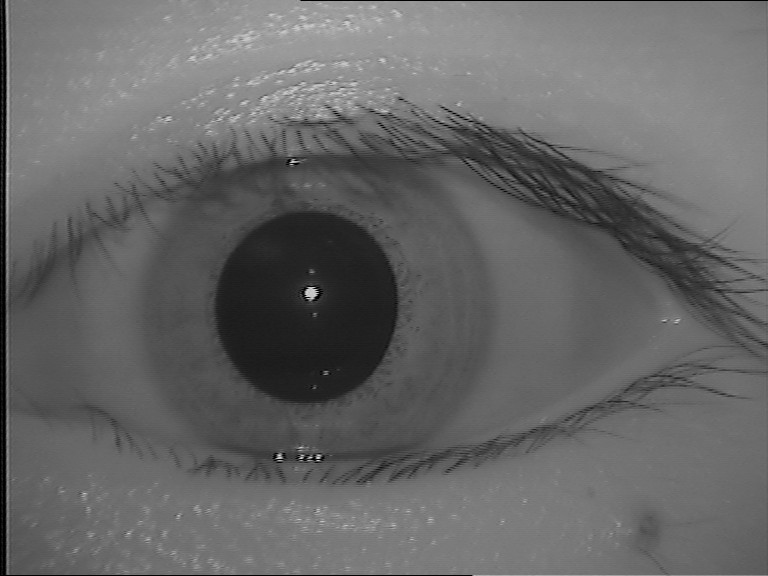}
\includegraphics[width=\imgsize\columnwidth]{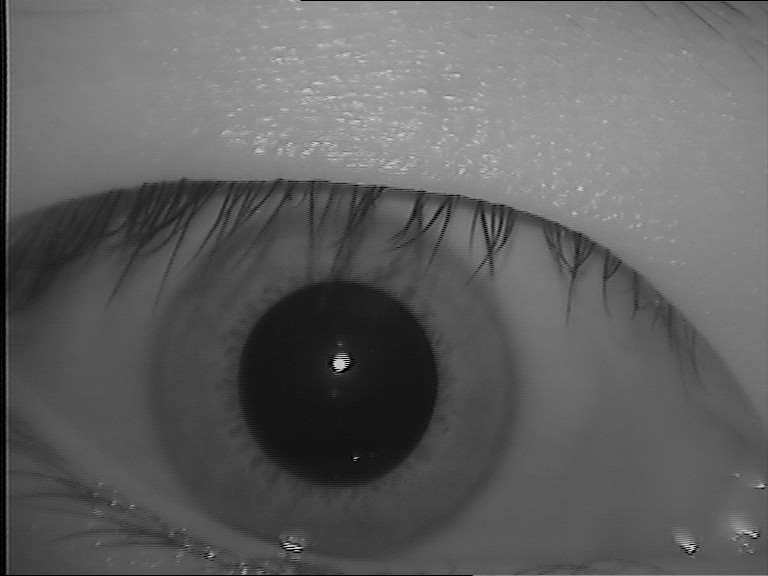}
\includegraphics[width=\imgsize\columnwidth]{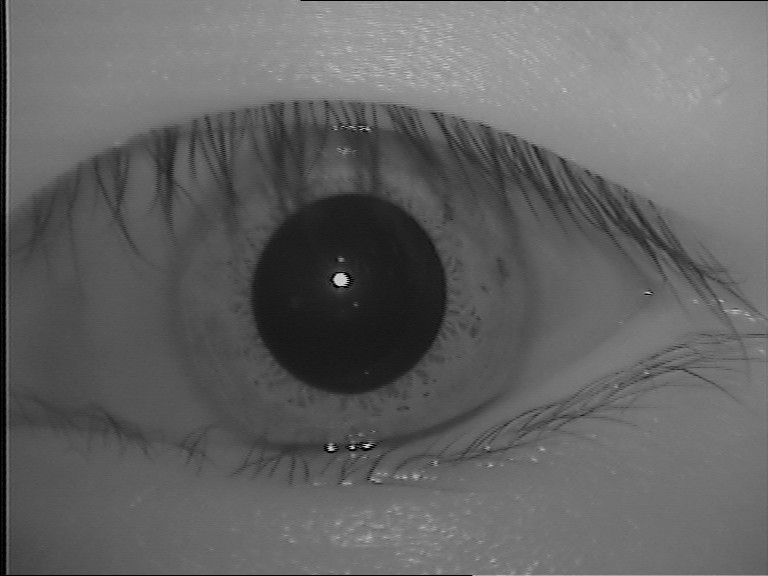}

\vspace{1mm}

\includegraphics[width=\imgsize\columnwidth]{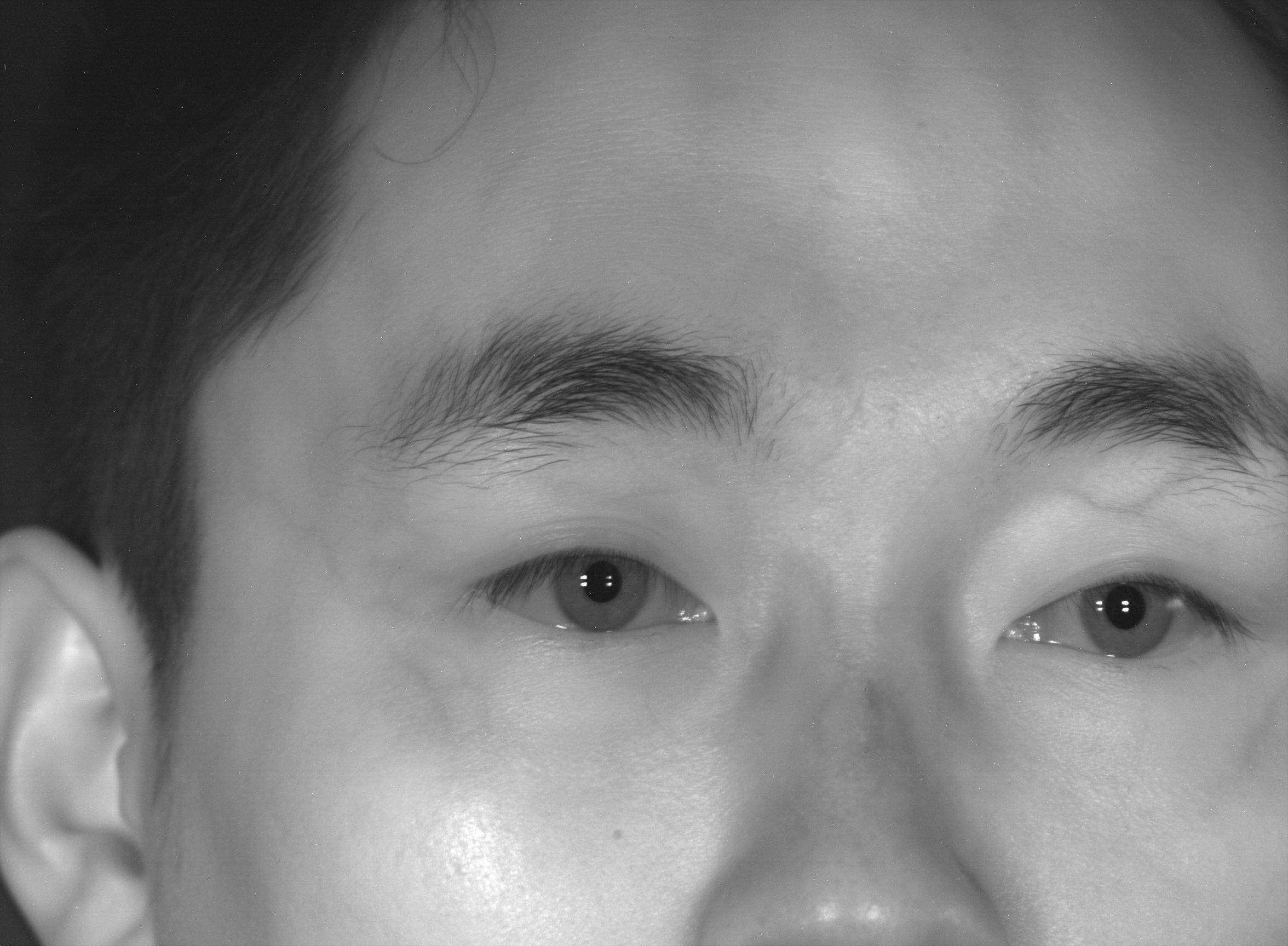}
\includegraphics[width=\imgsize\columnwidth]{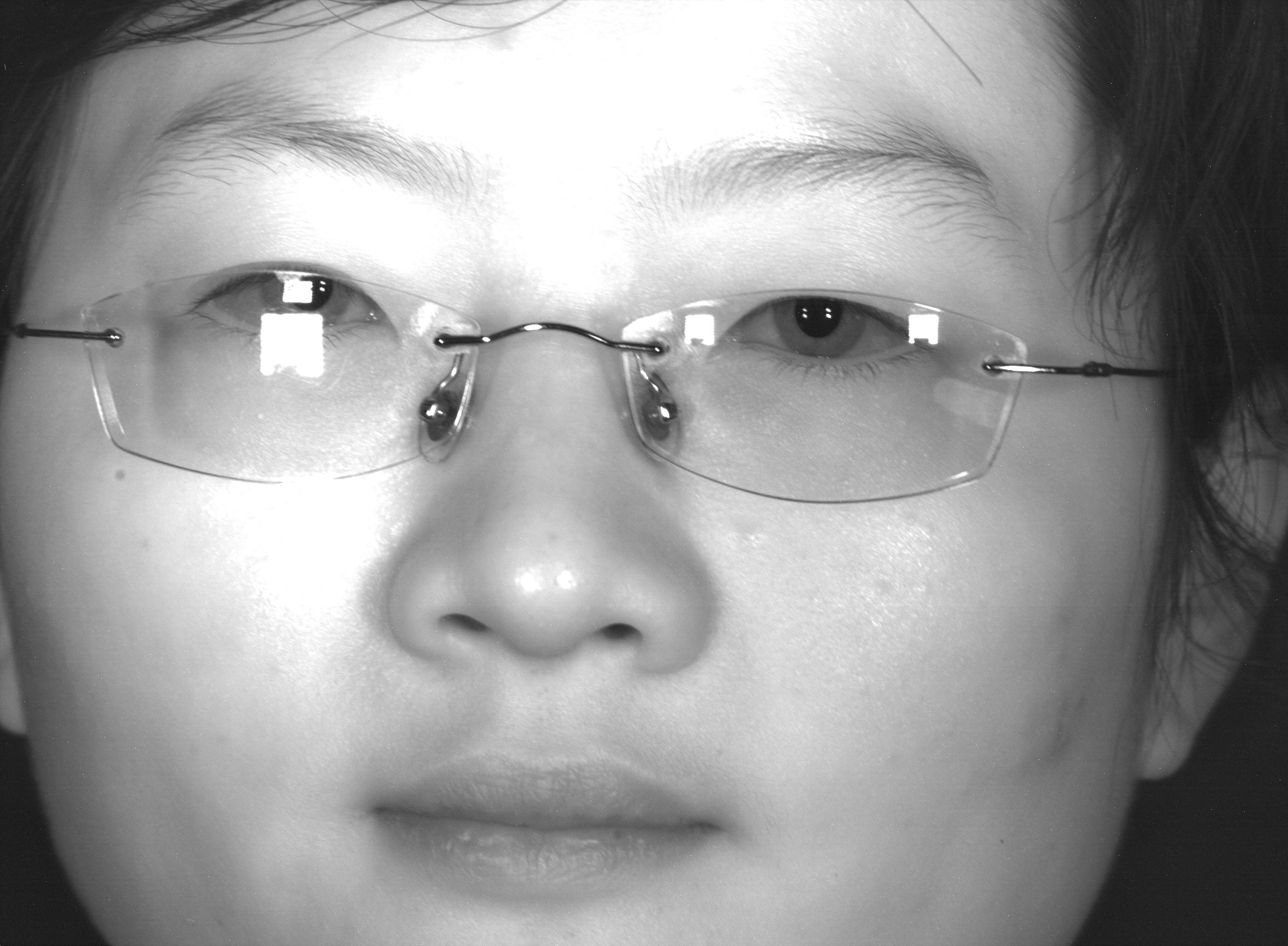}
\includegraphics[width=\imgsize\columnwidth]{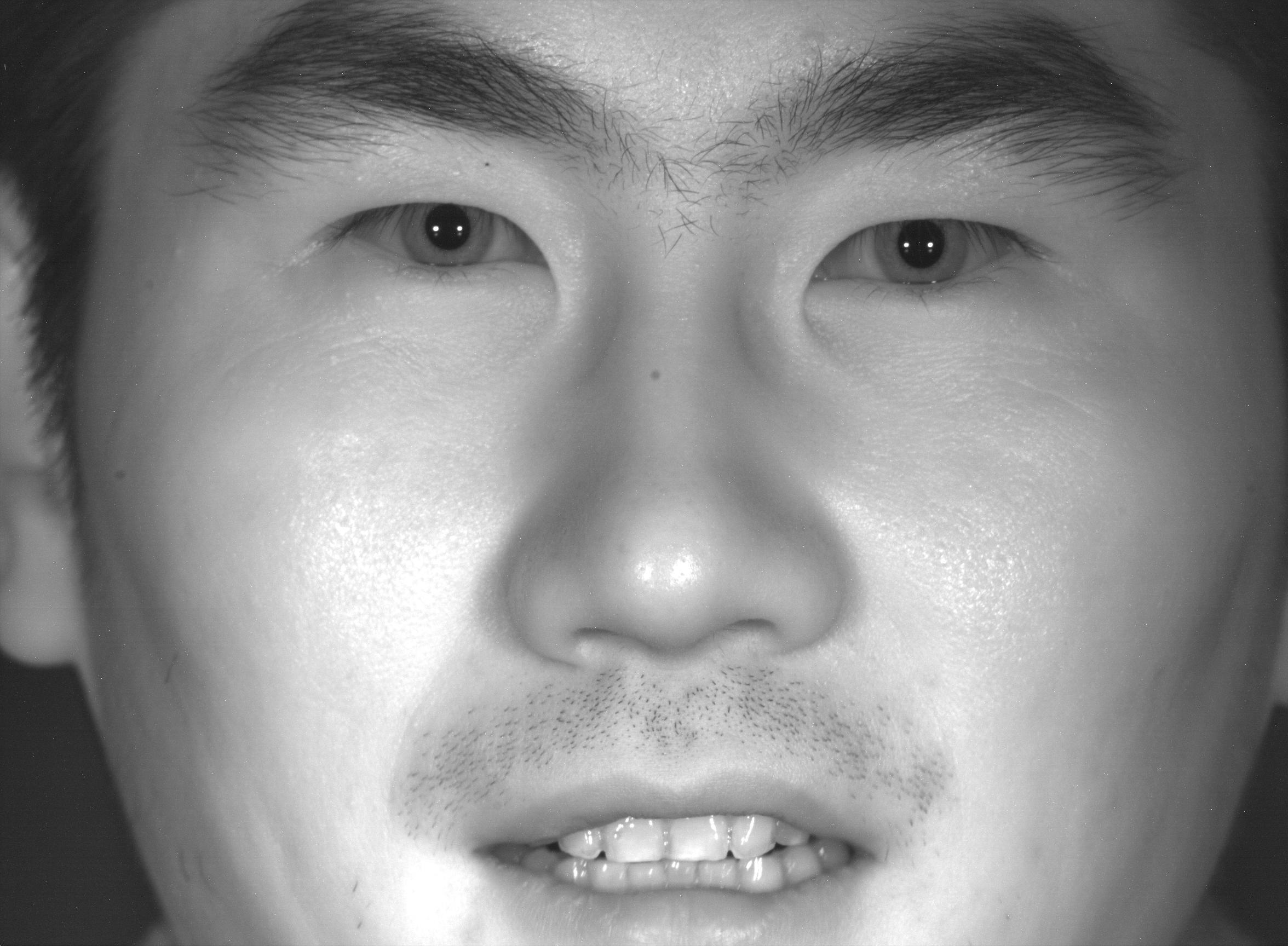}
\includegraphics[width=\imgsize\columnwidth]{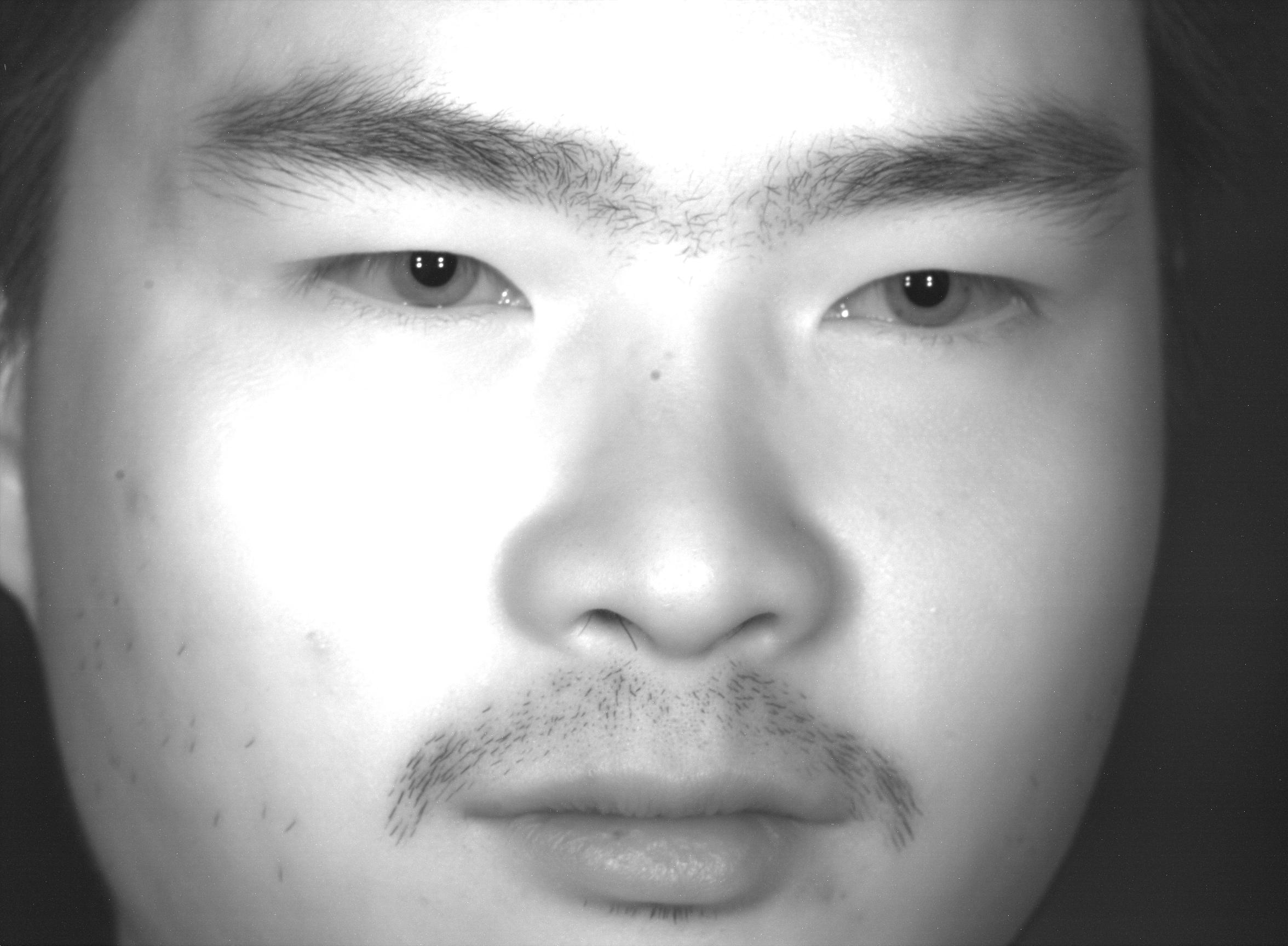}
\includegraphics[width=\imgsize\columnwidth]{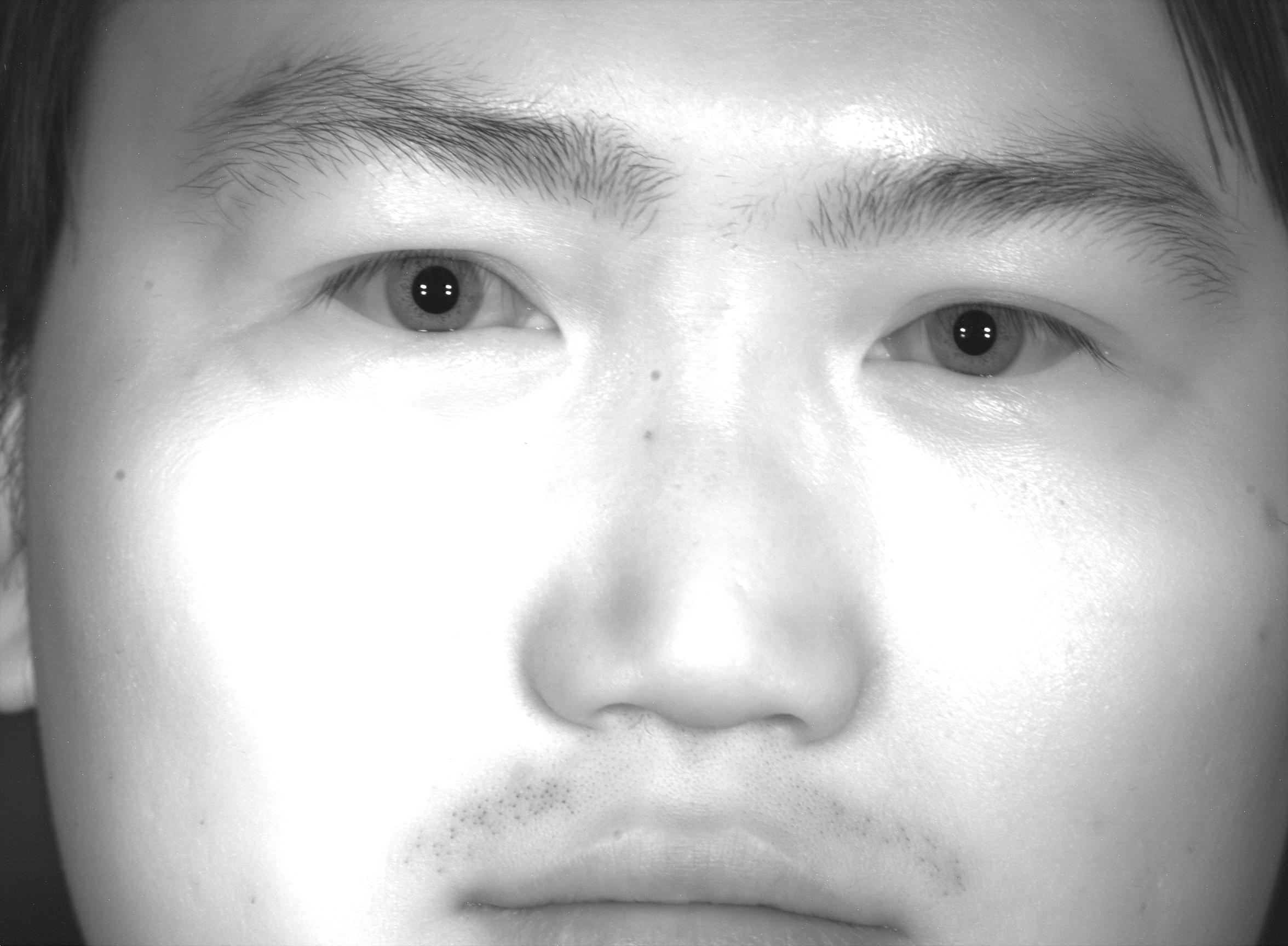}

\caption{From top to bottom: ocular image samples from the \mobbio~\cite{Sequeira2014a}, \sdumla~\cite{Yin2011} and \casiaDistance~\cite{CASIA2010} multimodal~databases.}
\label{fig:multisamples}
\end{figure}

In addition to the databases proposed specifically to assist the development and evaluation of new methodologies for iris/periocular recognition, some multimodal databases can also be used for this purpose.
Table~\ref{tab:multidata} show these databases.
As described in this subsection, most of these databases consist of iris images obtained at \gls{nir} wavelength.
Figure~\ref{fig:multisamples} shows samples of ocular images from some multimodal~databases.

\begin{table*}[!ht]
\setlength{\tabcolsep}{8pt}
\centering
\caption{Multimodal databases. Modalities: Iris [IR], Periocular [PR], Face [FC], Fingerprint [FP], Voice [VC], Speech [SP], Signature [SG], Handwriting [HW], Hand [HD], Finger vein [FV], Gait [GT] and KeyStroking [KS].}
\vspace{0.5mm}
\label{tab:multidata}
\resizebox{\linewidth}{!}{
\begin{tabular}{lccccccc}
\toprule
Database                         & Year    &\begin{tabular}[c]{@{}c@{}}Controlled\\ Environment\end{tabular}  & Wavelength  & Cross-sensor & Subjects & Images & Modality \\
\midrule
\biosec \cite{Fierrez2007}       & $2006$  & No   & NIR  & No  & $200$  & $3{,}200$           & [IR]/[FC]/[FP]/[VC] \\
\biosecur \cite{Fierrez2010}     & $2007$  & Yes  & NIR  & No  & $400$  & $12{,}800$          & [IR]/[FC]/[SP]/[SG]/[FP]/[HD]/[HW]/[KS] \\
\bmdb \cite{Ortega2010}          & $2008$  & Yes  & NIR  & No  & $667$  & $5{,}336$           & [IR]/[FC]/[SP]/[SG]/[FP]/[HD] \\
\mbgc \cite{MBGC2010}            & $2009$  & No   & NIR  & No  & *$268$ eyes & *$986$ videos  & [IR]/[FC] \\
\qFire \cite{Johnson2010}        & $2010$  & No   & NIR  & No  & $195$ & N/A                  & [IR]/[FC] \\
\focs \cite{FOCS2010}            & $2010$  & No   & NIR  & No  & $136$  & $9{,}581$           & [IR]/[PR]/[FC] \\
\casiaDistance \cite{CASIA2010}  & $2010$  & Yes  & NIR  & No  & $142$  & $2{,}567$           & [IR]/[PR]/[FC] \\
\sdumla \cite{Yin2011}           & $2011$  & Yes  & NIR  & No  & $106$  & $1{,}060$           & [IR]/[FC]/[FV]/[GT]/[FP] \\
\mobbio \cite{Sequeira2014a}     & $2013$  & No   & VIS  & No  & $105$  & $1{,}680$           & [IR]/[FC]/[VC] \\
\gbMOD \cite{Rios-Sanchez2016}   & $2015$  & Yes  & NIR  & No  & $60$  & *$600$ videos        & [IR]/[FC]/[HD] \\
\bottomrule

\end{tabular}
}
\end{table*}

The \biosec baseline database, proposed by Fierrez et al.~\cite{Fierrez2007}, has biometric data of fingerprint, face, iris and voice.
Data were acquired from $200$ subjects in two acquisition sessions, with environmental conditions (e.g., lighting and background noise) not being controlled to simulate a real situation.
There are $3{,}200$ \gls{nir} iris images, being $4$ images of each eye for each session, captured by an LG IrisAccess EOU3000~camera~\cite{Fierrez2007}.

The \biosecur multimodal database consists of $8$ unimodal biometric traits: iris, face, speech, signature, fingerprints, hand, handwriting, and keystroking~\cite{Fierrez2010}.
The authors collected data from $400$ subjects in four acquisition sessions through $4$ months at six Spanish institutions.
The iris images were captured at \gls{nir} by an LG Iris Access EOU 3000 camera with a resolution of $640\times480$ pixels.
Four images of each eye were obtained in each of the $4$ sessions, totaling $32$ images per individual and a final set of $12{,}800$ iris images.

The \bmdb (multienvironment multiscale BioSecure Multimodal Database)~\cite{Ortega2010} has biometric data from more than $600$ subjects, obtained from $11$ European institutions participating in the BioSecure Network of Excellence~\cite{Ortega2010}.
This database contains biometric data of iris, face, speech, signature, fingerprint and hand, and is organized into three subsets: DS1, which has data collected from the Internet under unsupervised conditions;
DS2, with data obtained in an office environment under supervision;
and DS3, in which mobile hardware was used to take data indoor and outdoor.
The iris images belong to the DS2 subset and were obtained in $2$ sessions at \gls{nir} wavelength in an indoor environment with supervision.
For the acquisition, the use of contact lenses was accepted, but glasses needed to be removed.
Four images (2 of each eye) were obtained in each session for each of the $667$ subjects, totaling $5{,}336$ images.
These images have a resolution of $640\times480$ pixels and were acquired by an LG Iris Access EOU3000~sensor.

The goal of the Multiple Biometrics Grand Challenge (\mbgc)~\cite{MBGC2010} was the evaluation of iris and face recognition methods using data obtained from still images and videos under unconstrained conditions~\cite{Phillips2009}.
The \mbgc is divided into three problems: the portal challenge problem, the still face challenge problem, and the video challenge problem~\cite{MBGC2010}.
This competition has two versions.
The first one was held to introduce the problems and protocol, whereas version 2 was released to evaluate the approaches in large databases~\cite{Phillips2009}.
The iris images were obtained from videos captured at \gls{nir} by an Iridian LG EOU 2200 camera~\cite{Hollingsworth2009}.
The videos present variations such as pose, illumination, and camera angle.
The \mbgc database has $986$ iris videos from $268$ eyes collected in~2008~\cite{Hollingsworth2009}. 

The \qFire database (Quality in Face and Iris Research Ensemble) has iris and face images from $195$ subjects, obtained through videos at different distances~\cite{Johnson2010}.
This database has $28$ and $27$ videos of face and iris, respectively, captured in $2$ sections, with varying camera distance between $5$, $7$, $11$, $15$ and $25$ feet.
The videos have approximately $6$ seconds each and were captured at approximately $25$ frames per second.
A Dalsa 4M30 infrared camera equipped with a Tamron AF 70-300mm 1:4.5-5.6 LD DI lens were used to capture iris videos.
For distances of $15$ and $25$ feet, a Sigma APO 300-800mm F5.6 EX DG HSM lens was used.
The most attractive distance of capture for iris is $5$~($300\times280$ pixels), $7$~($220\times$200 pixels), and $11$~($120\times100$ pixels) feet since they respectively represent high, medium and low resolution, based on the number of pixels in the iris diameter.
The images also have variations of illumination, defocus, blur, eye angles, motion blur, and occlusions~\cite{Johnson2010}.

The \gls{nir} images from the ocular region (iris and periocular) of the \focs database~\cite{FOCS2010} were extracted from the \mbgc database~\cite{MBGC2010} videos, which were collected from moving subjects~\cite{Matey2006}.
These videos were captured in an uncontrolled environment presenting some variations such as noise, gaze, occlusion and lighting.
The database has $9{,}581$ images ($4{,}792$ left, $4{,}789$ right) with a resolution of $750\times600$ pixels from $136$ subjects~\cite{Smereka2015}.

Their system can recognize users from up to $3$ meters ($10$ feet) using a system with an active search for iris, face or palmprint patterns.
The images were taken using a camera with high resolution so that a single image includes regions of interest for both eyes and face traits.
Information from the face trait such as skin pattern can also be used for multi-modal fusion.
The database has $2{,}567$ images from $142$ individuals and $284$ classes with a resolution of $2352\times1728$~pixels.

The \sdumla multimodal database contains biometric traits of iris, face, finger vein, gait, and fingerprint~\cite{Yin2011}.
All data belong to $106$ subjects and were collected at Shandong University in China.
The iris images were collected at \gls{nir} and under a controlled environment at a distance of $6$~cm to $32$~cm between the camera and the subject.
In total, the authors collected $1{,}060$ iris images with $768\times576$ pixels of resolution, being $10$ images ($5$ of each eye) from each subject~\cite{Yin2011}.

Sequeira et al.~\cite{Sequeira2014a} created the \mobbio database due to the growing interest in mobile biometric applications, as well as the growing interest and application of multimodal biometrics.
This database has data from iris, face, and voice belonging to $105$ subjects.
The data were obtained using an Asus TPad TF~300T mobile device, and the images were captured using the rear camera of this device in $8$ MP of resolution.
The iris images were obtained at \gls{vis} and in two different illumination conditions varying eye orientations and occlusion levels.
For each subject, $16$ images ($8$ of each eye, cropped from an image of both eyes) were captured.
The cropped images have a resolution of $300\times200$ pixels.
Manual annotations of the iris and pupil contours are provided along with the database, but iris illumination noises are not~identified.

The gb2s$\mu$MOD database is composed of $8{,}160$ iris, face and hand videos belonging to $60$ subjects and captured in three sessions with environment condition variation~\cite{Rios-Sanchez2016}.
Sessions $1$ and $2$ were obtained in a controlled environment, while session $3$ was acquired in an uncontrolled environment.
The iris videos were recorded only in sessions $1$ and $2$ with a \gls{nir} camera (850~nm) held by the subject himself as close to the face as possible capturing both eyes.
The diameter of the iris in such videos is approximately $60$ pixels.
Ten iris videos were collected in two ($5$ in each session) for each one of the $60$ subjects.
Along with the videos, information such as name, ID card number, age, gender, and handedness are also available.

All databases described in this subsection contain iris and/or periocular subsets, however, some databases that do not have such subsets can also be employed for iris/periocular recognition. For example, the FRGC~\cite{Phillips2005} database, which is a database of face images, has already been used for iris~\cite{Tan2013} and periocular~\cite{Woodard2010, Park2011, Smereka2015} recognition in the~literature.
\section{Ocular Recognition Competitions}
\label{sec:eyeCont}

In this section, we describe the major recent competitions and the algorithms that achieved the best results in iris and/or periocular region information.
Through these competitions, it is possible to demonstrate the advancement in terms of methodologies for ocular biometrics and also the current challenges in this research~area.

The competitions usually provide a database in which the competitors must perform their experiments and submit their algorithms.
Once submitted, the algorithms are evaluated with another subset of the database, according to the metrics established by the competition protocol.
In this way, it is possible to fairly assess the performance of different methodologies for specific objectives.

In ocular biometrics including iris and periocular recognition, there are several competitions aimed at evaluating different situations, such as 
recognition in images captured at~\gls{nir} and/or~\gls{vis} wavelengths, images captured in an uncontrolled environment, images obtained with mobile devices, among others.
For each competition, we describe the approaches that achieved the best results using fused information from iris and periocular region, and also the best performing methodologies using only iris information.
Table~\ref{contest} presents the main competitions held in recent years and the best results achieved, while Table~\ref{method_contest} details the methodologies that obtained the best results in these~competitions.

\begin{table*}[!ht]
\setlength{\tabcolsep}{8pt}
\centering
\caption{Best results achieved in ocular biometric competitions.}
\label{contest}
\resizebox{\linewidth}{!}{
\begin{tabular}{lccccccc}
\toprule
Competition                             & Year & Database               & Wavelength     & Best Result                                                      & Traits                                                      \\
\midrule
NICE.II \cite{Proenca2012}              & 2010 & portion of UBIRIS v2   & VIS            &  DI = $2.57$~\cite{Tan2012}                                      & Iris + Periocular \\
NICE.II \cite{Proenca2012}              & 2010 & portion of UBIRIS v2   & VIS            &  DI = $1.82$~\cite{Wang2012}                                     & Iris              \\
MICHE-II \cite{DeMarsico2017}           & 2016 & \miche and MICHE-II    & VIS            & AVG = $1.00$~\cite{Ahmed2016, Ahmed2017}                         & Iris + Periocular \\
MICHE-II \cite{DeMarsico2017}           & 2016 & \miche and MICHE-II    & VIS            & AVG = $0.86$~\cite{Raja2017}                                     & Iris              \\
MIR \cite{Zhang2016}                    & 2016 & MIR-Train and MIR-Test & NIR            & FNMR4 = $2.24\%$, EER = $1.41\%$ e DI = $3.33$~\cite{Zhang2016}  & Iris              \\
VISOB 1.0~\cite{Rattani2016}            & 2016 & \visob                 & VIS            & EER = $0.06\%$ - $0.20\%$~\cite{Raghavendra2016}                 & Periocular        \\
\crossEyed~\cite{Sequeira2016}          & 2016 & \crossEyed             & Cross-spectral & GF2 = $0.00\%$ and EER = $0.29\%$~($HH_1$)~\cite{Sequeira2016}   & Periocular        \\
\crossEyed~\cite{Sequeira2016}          & 2016 & \crossEyed             & Cross-spectral & GF2 = $3.31\%$ and EER = $2.78\%$~($NTNU_6$)~\cite{Sequeira2016} & Iris              \\
2$^{nd}$ \crossEyed~\cite{Sequeira2017} & 2017 & \crossEyed             & Cross-spectral & GF2 = $0.00\%$ and EER = $0.05\%$~($NTNU_4$)~\cite{Sequeira2017} & Iris              \\
2$^{nd}$ \crossEyed~\cite{Sequeira2017} & 2017 & \crossEyed             & Cross-spectral & GF2 = $0.74\%$ and EER = $0.82\%$~($HH_1$)~\cite{Sequeira2017}   & Periocular        \\
VISOB 2.0~\cite{nguyen2020visob2}  & 2020 & \visobtwo    & VIS       & EER = $5.25\%$ and AUC = $98.8\%$~\cite{zanlorensi2019cross}                   & Periocular   \\

\bottomrule

\end{tabular}

}

\end{table*}

\begin{table*}[!ht]
\setlength{\tabcolsep}{8pt}
\centering
\caption{Best methodologies in ocular biometric competitions.}
\label{method_contest}
\resizebox{\linewidth}{!}{
\begin{tabular}{@{}L{4.4cm}C{4.5cm}C{4.5cm}C{4.5cm}C{4.5cm}C{2cm}@{}}
\toprule
Contest/Author          & Periocular Features                                      & Iris Features                         & Periocular Matching                                       & Iris Matching                                                    & Fusion Technique       \\
\midrule
NICE.II~\cite{Tan2012}  & Texton histogram and Semantic information                & Ordinal measures and color histogram  & chi-square distance and exclusive or                      & SOBoost and diffusion distance                                    & Sum rule               \\
NICE.II~\cite{Wang2012} & -                                                        & 2D Gabor                              & -                                                         & AdaBoost learning                                                & -                      \\
MICHE-II~\cite{Ahmed2016, Ahmed2017}  & Multi-Block Transitional Local Binary Pattern (MB-TLBP)  & 1D Log-Gabor filter                   & chi-square distance                                       & Hamming distance                                                 & Weighted sum of scores \\
MICHE-II~\cite{Raja2017}      & -                                                        & Deep sparse filters                   & -                                                         & Maximized likelihood in a collaborative subspace representation  & -                      \\
MIR~\cite{Zhang2016}         & -                                                        & Gabor wavelet                         & -                                                         & Cosine distance and hamming distance                             & -                      \\
VISOB 1.0~\cite{Raghavendra2016}        & Maximum Response (MR) filters                            & -                                     & DNN based on deeply coupled autoencoders  & -                                                                & -                      \\
\crossEyed~$HH_1$~\cite{Sequeira2016}  & SAFE, GABOR, SIFT, LBP and HOG                           & -                                     & Probabilistic bayesian                          & -                                                                & -                      \\
\crossEyed~$NTNU_6$~\cite{Sequeira2016}  & -                                                        & M-BSIF                                & -                                                         & chi-square distance and SVM                                      & -                      \\

2$^{nd}$~\crossEyed~$NTNU_4$~\cite{Sequeira2017}  & -  & M-BSIF                                     & -                          & chi-square distance                                                                & -                      \\
2$^{nd}$ \crossEyed~$HH_1$~\cite{Sequeira2017}  & SAFE, GABOR, SIFT, LBP and HOG                                                        & -                                & Probabilistic bayesian                                                         & -                                      & -                      \\
VISOB 2.0~\cite{nguyen2020visob2}        & ResNet-50                            & -                                     & Cosine Distance  & -                                                                & -                      \\

\bottomrule

\end{tabular}
}
\end{table*}

\subsection{NICE - Noisy Iris Challenge Evaluation}

The Noisy Iris Challenge Evaluation~(NICE) competition contains two different contests.
In the first one (NICE.I), held in 2008, the goal was the evaluation of methods for iris segmentation to remove noise factors such as specular reflections and occlusions.
Regarding the evaluation of encoding and matching methods, the second competition (NICE.II), was carried out in 2010.
The databases used in both competitions are subsets of~UBIRIS.v2~\cite{proenca2010}, which contains~\gls{vis} ocular images captured under uncontrolled~environments.

Described by Proença and Alexandre~\cite{Proenca2012}, the first competition aimed to answer: ``is it possible to automatically segment a small target as the iris in unconstrained data (obtained in a non-cooperative environment)?''
In total, $97$ research laboratories from $22$ countries participate in the competition.
The training set consisted of $500$ images, and their respective manually generated binary iris masks. 
The committee evaluated the proposed approaches using another $500$ images through a pixel-to-pixel comparison between the original and the generated segmentation masks.
As a metric, the organizers choose the following error rate based on pixel-level:

\begin{equation}
E_{j} = \frac{1}{n w h} \sum_{i=1}^{n} \sum_{r=1}^{h} \sum_{c=1}^{w} P_{i}(r,c) \otimes G_{i}(r,c) \, ,
\end{equation}

\noindent 
where $n$ refers to the number of test images, $w$ and $h$ are respectively the width and height of these images, $P_{i}(r,c)$ means the intensity of the pixel on row~$r$ and column~$c$ of the $i$th segmentation mask, $G_{i}(r,c)$ is the actual pixel value and $\otimes$ is the or-exclusive~operator.

According to the values of $E_{j}$,  NICE.I's best results are the following: $0.0131$~\cite{Tan2010}, $0.0162$~\cite{Sankowski2010}, $0.0180$~\cite{Almeida2010}, $0.0224$~\cite{Li2010}, $0.0282$~\cite{Jeong2010}, $0.0297$~\cite{Chen2010}, $0.0301$~\cite{DonidaLabati2010}, $0.0305$~\cite{Luengo-Oroz2010}.

The second competition (NICE.II) evaluated only the feature extraction and matching results.
Therefore, all the participants used the same segmented images, which were generated by the winner methodology in the NICE.I contest~\cite{Proenca2012}, proposed by Tan et al.~\cite{Tan2010}.
The main goal was to investigate the impact of noise presented inside the iris region in the biometric recognition process. 
As described in both competitions~\cite{Proenca2012}, these noise factors have different sources, e.g., specular reflection and occlusion,  caused by the uncontrolled environment where the images were taken.
This competition received algorithms sent by $67$ participants from $30$ countries.
The training set consists of $1{,}000$ images and their respective binary masks.
The proposed methods had to receive a pair of images followed by their masks as input and generate an output file containing the dissimilarity scores~($d$) of which pairwise comparison with the following conditions:
\begin{enumerate}
\item $d(I, I) = 0$
\item $d(I_{1}, I_{2}) = 0 \Rightarrow I_{1} = I_{2}$
\item $d(I_{1}, I_{2}) + d(I_{2}, I_{3}) \geq d(I_{1}, I_{3})$.
\end{enumerate}

The submitted approaches were evaluated using a new set of $1{,}000$ images with their binary masks.
Consider $IM = \{I_{1},...,I_{n}\}$ as a collection of iris images, $MA= \{M_{1},...,M_{n}\}$ as their respective masks, and $id(.)$ representing a function that identifies an image.
The comparison protocol one-against-all returns a match set $D^{I} = \{d^{i}_{1},...,d_{im}\}$ and a non-match set $D^{E} = \ d^{e}_{1},...,d_{ek}\}$ 
of dissimilarity scores, where $id(I_{i}) = id(I_{j})$ and $id(I_{i}) \neq id(I_{j})$, respectively.
The algorithms were evaluated using the decidability scores $d'$~\cite{Daugman2004}, which measure the separation level of two distributions. The following overlap area gives this decidability scores $d'$:

\begin{equation}
d' = \frac{|\mu_{E} - \mu_{I}|}{\sqrt{\frac{1}{2} (\sigma^{2}_{I} + \sigma^{2}_{E})}} \, ,
\end{equation}

\noindent
where the means of the two distributions are given by $\mu_{I}$ and $\mu_{E}$, and the standard deviations are represented by $\sigma_{I}$ and $\sigma_{I}$.

The best results of NICE.II ranked by their $d'$ scores are as follows: $2.5748$~\cite{Tan2012}, $1.8213$~\cite{Wang2012}, $1.7786$~\cite{Santos2012}, $1.6398$~\cite{Shin2012}, $1.4758$~\cite{Li2012a}, $1.2565$~\cite{DeMarsico2012}, $1.1892$~\cite{Li2012b}, $1.0931$~\cite{Szewczyk2012}.

The winner method, proposed by Tan et al.~\cite{Tan2012}, achieved a decidability value of $2.5748$ by fusing iris and periocular features.
The fusion process was performed at the score level by the sum rule method.
Therefore, for iris and periocular images, different features and matching techniques were used.
The iris features were extracted with ordinal measures and color histogram and for the periocular ones, texton histogram, and semantic information.
To compute the matching scores, the authors employed the following metrics: SOBoost learning, diffusion distance, chi-square distance, and exclusive OR operator.

Wang et al.~\cite{Wang2012} proposed a method using only iris information.
Their approach was ranked second in the competition, achieving a decidability value of $1.8213$.
The algorithm performed the segmentation and normalization of iris using the Daugman technique~\cite{Daugman1993}.
Features were extracted by applying the Gabor filters from different patches generated from the normalized image.
The AdaBoost algorithm computed a selection of features and the similarity.

The main contribution of NICE competitions was the evaluation of iris segmentation and recognition methods independently, as several iris segmentation methodologies were evaluated in the first competition and the best one was applied to generate the binary masks used in the second one, in which the recognition task was evaluated.
Hence, the approaches described in both competitions can be fairly compared since they employed the same images for training and~testing.

Although NICE.II was intended to evaluate iris recognition systems, some approaches using information from the periocular region were also included in the final ranking.
The winning method fused iris and periocular information, however, it should be noted that some approaches that also fused these two traits achieved lower results than methodologies that used only iris~features.
Moreover, it would be interesting to analyze the best performing approaches in the NICE.II competition in larger databases to verify the scalability of the proposed methodologies, as the database used in these competitions was not composed of a large number of~images/classes.

Some recent works applying deep~\gls{cnn} 
models have achieved state-of-the-art results in the NICE.II database using information from the iris~\cite{Zanlorensi2018}, periocular region~\cite{Luz2018} and fusing iris/periocular traits~\cite{Silva2018} with decidability values of $2.25$, $3.47$, $3.45$,~respectively.

\subsection{MICHE - Mobile Iris Challenge Evaluation}

In order to assess the performance that can be reached in iris recognition without the use of special equipment, the Mobile Iris CHallenge Evaluation II, or simply MICHE-II competition, was held~\cite{DeMarsico2017}.
The \miche database, introduced by De~Marsico et al.~\cite{DeMarsico2015} has $3{,}732$ images taken by mobile devices and was made available to the participants to train their algorithms, while other images obtained in the same way were employed for the~evaluation.

Similarly to NICE.I and NICE.II, MICHE is also divided into two phases.
MICHE.I and MICHE.II focused on iris segmentation and recognition, respectively.
Ensuring a fair assessment and targeting only the recognition step, all MICHE.II participants used the segmentation algorithm proposed by Haindl and Krupicka~\cite{Haindl2015}, which achieved the best performance on~MICHE.I.

The performance of each algorithm was evaluated through dissimilarity.
Assuming $I$ as a set of the MICHE.II database and that $I_{a}, I_{b} \in I$, the dissimilarity function $D$ is defined by:

\begin{equation}
D(I_{a},I_{b}) \Rightarrow [0,1] \subset \mathbb{R} \, ,
\end{equation}
\noindent
satisfying the following properties:
\begin{enumerate}
\item $D(I_{a}, I_{a}) = 0$
\item $D(I_{a}, I_{b}) = 0 \Rightarrow I_{a} = I_{b}$
\item $D(I_{a}, I_{b}) = D(I_{b}, I_{a})$.
\end{enumerate}

Two metrics were employed to assess the algorithms.
The first, called~\gls*{rr}, was used to evaluate the performance in the identification problem~($1$:n), while the second, called~\gls*{auc}, was applied to evaluate the performance in the verification problem~($1$:$1$).
In addition, the methodologies were evaluated in two different setups: first comparing only images acquired by the same device and then using images obtained by two different devices~(cross-sensor).
The algorithms were ranked by the average performance of~\gls{rr} and~\gls{auc}.
The best results are listed in Table~\ref{miche2}.

\begin{table}[!htpb]
\scriptsize
\centering
\caption{Results of the MICHE.II competition. Average between~\gls{rr} and~\gls{auc}. Adapted from~\cite{DeMarsico2017}.}
\label{miche2}
\vspace{0.5mm}
\begin{tabular}{ccccc}
\toprule
\centering Authors                            & All$\times$All   & GS4$\times$GS4   & Ip5$\times$Ip5   & Average\\
\midrule
           Ahmed et al.~\cite{Ahmed2016, Ahmed2017}        & $0.99$      & $1.00$      & $1.00$      & $1.00$  \\
           Ahuja et al.~\cite{Ahuja2016a, Ahuja2017}       & $0.89$      & $0.89$      & $0.96$      & $0.91$  \\
           Raja et al.~\cite{Raja2017}                    & $0.82$      & $0.95$      & $0.83$      & $0.86$  \\
           Abate et al.~\cite{Abate2016, Abate2017}        & $0.79$      & $0.82$      & $0.88$      & $0.83$  \\
           Galdi and Dugelay~\cite{Galdi2016, Galdi2017}        & $0.77$      & $0.78$      & $0.92$      & $0.82$  \\
           Aginako et al.~\cite{Aginako2016a, Aginako2017a}  & $0.78$      & $0.80$      & $0.78$      & $0.79$  \\
           Aginako et al.~\cite{Aginako2016b, Aginako2017b}  & $0.75$      & $0.72$      & $0.77$      & $0.75$  \\
\bottomrule
\end{tabular}
\end{table}

Ahmed et al.~\cite{Ahmed2016, Ahmed2017} proposed the algorithm that achieved the best result.
Their methodology performs the matching of the iris and the periocular region separately and combines the final score values of each approach.
For the iris, they used the rubber sheet model normalization proposed by Daugman~\cite{Daugman1993}.
Then, the iris codes were generated from the normalized images with the 1-D Log-Gabor filter.
The matching was computed with the Hamming distance.
Using only iris information, an~\gls{eer} of $2.12$\% was reached.
Features from the periocular region were extracted with Multi-Block Transitional Local Binary Patterns and the matching was computed with the chi-square distance.
With features from the periocular region, an~\gls{eer} of $2.74$\% was reported.
The outputs of both modalities (iris and periocular) were normalized with z-score and combined with weighted scores.
The weights used for the fusion were $0.55$ for the iris and $0.45$ for the periocular region, yielding an~\gls{eer} of $1.22$\% and an average between~\gls{rr} and~\gls{auc} of~$1.00$.

The best performing approach using only iris information was proposed by Raja et al.~\cite{Raja2017}.
In their method, the iris region was first located through a segmentation method proposed by Raja et al.~\cite{Raja2015} and then normalized using the rubber sheet expansion model~\cite{Daugman2004}.
Each image band (red, green and blue) was divided into several blocks.
The features were extracted from these blocks, as well as from the entire image, using a set of deep sparse filters, resulting in deep sparse histograms.
The histograms of each block and each band were concatenated with the histogram of the entire image, forming the vector of iris features.
The features extracted were used to learn a collaborative subspace, which was employed for matching.
This algorithm achieved the third place in the competition, with an average between~\gls{rr} and~\gls{auc} of~$0.86$ and with~\gls{eer} values of~$0$\% in the images obtained by the iPhone 5S and~$6.55$\% in the images obtained by Samsung~S4.

This competition was the first to evaluate iris recognition using images captured by mobile devices and also to evaluate methodologies applied to the cross-sensor problem, i.e., to recognize images acquired by different sensors.

As in the NICE.II competition, one issue is the scalability evaluation of the evaluated approaches.
Although the reported results are very promising, we have to consider them as preliminary since the test set used for the assessment is very small, containing only $120$ images.
As expected, the best results were attained using iris and periocular region information, however, some approaches that used only iris information achieved better results than others that fused iris and periocular region~information.

\subsection{MIR - Competition on Mobile Iris Recognition}

The BTAS Competition on Mobile Iris Recognition~(MIR2016) was proposed to raise the state of the art of iris recognition algorithms on mobile devices under \gls*{nir} illumination~\cite{Zhang2016}.
Five algorithms, submitted by two participants, were eligible for the evaluation.

A database~(MIR-Train) was made available for training the algorithms and a second database~(MIR-Test) was used for the evaluation.
Both databases were collected under~\gls{nir} illumination.
The images of the two irises were collected simultaneously under an indoor environment.
Three sets of images were obtained, with distances of $20$ cm, $25$ cm, and $30$ cm, and $10$ images for each distance.
The images from both databases were collected in the same session.
The MIR-Train database is composed of $4{,}500$ images from $150$ subjects, while MIR-Test has $12{,}000$ images from $400$ subjects.
All images are grayscale with a resolution of $1968\times1024$ pixels.
The main sources of intra-class variation in the images are due to variations in lighting, eyeglasses and specular reflections, defocus, distance changes, and others.
Differently from NICE.II, the segmentation masks were not provided in MIR2016, thus, the methodologies submitted included iris detection, segmentation, feature extraction, and~matching.

For the evaluation, the organizing committee considered that the left and right irises belong to the same class; thus, a fusion of the matching scores of both irises was performed.
All possible intra-class comparisons (i.e., irises from the same subjects) were implemented to compute the~\gls*{fnmr}.
From each iris class, two samples were randomly selected to calculate the~\gls*{fmr}.
In total, $174{,}000$ intra-class and $319{,}200$ inter-class matches were used.
In cases where intra- or inter-class comparisons could not be performed due to failure enrollment or failure match, a random value between $0$ and $1$ was assigned to the score.
The classification of the participants was performed using the FNMR4 metric, but the~\gls{eer} and DI metrics were also reported.
The FNMR4 metric reports the~\gls{fnmr} value when the~\gls{fmr} equals to $0.0001$.
The~\gls{eer} is the value when~\gls{fnmr} is equal to the~\gls{fmr}, and the DI value is the decidability index, as explained~previously.

The best result was from the Beijing Bata Technology Co. Ltd. reporting FNMR4 = $2.24$\%, \gls{eer} = $1.41$\% and DI = $3.33$.
The methodology, described in~\cite{Zhang2016}, includes four steps: iris detection, preprocessing, feature extraction, and matching.
For iris detection, the face is found using the AdaBoost algorithm~\cite{Viola2004} and eye positions are found by using~\gls{svm}.
Next, to lessen the effect of light reflections, the irises and pupils are detected by the modified Daugmans Integro-Differential operator~\cite{Daugman2004}.
In pre-processing, reflection regions are located and then removed using a threshold and shape information.
Afterward, the iris region is normalized using the method proposed by Daugman~\cite{Daugman1993}.
Eyelashes are also detected and removed using a threshold.
An improvement in image quality is achieved through histogram equalization.
The features were extracted with Gabor wavelet, while \gls{pca} and~\gls{lda} were applied for dimensionality reduction.
The matching was performed using the cosine and Hamming distances, and the results~combined.

The second place was achieved by TigerIT Bangladesh Ltd. with FNMR4 = $7.07$\%, \gls{eer} = $1.29$\% and DI = $3.94$.
The proposed approach also made improvements in image quality through histogram equalization and smoothing.
After pre-processing, the iris was normalized using the rubber sheet model~\cite{Daugman2007}.
Features were then extracted with 2D Gabor wavelets, while the matching was performed employing the Hamming distance.
This methodology was classified in second place since it obtained a higher FNMR4 value than the first one, but the \gls{eer} and DI values were better than those reported by the winning algorithm of the competition.

The MIR2016's main contribution is to be the first competition using~\gls{nir} images acquired by mobile modules, in addition to the construction of a new database containing images from both eyes of each individual.
Unfortunately, the competition did not have many participants and the proposed methodologies consist only of classical literature~techniques.

\subsection{VISOB 1.0 and VISOB 2.0 Competitions on Mobile Ocular Biometric Recognition}
The \visob database was created for the VISOB 1.0 - ICIP 2016 Competition on mobile ocular biometric recognition, whose main objective was to evaluate the progress of research in the area of mobile ocular biometrics at the visible spectrum~\cite{Rattani2016}.
The front cameras of $3$ mobile devices were used to obtain the images: iPhone 5S at 720p resolution, Samsung Note 4 at 1080p resolution and Oppo N1 at 1080p resolution.
The images were captured in $2$ sessions for each one of the 2 visits, which occurred between 2 and 4 weeks, counting in the total $158{,}136$ images from $550$ subjects.
At each visit, it was required that each volunteer (subject) capture their own face using each one of the three mobile devices at a distance between $8$ and $12$ inches from the face.
For each session, images were captured under $3$ light conditions: regular office light, offices lights off but dim ambient lighting still present (dim light) and next to sunlit windows (natural daylight settings).
The collected database was preprocessed using the Viola-Jones eye detector and the region of the image containing the eyes was cropped to a size of $240\times160$ pixels.

The VISOB 1.0 competition was designed to evaluate ocular biometric recognition methodologies using images obtained from mobile devices in visible light on a large-scale database.
The database created and used for the competition was \visob (\visob Database ICIP2016 Challenge Version)~\cite{Rattani2016}.
This database has $158{,}136$ images from $550$ subjects, and is the database of images obtained from mobile devices with the largest number of subjects.
The images were captured by $3$ different devices (iPhone 5S, Oppo N1 and Samsung Note 4) under $3$ different lighting classes: `daylight', `office', and `dim light'.
Four different research groups participated in the competition and $5$ algorithms were submitted.
The metric used to assess the performance of the algorithms was~\gls{eer}.

In almost all competitions, participants submit an algorithm already trained and the evaluation is performed on an unknown portion of the database.
On the other hand, VISOB 1.0 competitors submitted an algorithm that was trained and tested on an unknown portion of the database.
Two different evaluations were carried out.
In the first one (see Table~\ref{icip1}), the algorithms were trained (enrollment) and tested for each device and type of~illumination.

\begin{table}[ht]
\scriptsize
\centering
\caption{EER (\%) rank by device and lighting condition. Adapted from~\cite{Rattani2016}.} 
\label{icip1}
\begin{tabular}{cccc}
\toprule
           \multicolumn{4}{c}{Day light} \\ 
\midrule
            
    Method & iPhone 5S  & Oppo N1  & Samsung Note 4 \\
 
\midrule
NTNU-1 \cite{Raghavendra2016} & $0.06$    & $0.10$  & $0.07$  \\ 
NTNU-2 \cite{Raja2016}        & $0.40$    & $0.43$  & $0.33$  \\ 
ANU                           & $7.67$    & $7.91$  & $8.42$  \\ 
IIITG  \cite{Ahuja2016a}      & $18.98$   & $18.12$ & $15.98$ \\ 
Anonymous                     & $38.09$   & $38.29$ & $62.23$ \\

\midrule
           \multicolumn{4}{c}{Office} \\  
 
\midrule
NTNU-1 \cite{Raghavendra2016} & $0.06$    & $0.04$  & $0.05$   \\ 
NTNU-2 \cite{Raja2016}        & $0.48$    & $0.63$  & $0.49$   \\ 
ANU                           & $10.36$   & $16.01$ & $9.10$   \\ 
IIITG  \cite{Ahuja2016a}      & $19.29$   & $19.79$ & $18.65$  \\ 
Anonymous                     & $35.26$   & $31.69$ & $72.84$  \\

\midrule
           \multicolumn{4}{c}{Dim light} \\  
 
\midrule
NTNU-1 \cite{Raghavendra2016} & $0.06$    & $0.07$  & $0.07$   \\ 
NTNU-2 \cite{Raja2016}        & $0.45$    & $0.16$  & $0.16$   \\ 
ANU                           & $8.44$    & $9.02$  & $11.89$  \\ 
IIITG  \cite{Ahuja2016a}      & $17.54$   & $19.49$ & $23.25$  \\ 
Anonymous                     & $31.06$   & $34.00$ & $67.20$  \\ 

\bottomrule

\end{tabular}
\end{table}

In the second evaluation, the algorithms were trained only with the images from the `office' lighting class for each of the $3$ devices.
To assess the effect of illumination on ocular recognition, the tests were performed with the $3$ types of illumination for each device.
The results are shown in Table~\ref{icip2}.

\begin{table}[ht]
\scriptsize
\centering
\caption{EER (\%) rank by device and lighting condition. The algorithms were trained only with the `office' lighting class~(O) and tested on all the others. Table adapted from~\cite{Rattani2016}.}
\vspace{0.5mm}
\label{icip2}
\begin{tabular}{cccc}
\toprule
                              \multicolumn{4}{c}{iPhone 5S} \\  
\midrule
Method                             & O-O     & O-Day & O-Dim  \\ 
\midrule
NTNU-1 \cite{Raghavendra2016} & $0.06$    & $0.13$  & $0.20$   \\ 
NTNU-2 \cite{Raja2016}        & $0.48$    & $1.82$  & $1.45$   \\ 
ANU                           & $10.36$   & $11.03$ & $16.64$  \\ 
IIITG  \cite{Ahuja2016a}       & $19.29$   & $32.93$ & $45.34$ \\ 
Anonymous                       & $35.26$   & $28.67$ & $42.29$  \\ 

\midrule
                              \multicolumn{4}{c}{Oppo N1} \\ 
\midrule
NTNU-1 \cite{Raghavendra2016} & $0.04$    & $0.10$  & $0.09$  \\ 
NTNU-2 \cite{Raja2016}        & $0.63$   & $1.90$  & $3.34$  \\ 
ANU                           & $16.01$   & $14.75$ & $18.24$ \\ 
IIITG  \cite{Ahuja2016a}      & $19.79$   & $38.24$ & $42.59$ \\ 
Anonymous                       & $31.69$   & $31.21$ & $37.17$ \\

\midrule
                              \multicolumn{4}{c}{Samsung Note 4} \\ 
\midrule
NTNU-1 \cite{Raghavendra2016} & $0.05$    & $0.13$  & $0.10$  \\ 
NTNU-2 \cite{Raja2016}        & $0.49$    & $2.50$  & $4.25$  \\ 
ANU                           & $9.10$    & $13.69$ & $19.57$ \\ 
IIITG  \cite{Ahuja2016a}      & $18.65$   & $34.29$ & $40.21$ \\ 
Anonymous                     & $27.73$   & $24.33$ & $50.74$ \\ 

\bottomrule

\end{tabular}
\end{table}

Raghavendra and Busch~\cite{Raghavendra2016} achieved an~\gls{eer} between $0.06$\% and $0.20$\% in all assessments, obtaining the best result of the competition.
The proposed approach extracted periocular features using Maximum Response~(MR) filters from a bank containing $38$ filters, and a deep neural network learned with a regularized stacked autoencoders~\cite{Raghavendra2016}.
For noise removal, the authors applied a Gaussian filter and performed histogram equalization and image resizing. 
Finally, the classification was performed through a deep neural network based on deeply coupled~autoencoders.

All participants explored features based on the texture of the eye images, extracted from the periocular region.
None of the submitted algorithms extracted features only from the iris.
The organizing committee compared the performance of the algorithms using images obtained only by the same devices, that is, the algorithms were not trained and tested on images from different devices~(cross-sensor).
Thus, the main contributions of this competition were a large database containing images from different sensors and environments, along with the assessments on these different~setups.

The second edition of this competition, called VISOB 2.0, was carried out at IEEE WCCI in 2020~\cite{nguyen2020visob2}.
A new \visob's subset with eye images from $250$ subjects captured by two mobile devices: Samsung Note 4 and Oppo N1, was employed to compare the submitted approaches.
This competition evaluated ocular biometrics recognition methods using stacks of five images in the open-world (subject-independent) protocol in different lighting conditions: Dark, Office, and Daylight.
In the development (training) stage, the competitors were provided with stacks of images from $150$ subjects.
Regarding the subject-independent evaluation, the comparison of the submitted methods was performed employing samples from other $100$ subjects that were not available in the training stage.
The main idea of using multi-frame (stacks) captures for ocular biometrics is to avoid degradation in the images caused by variations in illumination, noise, blur, and user to camera distance.
Two participants submitted algorithms based on deep representations and one based on hand-crafted features.
Table~\ref{visob2} presents the results.
\begin{table*}[ht]
\scriptsize
\centering
\caption{EER (\%) rank by device and lighting condition: Dark (DK), Daylight (DL), and Office (O). Table adapted from~\cite{nguyen2020visob2}.}
\vspace{0.5mm}
\label{visob2}
\resizebox{\linewidth}{!}{
\begin{tabular}{cccccccccc}
\toprule
                              \multicolumn{10}{c}{Samsung Note 4} \\  
\midrule
Method                          & DK-DK   & DK-DL   & DK-O   & DL-DK   & DL-DL   & DL-O     & O-DK    & O-DL    & O-O  \\ 
\midrule
UFPR~\cite{zanlorensi2019cross} & $7.46$  & $10.03$ & $6.66$  & $11.46$ & $7.76$  & $6.72$  & $12.10$ & $8.06$  & $5.26$   \\ 
Bennett University              & $35.01$ & $40.47$ & $42.15$ & $41.45$ & $30.68$ & $34.40$ & $43.65$ & $34.31$ & $27.05$  \\ 
Anonymous                       & $42.07$ & $44.69$ & $43.44$ & $44.41$ & $40.69$ & $42.51$ & $46.09$ & $42.69$ & $39.77$  \\ 

\midrule
                              \multicolumn{10}{c}{Oppo N1} \\ 
\midrule
                                & DK-DK   & DK-DL   & DK-O    & DL-DK   & DL-DL   & DL-O    & O-DK    & O-DL    & O-O  \\ 
\midrule
UFPR~\cite{zanlorensi2019cross} & $6.39$  & $9.40$  & $8.08$  & $8.28$  & $8.11$  & $6.67$  & $9.76$  & $8.65$  & $6.49$   \\ 
Bennett University              & $34.33$ & $40.36$ & $40.90$ & $41.99$ & $29.70$ & $31.91$ & $42.95$ & $31.79$ & $26.21$  \\ 
Anonymous                       & $40.30$ & $44.94$ & $43.71$ & $45.41$ & $42.46$ & $45.14$ & $46.68$ & $45.70$ & $42.05$  \\ 
\bottomrule

\end{tabular}
}
\end{table*}

The rank $1$ algorithm proposed by Zanlorensi et al.~\cite{zanlorensi2019cross} (UFPR) consists of an ensemble of ResNet-50 models ($5$ models, one for each image in the stack) pre-trained for face-recognition using the VGG-Face database.
The authors had previously employed this method for cross-spectral ocular recognition achieving state-of-the-art results on the \crossEyed and the \polyu databases using iris and periocular traits.
In this method, each ResNet-50 model was fine-tuned using the periocular images from \visobtwo.
The only modification in the model was the addition of a fully connected layer containing $256$ neurons at the top to reduce the feature dimensionality.
The training was computed in the identification mode using a Softmax cross-entropy loss function as a prediction layer.
Then, in the evaluation, the prediction layer was removed, and the final combined feature vector with a size of $1280$ ($5\times256$) was used to match samples by computing the cosine distance similarity.
This algorithm's best result was $5.26\%$ of EER using images in the Office vs. Office lighting condition.

The second-place method (\textit{Bennet University}) used Directional Threshold Local Binary Pattern (DTLBP), and wavelet transform for feature extraction (handcrafted features).
Then, the Chi-square distance was employed to compute the similarity between the stack of images.
This method's best result was $26.21\%$ of EER in the Office vs. Office lighting condition.
Finally, the third approach employed the GoogleNet model pre-trained in the ImageNet database for feature extraction and euclidean distance to compute the similarity between the pairs of images.
A Long Short Term Memory (LSTM) model using the euclidean distance scores as input was used to predict whether the pair of images is from the same subject or not.
This method's best result was $39.77$ of EER in the Office vs. Office lighting condition.

To the best of our knowledge, VISOB 2.0 was the first competition to use multi-frame ocular recognition.
The results show that comparison across different illumination was the most difficult for all methods.
The open-world (subject-independent) protocol is a realistic scenario for applications in environments without restriction and prior knowledge of the subjects.
Finally, the submitted algorithms' performance shows that there is still room for improvement in this area.

\subsection{Cross-Eyed - Cross-Spectral Iris/Periocular Competition}

The first Cross-Eyed competition was held in $2016$ at the $8$th IEEE International Conference on Biometrics: Theory, Applications, and Systems~(BTAS).
The aim of the competition was the evaluation of iris and periocular recognition algorithms using images captured at different wavelengths.
The \crossEyed database~\cite{Sequeira2016, Sequeira2017}, employed in the competition, has iris and periocular images obtained simultaneously at the~\gls{vis} and~\gls{nir} wavelengths.

Iris and periocular recognition were evaluated separately.
To avoid the use of iris information in the periocular evaluation, a mask excluding the entire iris region was applied.
Six algorithms submitted by $2$ participants, named \textbf{HH} from Halmstad University and \textbf{NTNU} from Norway Biometrics Laboratory, qualified.
The final evaluation was carried out with another set of images, containing $632$ images from $80$ subjects for periocular recognition and $1{,}280$ images from $160$ subjects for iris~recognition.

The evaluation consisted of enrollment and template matching of intra-class (all~\gls{nir} against all~\gls{vis} images) and inter-class comparisons ($3$~\gls{nir} against $3$~\gls{vis} images -- per class).
A metric based on~\gls{gfar} and~\gls{gffr} was used to verify the performance of the submitted algorithms. 
These metrics generalize the~\gls{fmr} and the~\gls{fnmr}, including~\gls{fte} and~\gls{fta}, complying with the ISO/IEC standards~\cite{ISO19795-1}.
Finally, to compare the algorithms, the GF2 metric (GFRR@GFAR = $0.01$) was~employed.

Halmstad University~(HH) team submitted $3$ algorithms. 
The approaches consist of fusing features extracted with~\gls{safe},~\gls{gabor},~\gls{sift},~\gls{lbp} and~\gls{hog}.
These fusions were evaluated combining scores from images obtained by the same sensors and also by different sensors.
The evaluated algorithms differ by the fusion of different features: $HH_1$~fusing all the features; $HH_2$~fusing~\gls{safe},~\gls{gabor},~\gls{lbp} and~\gls{hog}; and $HH_3$~fusing~\gls{gabor},~\gls{lbp} and~\gls{hog}.
The algorithms were applied only to periocular recognition, and the best performance was achieved by $HH_1$, which achieved an~\gls{eer} of~$0.29$\% and GF2 of~$0.00$\%.
More details can be found in~\cite{Sequeira2016}.

The Norwegian Biometrics Laboratory (NTNU) also submitted $3$ algorithms, which applied the same approaches for feature extraction from iris and periocular traits.
The iris region was located using a technique based on the approach proposed by Raja et. al.~\cite{Raja2015}, and features were extracted through histograms resulting from the multi-scale BSIF, a bank of independent binarized statistical filters.
These histograms were compared using the Chi-Square distance metric.
Lastly, an~\gls{svm} was employed to obtain the fusion and scores corresponding to each filter.
The best approach achieved~\gls{eer} of~$4.84$\% and GF2 of~$14.43$\% in periocular matching, and~\gls{eer} of~$2.78$\% and GF2 of~$3.31$\% in iris~matching.

In 2017, the second edition of this competition was held~\cite{Sequeira2017}.
Similarly to the first competition, the submitted approaches were ranked by~\gls{eer} and GF2 values.
Comparisons in periocular images were made separately for each eye, i. e., the left eyes were compared only with left eyes, and the same for the right eyes.
The main difference was in the database used, as the training set consisted of the \crossEyed database and the test set was made with $55$ subjects.
As in the first competition, the matching protocol consisted of intra- and inter-class comparisons, in which all intra-class comparisons were performed and only $3$ random images per class were applied in the inter-class comparisons.
Results and methodologies of $4$ participants were reported, being $4$ participants with $11$ algorithms for periocular recognition, and $1$ participant with $4$ algorithms for iris recognition.
Two of these participants took part in the first competition, Halmstad University (HH) and Norwegian Biometrics Laboratory (NTNU). The other three competitors were IDIAP from Switzerland, IIT Indore from India, and an~anonymous.

The best method using periocular information was submitted by $HH_1$, which fused features based on~\gls{safe},~\gls{gabor},~\gls{sift},~\gls{lbp} and~\gls{hog}.
Their approach, similar to the one proposed in the first competition, reached~\gls{eer} and GF2 values of $0.82$\% and~$0.74$\%,~respectively. 
For iris recognition, the best results were attained by $NTNU_4$, which was based on BSIF features and reported \gls*{eer} and GF2 values of~$0.05$\% and~$0.00$\%,~respectively.

We point out two main contributions of these competitions: (i)~the release of a new cross-spectral database, and (ii)~the evaluation of several approaches using iris and periocular traits with some promising strategies that can be applied for cross-spectral ocular recognition.
Nevertheless, we also highlight some problems in their evaluation protocols.
First, the periocular evaluation in the second competition only matches left eyes against left eyes and right eyes against right eyes using prior knowledge of the database.
Another problem is the comparison protocol, which uses only $3$ images per class in inter-class comparisons instead of all images without specifically reporting which ones were used.
There is also no information on code availability, and details of the methodologies are lacking, limiting the~reproducibility.
\section{Deep Learning in Ocular Recognition}
\label{sec:deepeye}

Recently, deep learning approaches have won many machine learning competitions, even achieving superhuman visual results in some domains~\cite{lecun2015deep}.
Therefore, in this section, we describe recent works that applied deep learning-based techniques focusing on encoding and matching, i.e., not covering iris preprocessing methods to ocular biometrics including iris, periocular and sclera recognition, gender and age classification, and subject-independent recognition. 

\subsection{Iris approaches}

Liu et al.~\cite{Liu2016} presented one of the first works applying deep learning to iris recognition.
Their approach, called \emph{DeepIris}, was created for recognizing heterogeneous irises captured by different sensors.
The proposed method was based on a \gls*{cnn} model with a bank of Pairwise filters, which learns the similarity between a pair of images.
The evaluation in verification protocol was carried out in the \qFire and CASIA cross-sensor databases and reported promising results with \gls*{eer} of $0.15$\% and $0.31$\%, respectively.

Gangwar and Joshi~\cite{Gangwar2016} also developed a deep learning method for iris verification on the cross-sensor scenario, called \emph{DeepIrisNet.}
They presented two \gls*{cnn} architectures for extracting iris representations and evaluated them using images from the \ndIris and \ndcsIris databases.
The first model was composed of $8$~ standard convolutional, $8$~normalization, and $2$~dropout layers.
The second one, on the other hand, has inception layers~\cite{Szegedy2015} and consists of $5$ convolutional layers, $7$ normalization layers, $2$ inception layers, and $2$ dropout layers.
Compared to the baselines, their methodology reported better robustness on different factors such as the quality of segmentation, rotation, and input, training, and network~sizes.

To demonstrate that generic descriptors can generate discriminant iris features, Nguyen et al.~\cite{Nguyen2018} applied distinct deep learning architectures to \gls*{nir} databases obtained in controlled environments.
They evaluated the following \gls{cnn} models pre-trained using images from the ImageNet database~\cite{Imagenet2009}: AlexNet, VGG, Inception, ResNet and DenseNet.
Iris representations were extracted from normalized images at different depths of each \gls{cnn} architecture, and a multi-class \gls{svm} classifier was employed for the identification task.
Although no fine-tuning process was performed, interesting results were reported in the LG2200 (\ndcsIris) and \casiaThousand databases.
In their experiments, the representations extracted from intermediate layers of the networks reported better results than the representations from deeper layers.

The method proposed by Al-Waisy et al.~\cite{Al-Waisy2017} used left and right irises information for the identification task.
In this approach, each iris was first detected and normalized, and then features were extracted and matched.
Finally, the left and right irises matching scores were fused.
Several \gls{cnn} configurations and architectures were evaluated during the training phase and, based on a validation set, the best one was chosen.
The authors also evaluated other training strategies such as dropout and data augmentation.
Experiments carried out on three databases (i.e.,~\sdumla, \casiaInterval, and \iitdIris) reported a $100$\% rank-1 recognition rate in all of~them.

Generally, an iris recognition system has several preprocessing steps, including segmentation and normalization (using Daugman's approach~\cite{Daugman1993}).
In this context, Zanlorensi et al.~\cite{Zanlorensi2018} analyzed the impact of these steps when extracting deep representations from iris images.
Applying deep representations extracted from an iris bounding box without both segmentation and normalization processes, they reported better results compared to those obtained using normalized and segmented images.
The authors also fine-tuned two pre-trained models for face recognition (i.e.,~VGG-16 and ResNet50) and proposed a data augmentation technique by rotating the iris bounding boxes.
In their experiments, using only iris information, an \gls*{eer} of $13.98$\% (i.e.,~state-of-the-art results) was reached in the NICE.II database.

As the performance of many iris recognition systems is related to the quality of detection and segmentation of the iris, Proen\c{c}a and Neves~\cite{Proenca2017irina} proposed a robust method for inaccurately segmented images.
Their approach consisted of corresponding iris patches between pairs of images, which estimates the probability that two patches belong to the same biological region.
According to the authors, the comparison of these patches can also be performed in cases of bad segmentation and non-linear deformations caused by pupil constriction/dilation.
The following databases were used in the experiments: \casiaLamp, CASIA-IrisV4-Lamp, \casiaThousand, and WVU.
The authors reported results using good quality data as well as data with severe segmentation errors.
Using accurately segmented data, they achieved~\gls{eer} values of $0.6$\% (\casiaLamp), $2.6$\% (CASIA-IrisV4-Lamp), $3.0$\% (\casiaThousand) and~$4.2$\%~(WVU).

In~\cite{Wang2019}, Wang and Kumar claimed that iris features extracted from \gls{cnn} models are generally sparse and can be used for template compression.
In the cross-spectral scenario, the authors evaluated several hashing algorithms to reduce the size of iris templates, reporting that the supervised discrete hashing was the most effective in terms of size and matching.
Features were extracted from normalized iris images with some deep learning architectures, e.g., \gls{cnn} with softmax cross-entropy loss, Siamese network, and Triplet network. 
Promising results were reported by incorporating supervised discrete hashing on the deep representations extracted with a \gls{cnn} model trained with a softmax cross-entropy loss.
The proposed methodology was evaluated on a cross-spectral scenario and achieved \gls{eer} values of $12.41$\% and $6.34$\% on the \polyu and \crossEyed databases, respectively.

Zanlorensi et al.~\cite{zanlorensi2019cross} performed extensive experiments in the cross-spectral scenario applying two \gls{cnn} models: ResNet-50~\cite{Cao2017} and VGG16~\cite{Omkar2015}.
Both models were first pre-trained for face recognition and then fine-tuned using periocular and iris images.
The results of the experiments, carried out in two databases: \crossEyed and \polyu, indicated that it is possible to apply a single \gls{cnn} model to extract discriminant features from images captured at both \gls{nir} and \gls{vis} wavelengths.
The authors also evaluate the impact of representation extraction at different depths from the ResNet-50 model and the use of different weights for fusing iris and periocular features.
For the verification task, their approach achieved state-of-the-art results in both databases on intra- and cross-spectral scenarios using iris, periocular, and fused features.

Wang and Kumar~\cite{wang2019dilated} proposed a deep learning-based approach for iris recognition composed of a residual network combined with dilated convolutional kernels, which optimizes the training process and aggregates contextual information from the iris images.
The proposed method outperformed matching accuracy compared with classical and state-of-the-art approaches for iris recognition.

Ren et al.~\cite{ren2019alignment} proposed a unified feature-level solution regarding intra-class variation in iris recognition caused by variations on illumination, eye angle, and eye gaze.
Their method is composed of an encoder based on a trainable~\gls{vlad} and a deformable convolution.
The authors performed extensive experiments on three iris databases showing that the proposed method outperformed state-of-the-art recognition approaches.

In~\cite{ren2020dynamic}, the authors proposed a framework using \gls{cnn} and graphical models to learn dynamic graph representations in order to solve occlusion that occurs in biometrics.
Their approach consists of build feature graphs based on node representations generated by convolutional features re-crafted using a graph generator establishing connections among spatial parts.
The authors stated that it is possible to adaptively remove the nodes representing the occluded parts using their similarities.
Additionally, a novel strategy to measure the distances of nodes and adjacent matrices was proposed.
Experiments using iris and face databases showed that the proposed framework can achieve promising results on occluded biometrics recognition.

Wei et al.~\cite{wei2019cross} proposed a method using adversarial strategy and sensor-specific information regarding the problem of cross-sensor iris recognition.
Their approach consists of alleviating the degradation in the cross-sensor recognition by applying the adversarial strategy and weakening interference of sensor information.
The method comprises three components: feature extractors containing sensor-specific information to narrow the distribution gap, an alignment feature distribution using~\gls{gan}, and a triplet loss function to reduce the discrepancy of images from different sensors.
The authors validated their method on two cross-sensor iris databases.

\subsection{Periocular approaches}

Luz et al.~\cite{Luz2018} designed a biometric system for the periocular region employing the VGG-16 model~\cite{Omkar2015}.
Promising results were reported by performing transfer learning from the face recognition domain and fine-tuning the system for periocular images.
This model was compared to a model trained from scratch, showing that the proposed transfer learning and fine-tuning processes were crucial for obtaining state-of-the-art results.
The evaluation was performed in the NICE.II and \mobbio databases, reporting~\gls{eer} values of~$5.92$\% and~$5.42$\%,~respectively.

Using a similar methodology, Silva et al.~\cite{Silva2018} fused deep representations from iris and periocular regions by applying the Particle Swarm Optimization (PSO) to reduce the feature vector dimensionality.
The experiments were performed in the NICE.II database and promising results were reported using only iris information and also fusing iris and periocular traits, reaching~\gls{eer} values of~$14.56$\% and~$5.55$\%,~respectively.

Proen\c{c}a and Neves~\cite{Proenca2018} demonstrated that periocular recognition performance can be optimized by first removing the iris and sclera regions. 
The proposed approach, called \emph{Deep-PRWIS}, consists of a \glspl*{cnn} model that automatically defines the regions of interest in the periocular input image.
The input images were generated by cropping the ocular region (iris and sclera) belonging to an individual and pasting the ocular area from another individual in this same region. 
They obtained state-of-the-art results (closed-world protocol) in the \ubirisvTwo and FRGC databases, with \gls*{eer} values of $1.9$\% and~$1.1$\%,~respectively.

Zhao and Kumar~\cite{Zhao2018critical} developed a CNN-based method for periocular verification.
This method first detects eyebrow and eye regions using a~\gls{fcn} and then uses these traits as key regions of interest to extract features from the periocular images.
The authors also developed a verification oriented loss function (\emph{Distance-driven Sigmoid Cross-entropy loss~(DSC))}.
Promising results were reported on six databases both in closed- and open-world protocols, achieving ~\gls{eer} values of $2.26$\%~(\ubipr), $8.59$\%~(FRGC), $7.68$\%~(\focs), $4.90$\%~(\casiaDistance), $0.14$\%~(\ubirisvTwo) and~$1.47$\%~(\visob).

Using~\gls{nir} images acquired by mobile devices, Zhang et al.~\cite{Zhang2018fusion} developed a method based on \gls{cnn} models to generate iris and periocular region features.
A weighted concatenation fused these features.
These weights and also the parameters of convolution filters were learned simultaneously.
In this sense, the joint representation of both traits was optimized. 
They performed experiments in a subset of the \casiaMob database reporting \gls{eer} values of~$1.13$\% (Periocular), $0.96$\%~(Iris) and~$0.60$\%~(Fusion).

\subsection{Sclera approaches}

In ocular biometrics using the sclera region, deep learning techniques are generally applied in the segmentation stage~\cite{lucio2018fully, das2019sclera, naqvi2019scleranet, wang2019sclerasegnet}, helping the recognition system by locating traits as the sclera itself and the iris.
As described by Vitek et al.~\cite{vitek2020comprehensive}, the recognition is often performed using the segmented sclera vasculature by employing key-point and dense-grid descriptors as SIFT, SURF, ORB, and Dense SIFT.
As the sclera is a relatively new ocular biometric trait, there are currently few deep learning-based approaches to perform person recognition~\cite{rot2020scleranet, maheshan2020convsclera}.

Regarding segmentation methods, Lucio et al.~\cite{lucio2018fully} proposed two approaches based on~\gls{fcn} and~\gls{gan} to segment the sclera region. 
Experiments performed on two ocular databases demonstrated that the FCN model achieved better results on a single-sensor configuration.
In contrast, for the cross-sensor scenario, the GAN model reached higher scores.
Wang et al.~\cite{wang2019sclerasegnet} presented the ScleraSegNet, which is based on the U-Net model.
The authors also proposed and compared different embed attention modules in the U-Net model regarding learning discriminative features.
Extensive experiments using three ocular databases showed that the channel-wise attention module was the most effective for performing the segmentation and that data augmentation techniques improved the generalization ability.
Naqvi and Loh~\cite{naqvi2019scleranet} proposed a model for sclera segmentation employing a residual encoder and decoder network, called Sclera-Net.
The authors also addressed sclera segmentation in images acquired by different sensors achieving promising results in this work.
Recent competitions on sclera segmentation~\cite{das2019sclera, vitek2020ssbc} demonstrated that deep learning-based methods achieved the highest results, mainly models based on the U-Net and FCN architectures.
The results reached in these competitions show that sclera segmentation is still an open and challenging problem.

Regarding the sclera recognition task based on deep learning methods, one of the first approaches found in the literature is the ScleraNET~\cite{rot2020scleranet}.
In this work, the authors proposed a multi-task CNN model combining losses from the identity and gaze direction recognition.
This model extracts vasculature descriptors and uses them to infer the identity of the subject.
Promising results were achieved and compared with handcrafted-based methods.
Maheshan et al.~\cite{maheshan2020convsclera} also proposed a method based on~\gls{cnn} for sclera recognition.
The model comprises four convolutional layers, followed by a max-pooling layer and a fully connected layer at the top.
The proposed model was evaluated and compared with the top $2$ ranked algorithms in the SSRBC 2016 Sclera Segmentation and Recognition Competition~\cite{das2016ssrbc} reaching the higher~scores.

\subsection{Gender and age classification}

Soft biometrics, such as gender and age classification, using ocular traits are tasks that have gained attention in research in recent years~\cite{krishnan2020probing, rattani2017gender, rattani2018gender, kuehlkamp2019gender, zanlorensi2020ufprperiocular}.
It can be used as primary biometric information to improve the accuracy of biometric systems~\cite{rattani2017gender}.
A few works in the literature employ ocular traits (iris and periocular region) using~\gls{vis} images for gender and age estimation/classification based on deep learning techniques~\cite{rattani2017age,rattani2018gender, kuehlkamp2019gender, angeloni2019age, zanlorensi2020ufprperiocular}.

Kuehlkanp and Bowyer~\cite{kuehlkamp2019gender} performed extensive experiments using hand-crafted and deep-representations with iris and periocular traits for gender classification.
The results sustain that gender prediction using periocular images is at least $17\%$ more accurate than normalized iris images, regardless of the classifier (hand-crafted or deep representations).
Krishnan et al.~\cite{krishnan2020probing} investigated the fairness of ocular biometrics methods using mobile images across gender.
The evaluation employing the ResNet, LightCNN, and MobileNet models for periocular biometrics presented an equivalent verification performance for males and females.
However, in gender classification, males outperformed females by a difference of~$22.58\%$.

Rattani et al.~\cite{rattani2017age} investigated age classification using~\gls{vis} ocular images acquired by mobile devices.
The proposed method consists of a $6$-layer~\gls{cnn} model comprising convolution, max-pooling, batch-normalization, and fully connected layers.
Ages were grouped into $8$ ranges, and a soft-max activation was employed to compute each group's probability.
Experiments conducted on a $5$-fold cross-validation protocol using only the ocular region (both eyes, eyebrows, and periocular region) reported closer and promising results than full-face methods for age estimation, achieving an accuracy ($\%$) of $46.97\pm2.9$ against $49.5\pm4.4$, respectively.
Angeloni et al.~\cite{angeloni2019age} proposed a multi-stream \gls{cnn} model using facial parts for age classification.
The model consists of $4$ streams, each one for the following traits: eyebrows, eyes, nose, and mouth.
The proposed approach reached better results in accuracy than methods employing images from the entire face.
Furthermore, an ablation study on the method reported that the eyes region was the most important trait to improve the entire approach accuracy.

In a recent work~\cite{zanlorensi2020ufprperiocular}, the authors proposed a multi-task learning network for periocular recognition using~\gls{vis} images acquired by mobile devices.
The~\gls{cnn} architecture was composed of a MobileNetV2 as a base model and $5$ fully connected layers followed by soft-max layers for the following soft biometrics tasks: identity, age, gender, eye side, and smartphone model classification.
The proposed multi-task model reached better results than several~\gls{cnn} architectures for verification and identification tasks on experiments conducted on closed- and open-world (subject-independent) protocols.
Moreover, performing an ablation study, the authors stated that age, gender, and mobile device classification were critical components regarding the accuracy of the method for the identification task.

\subsection{Subject-independent recognition}

The term subject-independent comprises open-world, cross-dataset, and open-set protocols.
There are samples from different subjects in the training and test (evaluation) stages in this scenario.
It is generally employed in the evaluation of methods developed with representation learning.
Regarding deep learning applications for ocular biometrics, the subject-independent evaluation is generally related to the method's robustness, and it is evaluated for the verification task.
Some works compared ocular (iris and periocular) biometric approaches on both subject-dependent (closed-world) and subject-independent (open-world) protocols showing that the latter is the most challenging~\cite{Zhao2018critical, reddy2018ocularnet, zanlorensi2019cross, Proenca2019segmentation, zanlorensi2020ufprperiocular}.
Furthermore, the results reached on VISOB 1.0 (subject-dependent)~\cite{Rattani2016} and VISOB 2.0~\cite{nguyen2020visob2} (subject-independent) competitions sustains this statement.

The methodology proposed in~\cite{Proenca2019segmentation} does not require preprocessing steps, such as iris segmentation and normalization, for iris verification.
In this approach, based on deep learning models, the authors used biologically corresponding patches to discriminate genuine and impostor comparisons in pairs of iris images, similarly to \emph{IRINA}~\cite{Proenca2017irina}.
These patches were learned in the normalized iris images and then remapped into a polar coordinate system.
In this way, only a detected/cropped iris bounding box is required in the matching stage.
The model's input is a pair of images, and the output informs whether they are from the same subject or not.
State-of-the-art results were reported in three~\gls{nir} databases, achieving~\gls{eer} values of $0.6$\%, $3.0$\%, and $6.3$\% in the CASIA-Iris-V4-Lamp, \casiaThousand, and WVU, respectively, in the subject-independent (open-world)~protocol.

Regarding ocular images captured in the~\gls{vis} spectrum, Reddy et al.~\cite{reddy2018ocularnet} proposed a patch-based method employing deep learning networks.
The model crops $6$ overlapping patches from the ocular/periocular region and extracts features employing a small~\gls{cnn} network for each patch.
For a given image pair, the matching is computed by a Euclidean distance between each patch's features.
The final score is then generated by combining the distances with the mean, median, and minimum of patches scores.
Promising results were achieved in $3$~\gls{vis} and $1$ cross-spectral periocular~databases.

Some works~\cite{Wang2019, zanlorensi2020ufprperiocular} evaluated the most employed CNN architectures for the verification task on the subject-independent setting.
These approaches are generally based on Pairwise filters, Siamese, and Triplet networks.
Regarding only these kinds of architectures, in~\cite{Wang2019}, the Siamese model achieved better results than the Triplet network.
On the other hand, in~\cite{zanlorensi2020ufprperiocular} the Pairwise filters network reached better results than the Siamese network.
It is important to note that in both works~\cite{Wang2019, zanlorensi2020ufprperiocular}, even in the subject-independent setting, the best results for the verification task were achieved employing~\gls{cnn} models using a soft-max layer in the training~stage.

\subsection{Final remarks}

Regarding the works described in this section, we point out that some deep learning-based approaches for iris recognition aim to develop end-to-end systems by removing preprocessing steps (e.g., segmentation and normalization) since a failure in such processes would probably affect recognition systems~\cite{Zanlorensi2018, Proenca2017irina, Proenca2019segmentation}.
Several works~\cite{Luz2018, Silva2018, Proenca2018, Zhao2018critical, Zhang2018fusion} show that the periocular region contains discriminant features and can be used, or fused with iris and sclera information, to improve the performance of biometric~systems.
Furthermore, recent works on soft-biometrics for periocular recognition~\cite{krishnan2020probing, rattani2017gender, rattani2018gender, kuehlkamp2019gender, zanlorensi2020ufprperiocular} reported promising results and stated that this kind of information can be used to improve the accuracy of the biometric system.
Finally, biometric systems evaluated in the subject-independent setting are still a challenging task since it is highly affected by the intra- and inter-class variability, especially in~\gls{vis} images collected in unconstrained~scenarios.

For completeness, there are several works and applications with ocular images using deep learning frameworks, such as: spoofing and liveness detection~\cite{Menotti2015, He2016}, left and right iris images recognition~\cite{Du2016}, contact lens detection~\cite{Silva2015}, iris location~\cite{severo2018benchmark}, sclera and iris segmentation~\cite{lucio2018fully, bezerra2018robust}, iris and periocular region detection~\cite{lucio2019simultaneous}, gender classification~\cite{Tapia2017}, iris/periocular biometrics by in-set analysis~\cite{proenca2019inset}, iris recognition using capsule networks~\cite{Zhao2019capsule}, and sensor model identification~\cite{Marra2017}.

\section{Challenges and Future Directions}
\label{sec:future}

In this section, we describe recent challenges and how approaches are being developed to address these issues.
We also point out some future directions and new trends in ocular biometrics.
The challenges and directions presented are as follows:

\begin{itemize}[leftmargin=0cm,itemindent=.3cm,labelwidth=\itemindent,labelsep=0cm,align=left]
    \item \textbf{Scalability}: The term scalability refers to the ability of a biometric system to maintain efficiency (accuracy) even when applied to databases with a large number of images and subjects.
    The largest ~\gls{nir} iris database available in the literature in terms of number of subjects is \casiaThousand~\cite{CASIA2010}, which has $20{,}000$ images taken in a controlled environment from $1{,}000$ subjects. In an uncontrolled environment and with~\gls{vis} ocular images, the largest database is \ufprPerioc~\cite{zanlorensi2020ufprperiocular}, which is composed of $33{,}660$ images from $1{,}122$ subjects.
    Although several proposed methodologies achieve high decidability index in these databases~\cite{Nguyen2018, Proenca2017irina, Proenca2019segmentation, Rattani2016, Raghavendra2016, Raja2016, Ahuja2016a}, indicating that these approaches have impressive and high separation of the intra- and inter-class comparison distribution, can we state that these methodologies are scalable? 
    In this sense, it is necessary to research new methods as well as new databases with a larger number of images/subjects to evaluate the scalability of existing approaches in the literature.

    \item \textbf{Multimodal biometric fusion in the visible spectrum:} 
    The periocular region traits are most utilized when there is a poor quality image of the iris region or part of the face is occluded, which commonly occurs in uncontrolled environments at~\gls{vis} wavelength~\cite{Park2009, Luz2018}.
    A promising solution in such scenarios is the fusion of several biometric traits contained in the images, for example, iris, periocular, ear, and the entire face. 
    In this way, there is still room for improvement in the detection/segmentation of biometric traits contained in the face region and also in algorithms for fusing features extracted from these traits into various levels, as feature extraction, matching score, and decision~\cite{Ross2003}.

    There are few publicly available multimodal databases, and those available combine ocular modalities with other popular biometric traits, such as face or speech.
    Researchers aiming to evaluate the fusion of ocular biometric modalities against other less common modalities need to create their own database or build a chimerical one. In~\cite{lopes2019chimericaldataset}, a protocol for the creation and evaluation of multimodal biometric chimerical databases is discussed. Although evaluation on chimeric databases is not an ideal condition, it may be an alternative to an initial/preliminary investigation~\cite{lopes2019chimericaldataset}.

    \item \textbf{Multi-session:} Regarding real-world applications, databases containing images captured in more than one session in an uncontrolled environment can be used to analyze the robustness of biometric systems, as images obtained at distinctive sessions often present high intra-class variations caused by environmental changes, lighting, distance, and other noises such as occlusion, reflection, shadow, focus, off-angle, etc. 
    Images obtained at different sessions are important for evaluating the variation of biometric traits through time and also the effect of imaging in different environments, e.g., indoor and outdoor environment, daylight (natural), office light (artificial), among others.
    Some studies~\cite{Rattani2016, Raghavendra2016, Raja2016, Ahuja2016a, Liu2016, Gangwar2016} show that images obtained in different sessions have a greater impact on the recognition of~\gls{vis} images than of~\gls{nir} images.
    This is because~\gls{nir} images are generally obtained under controlled environments while~\gls{vis} images are taken under uncontrolled environments and because the near-infrared spectrum best highlights the internal features of the iris~\cite{Liu2016, Gangwar2016, Rattani2016, Nalla2017, Nguyen2018}. 

    \item \textbf{Deep ocular representations:} Several works have explored strategies by modifying and/or evaluating input images for iris feature extraction using~\gls{cnn} models~\cite{Liu2016, Gangwar2016, Proenca2017irina, Zanlorensi2018, Proenca2018, Proenca2019segmentation, Zhao2018critical}.
    Zanlorensi et al.~\cite{Zanlorensi2018} showed that~\gls{cnn} models can extract more discriminating features from the iris region using images without classic preprocessing steps such as normalization and segmentation for noise removal.
    Proen\c{c}a and Neves~\cite{Proenca2018} demonstrated that by removing information from the eyeball region (iris and sclera), representations extracted from the periocular region yields better results in biometric systems and also that it is possible to train~\gls{cnn} models to define the region-of-interest automatically (i.e.,~ignoring the information contained in the eyeball region) in an implicit way.
    Recent works~\cite{Liu2016, Proenca2017irina, Proenca2019segmentation} attained promising results by training~\gls{cnn} models to detect/learn similar regions in image pairs using Pairwise filters, that is, using a pair of iris images as input and a binary output informing if the images belong to the same class.
    Features extracted from these models generally achieve better results when compared to models trained for verification tasks, e.g., Triplet and Siamese networks~\cite{Wang2019}.
    Within this context, we can state that improvements can be made by exploring different approaches to feed the~\gls{cnn} models and also by exploring different architectures and loss~functions.

    \item \textbf{Mobile Cross-sensor images:} Recently, some mobile (smartphones) ocular databases have been created (\miche, \vssiris, \csip and \visob) to study the use of images from different sensors and environments in ocular biometrics.
    The images contained in these databases are captured by the volunteer himself in uncontrolled environments and have several variabilities caused by occlusion, shadows, lighting, defocus, distance, pose, gait, resolution, image quality (usually affected by the environment lighting), among others.
    Due to these characteristics, iris recognition using such images may not be reliable; thus some methodologies using periocular region information have been proposed~\cite{Ahuja2016a, Ahmed2016, Aginako2017a, Ahuja2017}.
    Another factor evaluated in these databases is the recognition using cross-sensor images, i.e., the matching of features extracted from images captured by different sensors.
    In this scenario, the largest database in terms of subjects is~\visob~\cite{Rattani2016} with $550$ subjects and $158{,}136$ images captured using $3$ different sensors.
    In terms of number of sensors, the largest database is~\csip~\cite{Santos2015} with $7$ different sensors, however, it contains only $2{,}004$ images from $50$ subjects.
    A next step may be to create a mobile ocular database containing a larger number of different sensor models (compared to existing ones) in different sessions.
    Such a database can be used to assess biometric systems regarding the noise signature of each camera, as well as the variations caused by the environments (sessions). It is essential that this database has a large number of subjects, e.g., at least $1{,}000$ (\casiaThousand).

    \item \textbf{Cross-spectral scenario:}
    A recent challenge that still has room for improvement is the application of ocular biometric systems in a cross-spectral scenario/setting.
    The term cross-spectral refers to the matching of features extracted from images captured at different wavelengths, usually~\gls{vis} images against~\gls{nir} ones.
    Based on the configuration of the experiment, the feature extraction training step can be performed using images obtained at only one wavelength (\gls{vis} or~\gls{nir}) or both (\gls{vis} and~\gls{nir}).
    The challenge of this scenario is that the features present in~\gls{nir} images are not always the same as those extracted in~\gls{vis} images.
    We can mention some recent competitions and approaches that have been developed in this scenario~\cite{Sequeira2016, Sequeira2017, Nalla2017, Wang2019, zanlorensi2019cross}.

    \item \textbf{Protocols: closed-world, open-world, and cross-dataset:} 
    Deep learning-based biometric systems consist of learning distinct features from traits. 
    Those features can be used to generate a similar (or dissimilar) score to perform a verification task or can be fed to a classifier in order to perform an identification task. How learned features should be used is highly associated with the evaluation protocol.
    Ideally, experiments should be performed on different protocols such as closed-world, open-world, and cross-dataset to evaluate the robustness against different scenarios and the generalization ability of these models.
    Note that open-world and cross-dataset can also be reported as the subject-invariant protocol. 
   
    In the \textit{closed-world protocol}, different samples from the same classes are present in the training and test sets, facilitating the use of supervised classifiers for the biometric identification task. 
    This means that the system is not able to handle new classes. 
    This type of system (closed-world) is usually evaluated with accuracy or recognition rate metrics.

    The \textit{open-world protocol} must have samples from different classes in the training and test sets.
    Within this protocol, the biometric system must provide a score to allow the calculation of similarity (or dissimilarity) from a pair of samples. 
    The evaluation of open-world protocol is usually done with the biometric verification task. 
    Although the verification process is often performed in a pair-wise fashion (1:1) and, by definition, in the verification task, the identity of the subject to be verified is known a priori, in biometric competitions this information is also used to generate scores from impostor pairs in order to emulate spoofing attacks~\cite{Proenca2012, Zhang2016, Rattani2016, Sequeira2016, Sequeira2017, DeMarsico2017}.
    The number of impostor pairs is often the absolute majority during the assessment, which makes open-world protocol very challenging. 
    The evaluation of competitions using the open-world protocol are usually done by EER, AUC, or decidability.
   
    Finally, the \textit{cross-dataset protocol} consists of performing training and testing using data acquired with different devices (sensors). Therefore, two or more different databases are employed. 
    This type of evaluation brings another kind of issues in real environments, for example, the influence of sensor quality and light spectrum sensitivity. 
    Feature extraction methods should be robust enough to represent the samples under different conditions.
    
    In our opinion, the closed-world protocol is the most challenging one, followed by open-world and closed-world, respectively.
    We emphasize that, in order to assess robustness and generalization ability, all protocols should be considered by future~competitions.
    
    \item \textbf{Soft biometrics:}
    Considering that several recent periocular databases have labeled soft-biometrics as age, gender, race, and eye color~\cite{zanlorensi2020ufprperiocular, Rattani2016, Padole2012}, such information can be used to improve the biometric systems' performance/accuracy.
    The few works found in the literature exploring this kind of information generally present promising improvements by using soft biometrics data for both the training and evaluation stage~\cite{Marra2017, zanlorensi2020ufprperiocular}. 
    Recent research in this area aims to detect/classify these attributes~\cite{krishnan2020probing, rattani2017gender, rattani2018gender, kuehlkamp2019gender, zanlorensi2020ufprperiocular}.
    With the advancement of these approaches, we believe that soft biometrics will increasingly become an alternative to improve biometric systems' performance, serving as a pre-matching process to return the most likely matching samples.

\end{itemize}
\section{Conclusion}
\label{sec:conclusion}

This work presented a survey of databases and competitions for ocular recognition.
For each database, we described information such as image acquisition protocols, creation year, acquisition environment, images wavelength, number of images and subjects, and modality.
The databases were described and divided into three subsections:~\gls{nir},~\gls{vis} and cross-spectral, and multimodal databases.
Such databases included iris and periocular images for different applications such as recognition, liveness detection, spoofing, contact lens detection, synthetic iris creation, among others.
We also presented recent competitions in iris and periocular recognition and described the approaches that achieved the best results.
The top-ranked methodologies using only iris traits and also the better overall result (i.e., using both iris and periocular information) were detailed.
Finally, we reviewed recent and promising works that applied deep learning frameworks to ocular recognition tasks.

We also described recent challenges and approaches to these issues, point out some future directions and new trends in the ocular biometrics.
In this context, some research directions can be highlighted, for example, recognition using \textbf{(i)}~images taken in an uncontrolled environment \cite{Proenca2012, DeMarsico2017, Rattani2016}, \textbf{(ii})~images obtained from mobile devices at the \gls*{vis} wavelength~\cite{DeMarsico2017, Rattani2016}, and \textbf{(iii)}~cross-spectrum images~\cite{Wang2019, zanlorensi2019cross}.
Aiming to study the scalability of deep iris and periocular features and images obtained by smartphones, a very close real-world scenario, it may be interesting to create a database containing a larger number of devices/sensors and subjects compared with current databases \cite{Kim2016, DeMarsico2015, Raja2015, Santos2015, Rattani2016, zanlorensi2020ufprperiocular}, since the largest one in terms of sensors (\csip) have only $2{,}004$ images captured from $50$ subjects by $7$ different devices and the largest database in terms of subject (\ufprPerioc) have $33{,}660$ images captured from $1{,}122$ subjects by $196$ different sensors (not cross-sensor).
The application of machine learning techniques for segmentation, feature extraction, and recognition can still be greatly explored \cite{DeMarsico2016} since promising results have been achieved using them~\cite{Menotti2015, Gangwar2016, Liu2016, He2016, Du2016}.
Other directions that also deserve attention are ocular recognition at distance, liveness detection, multimodal ocular biometrics, and soft biometrics, which can be used to improve the performance of ocular biometric systems.

\section{Acknowledgments}
This work was supported by grants from the National Council for Scientific and Technological Development~(CNPq) (grant numbers~428333/2016-8, 313423/2017-2 and 306684/2018-2), and the Coordination for the Improvement of Higher Education Personnel~(CAPES)~(Social Demand Program), both funding agencies from Brazil.

\scriptsize
\balance
\setlength{\bibsep}{3pt}
\bibliographystyle{IEEEbib}
\bibliography{paper}

\end{document}